\documentclass{article}

\usepackage{arxiv}

\usepackage[utf8]{inputenc} 
\usepackage[T1]{fontenc}    
\usepackage{hyperref}       
\usepackage{url}            
\usepackage{booktabs}       
\usepackage{amsfonts}       
\usepackage{nicefrac}       
\usepackage{microtype}      
\usepackage{lipsum}         
\usepackage{xcolor}
\usepackage{graphicx}
\usepackage{natbib}
\usepackage{doi}
\usepackage{adjustbox}
\usepackage{caption,subcaption}
\usepackage[acronym]{glossaries}
\usepackage{tcolorbox}
\usepackage{fontawesome}
\usepackage{wrapfig}
\usepackage{array}

\usepackage{amsmath,amsfonts,bm,dsfont,amsthm}

\newcommand{\ot}[0]{{\text{OT}}}

\def\eqref#1{equation~\ref{#1}}

\def\1{\bm{1}}

\DeclareMathAlphabet{\mathsfit}{\encodingdefault}{\sfdefault}{m}{sl}
\SetMathAlphabet{\mathsfit}{bold}{\encodingdefault}{\sfdefault}{bx}{n}

\renewcommand{\min}[1]{\underset{#1}{\text{min}}\,}

\renewcommand{\inf}[1]{\underset{#1}{\text{inf}}\,}

\newcommand{\argmin}[1]{\underset{#1}{\text{arg min}}\,}

\newcommand{\arginf}[1]{\underset{#1}{\text{arg inf}}\,}

\newcommand{\expectation}[1]{\underset{#1}{\mathbb{E}}}

\newtheorem{theorem}{Theorem}[section]

\newtheorem{definition}{Definition}[section]
\newtheorem{remark}{Remark}[section]

\usepackage{cleveref}       

\usepackage[linesnumbered,ruled,vlined]{algorithm2e}

\SetKwComment{Comment}{\color{green!50!black}// }{}

\SetKwProg{Function}{function}{}{}

\title{Unsupervised Anomaly Detection through Mass Repulsing Optimal Transport}

\date{}

\newif\ifuniqueAffiliation
\uniqueAffiliationtrue

\ifuniqueAffiliation 
\author{
Eduardo Fernandes Montesuma\thanks{These authors have contributed equally to this work.}\\
CEA, List\\
Université Paris-Saclay\\
F-91120 Palaiseau, France
\And
Adel El Habazi$^{*}$\\
CEA, List\\
Université Paris-Saclay\\
F-91120 Palaiseau, France
\And
Fred Ngolè Mboula\\
CEA, List\\
Université Paris-Saclay\\
F-91120 Palaiseau, France
}
\else
\usepackage{authblk}

\newbox{\orcid}\sbox{\orcid}{\includegraphics[scale=0.06]{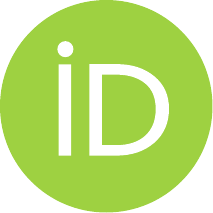}} 
\author[1]{%
	\href{https://orcid.org/0000-0000-0000-0000}{\usebox{\orcid}\hspace{1mm}David S.~Hippocampus\thanks{\texttt{hippo@cs.cranberry-lemon.edu}}}%
}
\author[1,2]{%
	\href{https://orcid.org/0000-0000-0000-0000}{\usebox{\orcid}\hspace{1mm}Elias D.~Striatum\thanks{\texttt{stariate@ee.mount-sheikh.edu}}}%
}
\affil[1]{Department of Computer Science, Cranberry-Lemon University, Pittsburgh, PA 15213}
\affil[2]{Department of Electrical Engineering, Mount-Sheikh University, Santa Narimana, Levand}
\fi

\renewcommand{\shorttitle}{Anomaly Detection through Mass Repulsing Optimal Transport}

\hypersetup{
pdftitle={Unsupervised Anomaly Detection through Mass Repulsing Optimal Transport},
pdfsubject={stats.ML},
pdfauthor={Montesuma, El Habazi and Ngole Mboula},
pdfkeywords={Optimal Transport,Anomaly Detection,Fault Detection,Cross Domain Fault Detection},
}

\newacronym{ot}{OT}{Optimal Transport}
\newacronym{munot}{MROT}{Mass Repulsing Optimal Transport}
\newacronym{kde}{KDE}{Kernel Density Estimation}
\newacronym{cdf}{CDF}{Cumulative Distribution Function}
\newacronym{ad}{AD}{Anomaly Detection}
\newacronym{pca}{PCA}{Principal Component Analysis}
\newacronym{svm}{SVM}{Support Vector Machine}
\newacronym{rf}{RF}{Random Forest}
\newacronym{kl}{KL}{Kullback-Leibler}
\newacronym{uot}{UOT}{Unbalanced Optimal Transport}

\newacronym{ols}{OLS}{Ordinary Least Squares}
\newacronym{svr}{SVR}{Support Vector Regression}

\newacronym{xgboost}{XGBoost}{eXtreme Gradient Boosting}

\newacronym{te}{TE}{Tennessee Eastmann}

\newacronym{knn}{$k-$NN}{$k-$Nearest Neighbors}
\newacronym{lof}{LOF}{Local Outlier Factor}
\newacronym{cblof}{CBLOF}{Clustering-based LOF}
\newacronym{isof}{IsoF}{Isolation Forest}
\newacronym{ocsvm}{OCSVM}{One Class SVM}

\newacronym{ecod}{ECOD}{Empirical Cumulative Distribution Functions}
\newacronym{copod}{COPOD}{Copula-Based OD}
\newacronym{hbos}{HBOS}{Histogram-based Outlier score}
\newacronym{loda}{LODA}{Lightweight On-line Detector of Anomalies}
\newacronym{dte}{DTE}{Diffusion Time Estimation}

\newacronym{nlp}{NLP}{Natural Language Processing}
\newacronym{tsne}{TSNE}{t-Stochastic Neighbor Embeddings}

\begin{document}
\maketitle

\maketitle

\begin{abstract}
Detecting anomalies in datasets is a longstanding problem in machine learning. In this context, anomalies are defined as a sample that significantly deviates from the remaining data. Meanwhile, Optimal Transport (OT) is a field of mathematics concerned with the transportation, between two probability distribution, at least effort. In classical OT, the optimal transportation strategy of a distribution to itself is the identity, i.e., each sample keeps its mass. In this paper, we tackle anomaly detection by forcing samples to displace its mass, while keeping the least effort objective. We call this new transportation problem Mass Repulsing Optimal Transport (MROT). Naturally, samples lying in low density regions of space will be forced to displace mass very far, incurring a higher transportation cost. In contrast, samples on high density regions are able to send their mass just outside an \emph{exclusion zone}. We use these concepts to design a new anomaly score. Through a series of experiments in existing benchmarks, and fault detection problems, we show that our algorithm improves over existing methods.

\begin{tcolorbox}

This is a pre-print. This paper was accepted at TMLR, see \url{https://openreview.net/forum?id=PPGJ3EvENv}. Our code is available at,

\begin{center}
\faGithub\,\,\,\url{https://github.com/eddardd/MROT}
\end{center}
\end{tcolorbox}

\end{abstract}

\section{Introduction}\label{sec:intro}

An anomaly, or an outlier, is a data point that is significantly different from the remaining data~\citep{aggarwal2017introduction}, to such an extent that it was likely generated by a different mechanism~\citep{hawkins1980identification}. From the perspective of machine learning, \gls{ad} wants to determine, from a set of examples, which ones are likely anomalies, typically through an interpretable score. This problem finds applications in many different fields, such as medicine~\citep{salem2013sensor}, cyber-security~\citep{siddiqui2019detecting}, and system monitoring~\citep{isermann2006fault}, to name a few. As reviewed in~\cite{han2022adbench}, existing techniques for \gls{ad} are usually divided into unsupervised, semi-supervised and supervised approaches, with an increasing need for labeled data. In this paper, we focus on unsupervised \gls{ad}, which does not need further labeling effort in constituting datasets.

Meanwhile, \gls{ot} is a field of mathematics concerned with the transportation of masses at least effort~\cite{villani2009optimal}. In its modern treatment, one can conceptualize transportation problems between probability distributions, which has made an important impact in machine learning research~\citep{montesuma2024recentadvancesoptimaltransport}. Hence, \gls{ot} is an appealing tool, as it can be estimated non-parametrically from samples from probability distributions. Likewise, the plethora of computational tools for computing \gls{ot}~\citep{peyre2020computationaloptimaltransport,flamary2021pot} further stresses its usability.

\begin{figure}[ht]
    \centering
    \includegraphics[width=\linewidth]{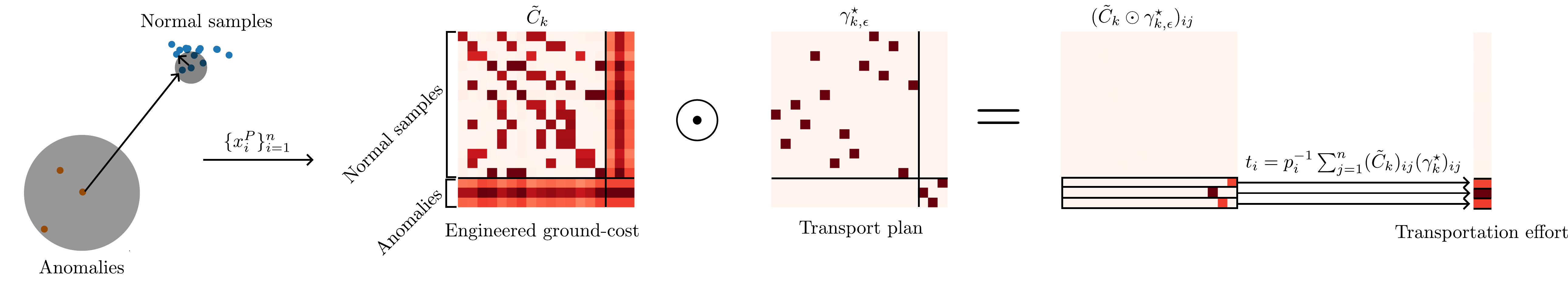}
    \caption{Mass Repulsive Optimal Transport. Our method engineers the ground-cost $c(x_{i}^{(P)},x_{j}^{(P)})$ between samples of a distribution $P$. That way, samples are forced to send their mass outside a vicinity defined through its nearest neighbors (shaded gray areas). This design leads to a transportation plan, such that anomalous samples incur in an higher transportation effort than normal samples. We use these efforts to build an interpretable and generalizable anomaly score.}
    \label{fig:mrot}
\end{figure}

In this context, the application of \gls{ot} for \gls{ad} is not straightforward, as we are interested in analyzing a single probability distribution. We present a new \gls{ot} problem between a distribution and itself, by restricting \emph{where} a sample can send its mass to. More specifically, we design an \emph{exclusion zone}, prohibiting samples from keeping its mass, or sending its mass to a small vicinity. Especially, we assume that anomalies lie in low-density regions of space, with only a few samples in their vicinity. By restricting the transport of mass in the vicinity of samples, anomalies are naturally forced to send their mass to the high-density region, which is assumed to be far away from the anomaly samples. Hence, anomalies will have an overall higher \emph{transportation effort} than normal samples, which can find nearby samples outside the exclusion zone. We show a conceptual illustration of our method in Figure~\ref{fig:mrot}.

Although \gls{ot} has been previously used to compare and aggregate signals in the context of \gls{ad}~\citep{alaoui2019unsupervised,alaoui2020semi}, to the best of our knowledge ours is the first general purpose \gls{ot}-based algorithm for \gls{ad}. Furthermore, we propose a new \gls{ot} problem based on the engineering of the ground-cost, which has links to \gls{ot} with repulsive costs~\citep{di2017optimal}. We benchmark our algorithm in a comprehensive list of datasets, including tabular, computer vision, and natural language processing proposed by~\cite{han2022adbench}, besides fault detection~\citep{reinartz2021extended,montesuma2024benchmarking}.

This paper is organized as follows. Section~\ref{sec:related_work} discusses related work in \gls{ad} and \gls{ot}. Section~\ref{sec:methodology} discusses our proposed method, called \gls{munot}. Section~\ref{sec:experiments} covers our experiments. Finally, section~\ref{sec:conclusion} concludes this paper.
\section{Related Work}\label{sec:related_work}

\noindent\textbf{Anomaly Detection.} Following~\cite{han2022adbench}, \gls{ad} methods can be mainly divided into 3 categories. First, supervised methods consider \gls{ad} through the lens of binary classification under class imbalance. Second, semi-supervised methods either consider partially labeled data, or \emph{labeled normal samples}, so that an algorithm can characterize what a normal sample is. The third, more challenging category is unsupervised \gls{ad}, where the training data contains both anomalies and normal samples and labels are not available. This paper considers precisely the last setting. Next, we review ideas in unsupervised \gls{ad}.

The first kind of methods rely on encoder-decode architectures to detect anomalies. The insight is that, by embedding data in a lower dimensional space, anomalies can be detected via the reconstruction error of the auto-encoding function. This is the principle of \gls{pca}~\cite{shyu2003novel}, which employs linear encoding and decoding functions, but also of kernelized versions~\cite{scholkopf1997kernel,hoffmann2007kernel}, as well as neural nets~\cite{vincent2008extracting,bengio2013representation}, which rely on non-linear embedding techniques.

The second type of strategies are based on the paradigm of 1-class classification. As~\cite{scholkopf1999support} puts, the idea is to define a function that outputs $0$ on a small, dense region of the space where normal samples lie, and $1$ elsewhere. In this context,~\cite{scholkopf1997kernel} extends the celebrated \gls{svm} to \gls{ad}, and~\cite{liu2008isolation} extends \glspl{rf} of~\cite{breiman2001random}.

A third kind of approaches focuses on \emph{neighborhoods and clustering}, to model the data underlying probability distribution, especially through the density of samples over the space. This is the case of \gls{knn}~\citep{ramaswamy2000efficient}, who use distances and nearest neighbors to determine anomalies, \gls{lof}~\cite{breunig2000lof}, who devised a score that measures the local deviation of a sample with respect its neighbors. Finally, \gls{cblof}~\cite{he2003discovering} proposed an extension of \gls{lof} based on the relative sizes of clusters within the data.

\noindent\textbf{Deep Learning-based Anomaly Detection.} As~\cite{pang2021deep} reviews, deep learning is used for \gls{ad} in mainly 3 ways: for i) learning powerful feature extractors, ii) learning representations of normality, and iii) building an end-to-end anomaly score. Besides iii, i and ii allow for the synergy between the deep and traditional \gls{ad} algorithms. In fact, one can use classical \gls{ad} strategies over the latent space of a neural net. We explore this possibility in Section~\ref{ex:adbench}.

In addition to this possibility, we give a few examples of how deep learning has contributed to \gls{ad}. For example,~\cite{ruff2018deep} proposed a strategy for performing \emph{deep} one class classification, which echoes the ideas of \gls{ocsvm}~\cite{scholkopf1999support}.

Furthermore, generative modeling has deeply contributed to \gls{ad}. For instance, generative adversarial nets~\citep{goodfellow2014generative} can serve as a modeling step for capturing features of the underlying distribution $P$. For instance, in~\cite{schlegl2019f} and~\cite{zenati2018efficient}, the authors combine reconstruction errors with the distance of the predictions of the discriminator for building an anomaly score.~\cite{akcay2018ganomaly} takes this idea further. This strategy creates a encoder-decoder-encoder architecture for the generator network. The model is trained to model normal images, by mapping them to the latent space, reconstructing back the image, then mapping again to the latent space. Anomalous images are then detected via the usual reconstruction loss, plus the distance between latent representations and the difference between discriminator's predictions. Another direction consists of using variational inference~\citep{dias2020anomaly} or diffusion models~\citep{livernoche2024on}.

As we cover in the next section, our method uses nearest neighbors and \gls{ot} to model, non-parametrically, the density of samples over the space. More specifically, we prohibit samples in \gls{ot} to keep their mass, or sending it over a region of space defined through their \gls{knn}. Differently from~\cite{ramaswamy2000efficient} and~\cite{breunig2000lof}, we do not rely on distances, which might not have a meaning in high-dimensions. Rather, we rely on the effort of transportation, measured through the samples' mass times the ground-cost.

\noindent\textbf{Optimal Transport with Repulsive Costs.} In general \gls{ot} theory (see, e.g., Section~\ref{sec:ot} below), samples are transported based on a ground-cost that measures how expensive it is to move masses between measures. In its original conception by~\cite{monge1781memoire} and~\cite{kantorovich1942transfer}, this ground cost is the Euclidean distance $c(\mathbf{x}_{1}, \mathbf{x}_{2}) = \lVert \mathbf{x}_{1} - \mathbf{x}_{2} \rVert_{2}^{2}$. As reviewed in~\cite{di2017optimal}, it may be interesting to consider \emph{repulsive costs}, i.e. functions $c$ that are big when $\mathbf{x}_{1}$ and $\mathbf{x}_{2}$ are close to each other and small otherwise. An example of such costs, arising from physics, is the Coulomb interaction $c(\mathbf{x}_{1}, \mathbf{x}_{2}) = (\lVert \mathbf{x}_{1} - \mathbf{x}_{2}\rVert_{2})^{-1}$. Still following~\cite{di2017optimal}, these kinds of transportation problems proved useful in physics, e.g., for quantum mechanics and $N-$body systems.

In this paper, we consider a different kind of repulsive cost, which we call the mass repulsive cost (see, e.g., Section~\ref{sec:munot} below). Our notion of cost defines an exclusion zone, based on its nearest neighbors, where sending mass is too costly. As a result, our approach captures the local characteristics of the probability distribution being analyzed, especially its density. A special characteristic of our approach is to give a sense of the transportation from a distribution to itself.

\noindent\textbf{Optimal Transport-based Anomaly Detection.} Previous works~\citep{alaoui2019unsupervised,alaoui2020semi} have considered \gls{ot} for \gls{ad}. These works proposed a distance-based detection mechanism, in which isolated samples are considered anomalies. \gls{ot} contributes to this setting, by defining a rich metric between samples. Especially, these works considered \gls{ad} in time series data, and \gls{ot} is used to compute distances between those time series in the frequency domain, under a Chebyshev ground cost. In comparison with these methods, ours is notably general purpose, that is, we do not assume data to be time series. Instead of using a Chebyshev ground cost, we model the \gls{ad} problem with a repulsive cost.
\section{Proposed Method}\label{sec:methodology}

\subsection{Optimal Transport}\label{sec:ot}

\gls{ot} is a field of mathematics, concerned with the displacement of mass between a source measure, and a target measure, at least effort. In the following, we cover the principles of \gls{ot} in continuous and discrete settings. We refer readers to~\cite{peyre2020computationaloptimaltransport} for a computational exposition of the main concepts, and~\cite{montesuma2024recentadvancesoptimaltransport} for applications in machine learning. In the following, we are particularly interested in the formulation by~\cite{kantorovich1942transfer}, which is defined as,
\begin{definition}{(Kantorovich Formulation)}\label{def:Kantorovich_continuous}
Let $P$ and $Q$ be 2 probability distributions over a set $\mathcal{X}$. Let $c:\mathcal{X}\times\mathcal{X}\rightarrow\mathbb{R}$ be a \emph{ground-cost}, measuring the effort of transporting units of mass from $x$ to $y$. Let $\Gamma(P, Q) = \{\gamma \in \mathbb{P}(\mathcal{X}\times\mathcal{X}): \int_{\mathcal{X}}\gamma(x,B)dx = Q(B) \text{ and }\int_{\mathcal{X}}\gamma(A,y)dy = P(A)\}$ be the set of \emph{transportation plans}, whose marginals are $P$ and $Q$. The optimal transportation problem is written as,
\begin{align}
    \gamma^{\star} = \text{OT}(P,Q) = \arginf{\gamma \in \Gamma(P, Q)}\int_{\mathcal{X}\times\mathcal{X}}c(x,y)d\gamma(x,y).\label{eq:kantorovich_continuous}
\end{align}
\end{definition}

Equation~\ref{eq:kantorovich_continuous} defines the transportation problem as an infinite dimensional linear program on the variable $\gamma$, called transport plan. In our case, instead of having access to a closed-form $P$, one has samples $\{\mathbf{x}_{i}^{(P)}\}_{i=1}^{n}$, each $\mathbf{x}_{i}^{(P)} \sim P$ with probability $p_{i}$. In such cases, $P$ may be approximated with an empirical measure,
\begin{align}
    \hat{P}(\mathbf{x}) = \sum_{i=1}^{n}p_{i}\delta(\mathbf{x} - \mathbf{x}_{i}^{(P)}),\,\sum_{i=1}^{n}p_{i}=1,p_{i}\,\geq 1, \forall i.\label{eq:emp_approx}
\end{align}

Plugging back equation~\ref{eq:emp_approx} into equation~\ref{eq:kantorovich_continuous} leads to a finite linear program,
\begin{align}
    \hat{\gamma} = \argmin{\gamma \in \Gamma(\mathbf{p},\mathbf{q})}\langle \gamma, \mathbf{C} \rangle_{F} = \sum_{i=1}^{n}\sum_{j=1}^{m}\gamma_{ij}\underbrace{c(\mathbf{x}_{i}^{(P)}, \mathbf{x}_{j}^{(Q)})}_{C_{ij}},\,\Gamma(\mathbf{p},\mathbf{q}) = \{\gamma:\sum_{i}\gamma_{ij}=q_{j},\,\sum_{i}\gamma_{ij}=p_{i}\text{ and }\gamma_{ij} \geq 0\}\label{eq:kantorovich_discrete}
\end{align}

\begin{wrapfigure}[16]{r}{0.5\linewidth}
\centering
\includegraphics[width=0.48\linewidth]{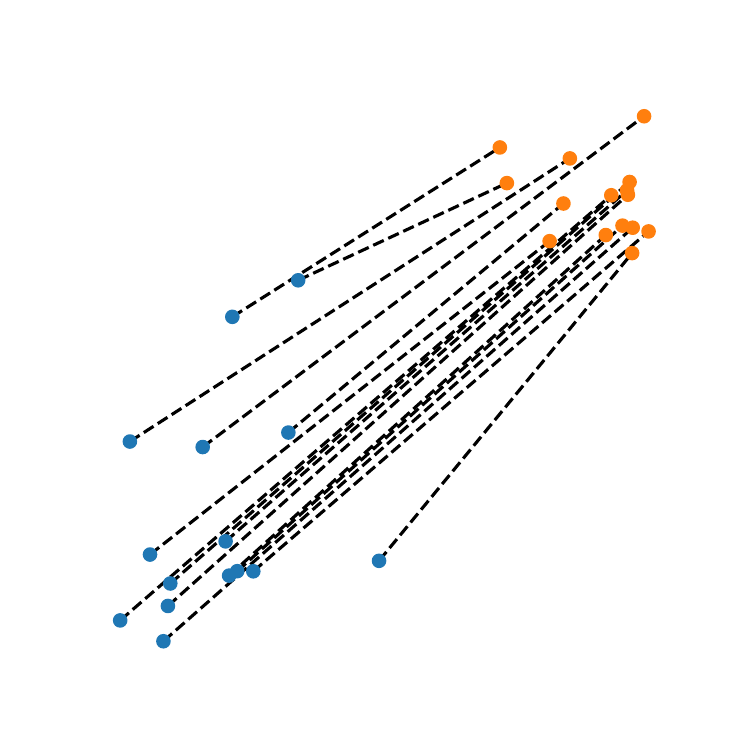}
\includegraphics[width=0.48\linewidth]{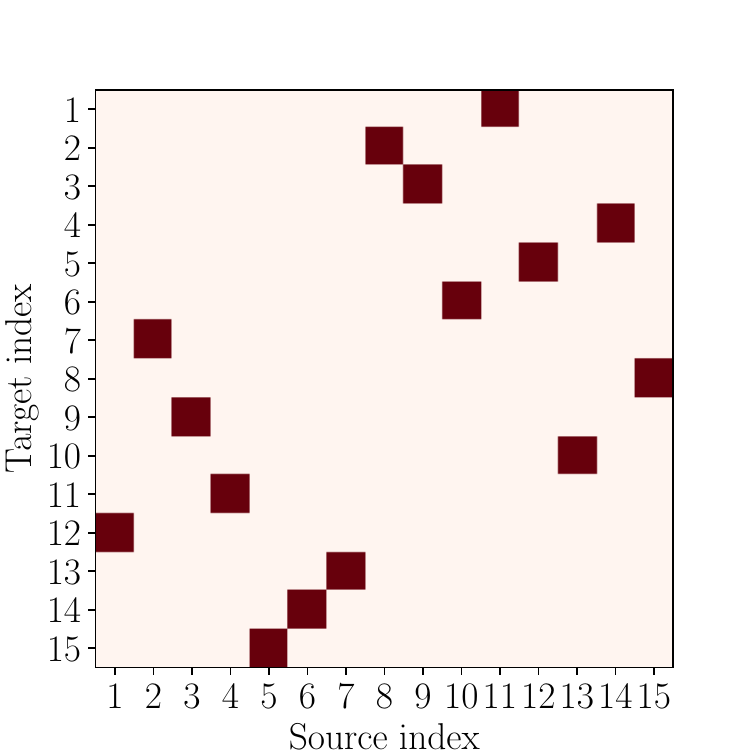}
\captionof{figure}{Optimal transport plan $\gamma$ between samples of $P$ (blue) and $Q$ (orange). On the left, we connect samples $(i,j)$ for which $\gamma_{ij}^{\star} > 0$. On the right, we show the entries of $\gamma^{\star}$.}
\label{fig:ot}
\end{wrapfigure}
where the optimization variables are the coefficients $\gamma_{ij}$ of the transport plan. Problem~\ref{eq:kantorovich_discrete} is a finite linear program, hence the solution $\gamma$ is a sparse matrix with at most $n + m - 1$ non-zero elements~\citep{peyre2020computationaloptimaltransport}. Solving it through the celebrated Simplex algorithm~\citep{dantzig1983reminiscences}, which has computational complexity $\mathcal{O}(n^{3}\log n)$ and storage complexity $\mathcal{O}(n^{2})$ (i.e., storing each $\gamma_{ij}$). We show an example of empirical \gls{ot} in Figure~\ref{fig:ot}.

A faster alternative was introduced by~\cite{cuturi2013sinkhorn}, who shown that adding an entropic regularization to equation~\ref{eq:kantorovich_discrete} leads to a problem that can be solved through Sinkhorn's algorithm~\citep{sinkhorn1967diagonal}. From a continuous perspective, this is equivalent to penalizing \gls{kl} divergence between $\gamma$, and the trivial coupling $P \otimes Q = P(x)Q(y)$. This regularization term is related to the entropy of $\gamma$ as discussed in~\cite[Chapter 4]{peyre2020computationaloptimaltransport}, hence this problem is called entropic \gls{ot}. Next, we define the entropic \gls{ot} problem,

\begin{definition}{(Entropic Optimal Transport)}\label{def:entropic_ot}
    Under the same conditions of Definition~\ref{def:Kantorovich_continuous}, let $\epsilon \geq 0$ be an entropic penalty. The entropic \gls{ot} problem is given by,
    \begin{align}
        \gamma^{\star}_{\epsilon} = OT_{\epsilon}(P, Q) = \arginf{\gamma \in \Gamma(P, Q)} \int_{\mathcal{X}\times\mathcal{X}}c(x,y)d\gamma(x,y) + \epsilon \text{KL}(\gamma|P\otimes Q),\label{eq:eot}
    \end{align}
    where $\text{KL}(\gamma|\xi) = \int_{\mathcal{X} \times \mathcal{X}}\log\biggr(\dfrac{d\gamma}{d\xi}(x,y)\biggr)d\gamma(x,y) + \int_{\mathcal{X}\times\mathcal{X}}(d\xi(x,y) - d\gamma(x,y))$ is the \gls{kl} divergence between measures $\gamma$ and $\xi$~\citep[Eq. 4.10]{peyre2020computationaloptimaltransport}.
\end{definition}

We can obtain an equivalent discrete formulation by plugging back equation~\ref{eq:emp_approx} into~\ref{eq:eot}, which leads to,
\begin{align}
    \hat{\gamma}_{\epsilon} = \argmin{\gamma \in \Gamma(\mathbf{p},\mathbf{q})}\langle \gamma, \mathbf{C} \rangle_{F} - \epsilon H(\gamma) = \sum_{i=1}^{n}\sum_{j=1}^{m}\gamma_{ij}C_{ij} + \epsilon\sum_{i=1}^{n}\sum_{j=1}^{m}\gamma_{ij}(\log \gamma_{ij} - 1),\label{eq:eot_discrete}
\end{align}
where $H(\gamma)$ denotes the entropy of the transportation plan. Since equation~\ref{eq:eot_discrete} relies on the~\cite{sinkhorn1967diagonal} algorithm rather than linear programming, it has $\mathcal{O}(Ln^{2})$ computational complexity, where $L$ is the number of iterations. In general, the \gls{kl} and entropic terms in equations~\ref{eq:eot} and~\ref{eq:eot_discrete} have a smoothing effect on the transportation plan $\gamma_{\epsilon}$. As a result, $\hat{\gamma}_{\epsilon}$ has more non-zero elements than $\hat{\gamma}$.

In the next section, we explore a new transportation problem with a single probability distribution. This problem is understood as the transportation of $P$ to itself, when the samples form $P$ are forced to send their mass outside of their immediate neighborhood.

\subsection{Mass Repulsing Optimal Transport}\label{sec:munot}

In this section, we propose a new \gls{ot} problem called \gls{munot}. This problem is inspired by the Kantorovich formulation, described in section~\ref{sec:ot}. However, instead of considering two different probability distributions $P$, $Q$, it considers the transport of $P$ to itself. Due to the properties of \gls{ot}, if we consider the \gls{ot} plan $\gamma^{\star} = \ot(P, P)$, it is supported in the set $\{(x, x), x \in \mathcal{X}\}$, that is, each point keeps its own mass~\citep{santambrogio2015optimal}. This motivates our new problem, in which we force points to \emph{repell} its mass. Henceforth, we assume that $c$ comes from a metric $(\mathcal{X}, d)$, i.e., $c(x, y) = d(x, y)^{p}$, $p \in [1, +\infty)$. For pairs $x \in \mathcal{X}$, $y \in \mathcal{X}$, and $L \in [0, +\infty)$,
\begin{align}
    \tilde{c}(x, y) = \begin{cases}
        c(x, y) & \text{ if }y \not\in \mathcal{N}(x),\\
        L & \text{ otherwise},
    \end{cases}\label{eq:engineered_cost}
\end{align}
where $\mathcal{N}(x)$ denotes the vicinity of $x$ (e.g., $k-$nearest neighbors, or an $\rho-$ball centered at $x$). In the next remark, we give some geometric intuition behind the cost engineering, as well as some motivation for the choice of $L$. More generally, in the next section we give a theoretically motivated choice for $L$.

\begin{figure}[ht]
    \centering
    \includegraphics[width=\linewidth]{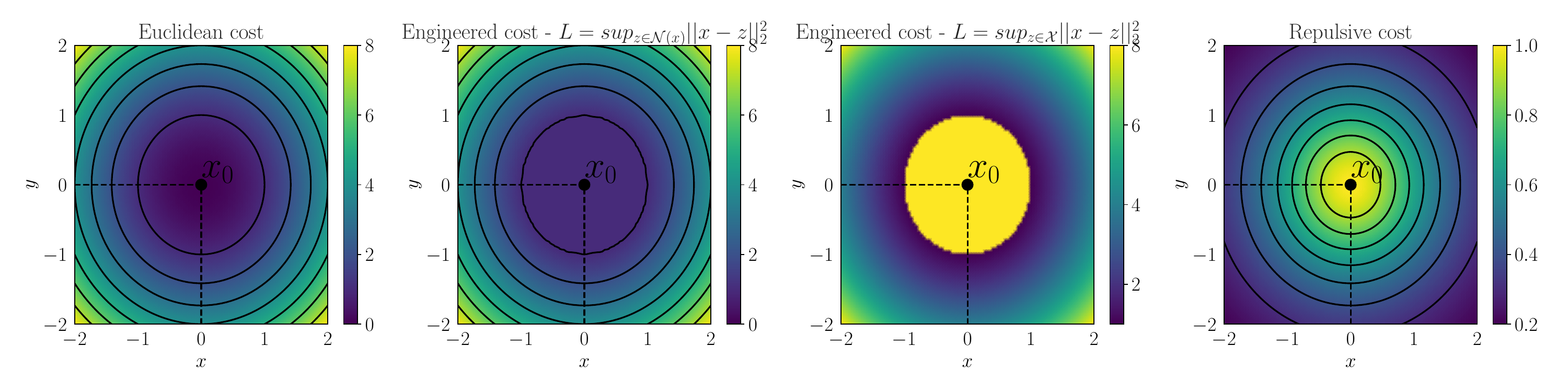}
    \caption{Comparison between different ground costs for $x_0 = (0, 0)$. From left to right: Squared euclidean cost $c(x, x_0) = \lVert x - x_0 \rVert_{2}^{2}$, Engineered cost (c.f., equation~\ref{eq:engineered_cost}) for two choices of $L$, and repulsive cost.}
    \label{fig:comparison-costs}
\end{figure}

\begin{remark}{(Geometric intuition)} 
Assume $x_0 \in \mathcal{X} = [-2, 2]^{2}$ fixed. Our cost engineering strategy replaces the base ground cost $c(x_0, y)$, in the neighborhood $\mathcal{N}(x_0)$ with a value $L$. Assume, for argumentation purposes, that $L \rightarrow +\infty$. This choice means that it is infinitely costly for each point to keep its own mass. As a result, samples are encouraged to send their mass to the immediate outside of their neighborhood  -- thus, we call this idea \emph{mass repulsive} \gls{ot}.

To ground some intuition, we show, in Figure~\ref{fig:comparison-costs}, a comparison using the squared Euclidean distance as the ground-cost, i.e., $c(x, y) = \lVert x - y \rVert_{2}^{2}$. Alongside the base cost, we show our engineered cost, and the repulsive cost $(1 + c(x, y))^{-1}$. Here, we define the neighborhood of $x_0$ through the $\rho-$closed ball centered at $x_0$, i.e., $\mathcal{N}(x_0) = \{y : \mathbb{R}^{2}:\lVert x - y \rVert_{2} \leq \rho\}$. Likewise, we use $\rho = 1$, and, $L = \text{sup}_{z \in \mathcal{N}(x_0)}c(x, z)$. This particular choice leads to nice theoretical properties for the continuous \gls{munot} problem (see equation~\ref{eq:mrot_continuous} below).

Due to the particular choices we made, the supremum takes its value on the frontier $\partial \mathcal{N}(x_0) = \{z : \mathbb{R}^{2}: \lVert x_0 - z \rVert_{2} = \rho\}$, in which case $c(x_0, z) = \rho^{2}$, independently of $x_0$. This choice for $L$ effectively flattens the ground cost around the vicinity of $x_0$. As we show in the next section, this is essential for ensuring the existence of the continuous \gls{munot} problem. In practice, it is beneficial to set $L$ to a large, finite constant to ensure points are transported outside $\mathcal{N}(x_0)$. Note that, as we remark in the next section, this does not harm the existence of the \textbf{discrete} \gls{munot} solution. We show an example of this choice by setting $L = \text{sup}_{z \in \mathcal{X}}\lVert x - z \rVert_{2}^{2}$.
\end{remark}

Our principle of mass repulsion is different from repulsive costs~\citep{di2017optimal}, which are designed to model the interaction between particles in multimarginal \gls{ot}~\citep{pass2015multi}. Indeed, we design a transportation problem from a probability distribution to itself. Hence, while repulsive costs incentive transportation towards distant points in space, our mass repulsing cost induces transportation just outside an exclusion zone. Henceforth, we focus on $p=2$ for the Euclidean distance. The ground-cost we propose essentially defines an \emph{exclusion zone} around points $\mathbf{x}$, where these points are discouraged from sending their mass.

Our main hypothesis for anomaly detection is that anomalous points lie in low density regions of $P$. On the one hand, If these points are forced to send its mass outside its vicinity, it will be forced to send it to high density regions of $P$ -- otherwise, mass conservation in \gls{ot} would not hold. On the other hand, \emph{if the exclusion zone is smaller than high density regions of $P$}, points on these regions will be sent close-by. As a result, anomalous points will have a higher transportation cost (c.f., equations~\ref{eq:teffort} and~\ref{eq:score} below) than normal points. We thus consider the following \gls{ot} problem,
\begin{align}
    \gamma^{\star}_{\epsilon} &= \text{MROT}_{\epsilon}(P) = \inf{\gamma \in \Gamma(P, P)}\int\tilde{c}(x,y)d\gamma(x,y)+\epsilon\text{KL}(\gamma|P\otimes P),\label{eq:mrot_continuous}
\end{align}
which, like the continuous entropic \gls{ot} problem in definition~\ref{def:entropic_ot}, also admits a discrete version when $P$ is approximated empirically through equation~\ref{eq:emp_approx},
\begin{align}
    \gamma^{\star}_{\epsilon} &= \text{MROT}_{\epsilon}(\hat{P}) = \min{\gamma \in \Gamma(\mathbf{p},\mathbf{p})}\langle \gamma, \tilde{\mathbf{C}}\rangle_{F} - \epsilon H(\gamma).\label{eq:discrete_munot}
\end{align}
Considering again Figure~\ref{fig:comparison-costs}, we focus on comparing the commonly used squared Euclidean cost, our engineered cost, and a repulsive cost associated with the Coulomb interaction~\citep{di2017optimal}. In traditional \gls{ot}, sending mass to distant regions of space is costly. As a result, points are encouraged to keep their own mass as close as possible. Consequently, \gls{ot} from $P$ to itself is the trivial plan $\gamma = \text{Id}$.

For our engineered cost, we use $\mathcal{N}(\mathbf{x}) = \{\mathbf{y}: \lVert \mathbf{x} - \mathbf{y} \rVert_{2}^{2} \leq \rho\}$, $\rho = 1$. Both our engineered cost $\tilde{c}$ and the repulsive cost $(1 - c(x, x_0))^{-1}$ assign high costs to points near $x_0$. In particular, within the neighborhood $\mathcal{N}(x_0)$, our engineered cost $\tilde{c}(x,x_0)$ reaches its maximum. Hence, there is an important difference between \gls{munot} and \gls{ot} with repulsive costs. Our engineered cost promotes transportation \emph{to the immediate region outside $\mathcal{N}$}, whereas the repulsive cost drives samples as far as possible from $x_0$. As we discussed previously, this feature of the ground-cost allows us to build an anomaly score.

\subsection{Theoretical Analysis}

In this section, we provide a proof for the existence of \gls{munot} plans. While at least one solution of equation~\ref{eq:mrot_continuous} exists, its uniqueness is not, in general, guaranteed. In this section, we denote the set of measures over a set $\mathcal{X}$ by $\mathcal{P}(\mathcal{X})$, and the power-set of $\mathcal{X}$ by $\mathbb{P}(\mathcal{X})$.

From classical \gls{ot} theory, the Kantorovich problem (c.f., equation~\ref{eq:kantorovich_continuous}) has a solution under mild conditions on the ambient space $\mathcal{X}$ and the ground-cost $c$, see~\cite{santambrogio2015optimal}. Indeed, $\mathcal{X}$ is required to be Polish (i.e., complete, separable metric space) and $c$ must be lower semi-continuous. For simplicity, we state our results on $d-$dimensional Euclidean spaces, which are the setting for our experiments. We start by re-stating~\cite[Theorem 1.5]{santambrogio2015optimal},

\begin{theorem}{(Existence of Optimal Transport plans)}
    Let $\mathcal{X}$ be a Polish space, $P$, and $Q \in \mathcal{P}(\mathcal{X})$, and $c:\mathcal{X}\times\mathcal{X} \rightarrow [0, +\infty]$ be lower semi-continuous. Then equation~\ref{eq:kantorovich_continuous} admits a solution.
\end{theorem}

Since we are dealing with $d-$dimensional Euclidean spaces, the condition on the ambient space is already satisfied. It remains the question of the lower semi-continuity of our engineered cost. In the next theorem, we prove that, for the squared Euclidean cost, a solution for \gls{munot} exists,

\begin{theorem}
    Let $c(x, y) = \lVert x- y \rVert_{2}^{2}$ and $\mathcal{N}(x) = \{y \in \mathbb{R}^{d}: \lVert x - y \rVert_{2} \leq \rho\}$. Then the continuous \gls{munot} problem (equation~\ref{eq:mrot_continuous}) admits a solution.
\end{theorem}

\begin{proof}
    Let $x_{0} \in \mathbb{R}^{d}$ and $\rho > 0$ be given. Our proof relies on the decomposition of $y \in \mathbb{R}^{d}$ in 3 regions: i) $\lVert x_0 - y \rVert_{2} > \rho$, ii) $\lVert x_0 - y \rVert_{2} < \rho$, and iii) $\lVert x - y \rVert_{2} = \rho$. For i) and ii), $\tilde{c}$ is a continuous function of its inputs. Indeed, for $\lVert x_0 - y \rVert_{2} > \rho$, $\tilde{c}(x, y) = \lVert x_0 - y\rVert_{2}^{2}$. Likewise, for $\lVert x_0 - y\rVert_{2} < \rho$, we have $\tilde{c}(x, y) = \rho^{2}$. Now, we need to prove that $\tilde{c}$ is continuous on $\lVert x_0 - y \rVert_{2} = \rho$. We sub-divide this into 2 cases. First, assume we have a sequence $(y_{n})$, $y_n \rightarrow y$, $y_n \not\in \mathcal{N}(x)$. Since $y_n \not\in \mathcal{N}(x_0)$, we have $\tilde{c}(x_0,y_n) = \lVert x_0 - y_n\rVert_{2}^{2}$. From the continuity of the squared Euclidean distance, $\tilde{c}(x_0,y_n) \rightarrow \lVert x_0 - y \rVert_{2}^{2} = \rho^{2}$. Second, assume we have a sequence $(y_n'), y_n' \rightarrow y$, $y_n' \in \mathcal{N}(x_0)$. Then, $\tilde{c}(x_0, y_n') = \rho^{2} \rightarrow \tilde{c}(x_0, y) = \rho^{2}$. Hence, $\tilde{c}$ is continuous.
\end{proof}

The principles used in the previous theorem can be generalized to other costs and neighborhoods, as long as: i) the base cost $c(x, y)$ is continuous, ii) the neighborhood $\mathcal{N}(x)$ is compact, and iii) the supremum $\text{sup}_{z \in \mathcal{N}(x)} c(x, z)$ is attained in the boundary of $\mathcal{N}(x)$, $\forall x \in \mathcal{X}$. The bottom line of our analysis is that, in general, one needs to carefully engineer the ground cost, through the choice of the base ground cost $c$, and the neighborhood function $\mathcal{N}$.

Although existence may hold, the fact that the engineered cost introduces a flat region on the neighborhood $\mathcal{N}(x)$ of $x$ makes it difficult to establish the uniqueness of the \gls{munot} plan. We recall from~\cite[Section 2.6]{figalli2021invitation} that, in classical \gls{ot}, the existence of a unique transport plan is related to the existence of a Monge map between $P$ and $Q$. From~\cite[2.7.1]{figalli2021invitation}, this means that we need 3 conditions: i) $x \mapsto \tilde{c}(x, y)$ is differentiable, ii) $y \mapsto \nabla_{x} \tilde{c}(x,y)$ is injective, and iii) for $\rho > 0$, and $B_{\rho} = \{x \in \mathcal{X}: \lVert x \rVert \leq \rho\}$, $|\nabla \tilde{c}(x, y)| \leq C_{\rho}$ for every $x \in B_\rho$. As it turns out, neither of these conditions holds for our engineered cost, so the \gls{munot} plan is likely not unique.

Now, let us focus on the discrete case, which is of practical interest to us. Note that $\Gamma(\mathbf{p}, \mathbf{p})$ is non-empty as $\gamma_{ij} = p_{i}p_{j} \in \Gamma(\mathbf{p},\mathbf{p})$. Furthermore, this set is compact, and the objective in equation~\ref{eq:discrete_munot} is continuous. Then, by Weierstrass theorem~\citep[Box 1.1]{santambrogio2015optimal}, there exists a minimizer to the discrete setting. Here, note that we did not have to impose continuity or lower semi-continuity on the neighborhood function $\mathcal{N}$ or the ground cost $\tilde{c}$. As in classical \gls{ot}, the solution to this problem is not unique, since the objective function is not strictly convex. Following this line, adding the entropic penalty (i.e., $\epsilon > 0$ in equation~\ref{eq:discrete_munot}) makes the objective strictly convex~\citep[Section 4.1]{peyre2020computationaloptimaltransport}, which guarantees the uniqueness of the \gls{ot} plan, and hence, the uniqueness of the \gls{munot} plan.


\subsection{Building and Generalizing an Anomaly Score}\label{sec:building-anomaly-score}

Through \gls{munot}, we want to build a score for the samples based on how anomalous they are. Assuming that anomalies lie in a low density region of space, our \gls{munot} problem forces
those samples to send their mass to distant parts of the feature space. In contrast, normal samples can send their mass to the immediate neighborhood outside the exclusion zone. As a result, we can sort out normal from anomalous samples using the transportation effort,
\begin{align}
    \mathcal{T}(x) = \expectation{y \sim \gamma(\cdot|x)}[\tilde{c}_{k}(x,y)] = \int_{\mathcal{X}}\tilde{c}_{k}(x, y)d\gamma(y|x),\label{eq:teffort}
\end{align}
where $\gamma(y|x)$ corresponds to the conditional probability, calculated through the joint $\gamma(x, y)$, given $x$. For empirical measures, this quantity can be calculated as follows,
\begin{align}
    t_{i} = \mathcal{T}(x_{i}^{(P)}) = \sum_{j=1}^{n}\dfrac{\gamma_{ij}^{\star}}{p_{i}}\tilde{c}_{k}(x_{i}^{(P)}, x_{j}^{(P)}),\label{eq:score}
\end{align}
where $p_{i}$ is the importance of the $i-$th sample (e.g, $p_{i} = n^{-1}$ for uniform importance). Interestingly, equation~\ref{eq:score} is similar to the barycentric map, widely used in domain adaptation~\cite{courty2016optimal}. $\mathcal{T}$ has 2 shortcomings as an anomaly score. First, it is hardly interpretable, as its range depends on the choice of ground-cost $c$. Second, it is only defined in the support of $\hat{P}$. We offer a solution to both of these problems.

Concerning interpretability, we propose to transform it using the \gls{cdf} of its values. Let $P_{\mathcal{T}}$ be the probability distribution associated with $\mathcal{T}(x)$. The \gls{cdf} is simply $F_{\mathcal{T}}(t) = P_{\mathcal{T}}((-\infty, t))$. Naturally, since $P_{\mathcal{T}}$ is not available, it may be approximated from samples $\{t_{i}\}_{i=1}^{n}$, obtained through equation~\ref{eq:score}. We do so through, \gls{kde},
\begin{align}
    \hat{P}_{\mathcal{T}}(t) &= \dfrac{1}{n\sigma}\sum_{i=1}^{n}\phi\biggr{(}\dfrac{t-t_{i}}{\sigma}\biggr{)}\text{, and, }\hat{F}_{\mathcal{T}}(t) = \int_{-\infty}^{t}\hat{P}_{\mathcal{T}}(s)ds\label{eq:kde}
\end{align}
where $\phi$ is a kernel function (e.g., the Gaussian kernel $\phi(x) = \exp(-\nicefrac{x^{2}}{2})$), and $\sigma$ is the bandwidth, controlling the smoothness of $\hat{P}_{\mathcal{T}}$ and determined through Scott's rule~\citep{scott1979optimal}. Equation~\ref{eq:kde} is an approximation for the density of transportation efforts $\{t_{i}\}_{i=1}^{n}$. The \gls{cdf} is appealing, as it is a monotonic function over transportation effort $t \in \mathbb{R}$, and it takes values on $[0, 1]$, both of which are desirable for an anomaly score.

The question of how to \emph{extrapolate} the anomaly score for new samples remains. For example, even if we use the \gls{cdf} values as anomaly scores, we need to recalculate $t = \hat{\mathcal{T}}(\mathbf{x})$ for a new sample $\mathbf{x} \sim P$, which is challenging, as $\hat{\mathcal{T}}$ is only defined on the support of $\hat{P}$. A naive approach would be to append $\mathbf{x}$ to the set $\{\mathbf{x}_{i}^{(P)}\}_{i=1}^{n}$ and solve a \gls{munot} problem again. Naturally, this is not feasible, as solving an \gls{ot} problem for each new sample is computationally expensive.

\begin{minipage}[t]{0.48\textwidth}
In this paper, we present the more efficient idea of modeling the relationship $\mathbf{x} \mapsto \hat{F}_{\mathcal{T}}(\hat{\mathcal{T}}(\mathbf{x}))$ from the samples we receive, that is, we create a labeled data set $\{\mathbf{x}_{i}^{(P)}, t_{i}\}_{i=1}^{n}$, where $t_{i} = \hat{\mathcal{T}}(\mathbf{x}_{i}^{(P)})$. Our anomaly score comes, then, through a function $\psi(\mathbf{x}_{i}^{(P)})$ fit to the labeled dataset through regression. The fitting $\psi$ can be done with standard regression tools, such as \gls{ols}, \gls{svr}~\citep{smola2004tutorial}, nearest neighbors or gradient boosting~\citep{friedman2002stochastic}. In general, one can expect the relationship between $\mathbf{x}_{i}^{(P)}$ and $\hat{F}_{\mathcal{T}}(\hat{\mathcal{T}}(\mathbf{x}_{i}^{(P)}))$ to be non-linear, hence, it is generally necessary to use a non-linear regression model. We show a summary of our strategy in Algorithm~\ref{alg:mrot}.
\end{minipage}
\hfill
\begin{minipage}[t]{0.48\textwidth}
\begin{algorithm}[H]
  \caption{Mass Repulsive Optimal Transport.}
  \label{alg:mrot}
  \Function{\textcolor{black}{mrot($\mathbf{X}^{(P)}, \epsilon$)}}{
    $\gamma_{k,\epsilon}^{\star} = \text{MROT}_{k,\epsilon}(\hat{P})$\;
    $t_{i} \leftarrow \sum_{j=1}^{n}(\nicefrac{(\gamma_{\epsilon}^{\star})_{ij}}{p_{i}})\tilde{c}(\mathbf{x}_{i}^{(P)},\mathbf{x}_{j}^{(P)})$\;
    $\hat{P}_{\mathcal{T}} \leftarrow \text{KDE}(\{t_{i}\}_{i=1}^{n})$\;
    $\psi \leftarrow \text{Regression}(\{\mathbf{x}_{i}^{(P)}, \hat{F}_{\mathcal{T}}(t_{i})\}_{i=1}^{n})$\;
    \Return{$\psi$}\;
  }
\end{algorithm}
\end{minipage}


\section{Experiments}\label{sec:experiments}

We divide our experiments in 3 parts. Section~\ref{ex:adbench} shows our results on AdBench~\citep{han2022adbench}. Section~\ref{ex:cdfd} shows our experiments in fault detection on the Tennessee Eastman Process~\citep{montesuma2024benchmarking,reinartz2021extended}. Finally, Section~\ref{sec:ablations} explores the robustness of our methods to various hyper-parameters and design choices.

In the following, we consider adaptive neighborhoods $\mathcal{N}_{k}(\mathbf{x})$ which take the $k-$nearest neighborhood of $\mathbf{x}$, and set $C_{ij} = \text{max}_{\ell=1,\cdots,n} C_{i\ell}$, where $j \in \mathcal{N}_{k}(\mathbf{x}_{i}^{(P)})$. Before going through our experiments, we show a toy example, going through all the steps in our algorithm. See Figure~\ref{fig:comparison-costs} for a comparison.

\textbf{An introductory example.} Before comparing our method with prior art, we give an introductory example that illustrates how we create anomaly scores out of samples. In this example, we sample normal examples $\mathbf{x}_{i} \sim \mathcal{N}(\mathbf{0}, 0.25\mathbf{I}_{2})$, and anomalous samples $\mathbf{y}_{j} \sim \mathcal{N}([-3, -3], 0.01\mathbf{I}_{2})$, where $\mathbf{I}_{2}$ is a $2 \times 2$ identity matrix. The dataset for this toy example consists of the concatenation $\mathbf{X}^{(P)} = \{\mathbf{x}_{1},\cdots,\mathbf{x}_{n},\mathbf{y}_{1},\cdots,\mathbf{y}_{m}\}$, where $n=500$ and $m=25$, which means that roughly $5\%$ of the total number of samples are anomalies. These samples are shown in Figure~\ref{fig:toy-example}.

\begin{wrapfigure}[15]{r}{0.3\linewidth}
    \centering
    \includegraphics[width=\linewidth]{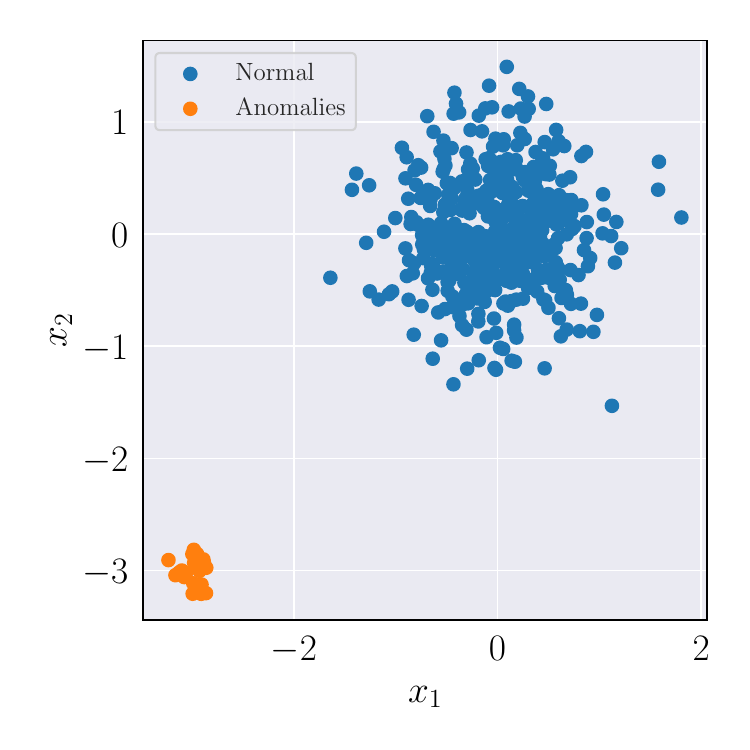}
    \caption{Toy example.}
    \label{fig:toy-example-samples}
\end{wrapfigure}
We compare our engineered cost with a regularized Coulomb interaction, $C_{ij} = (1+\lVert \mathbf{x}_{i}^{(P)} - \mathbf{x}_{j}^{(P)} \rVert_{2})^{-1}$. This cost is shown alongside our proposed engineered cost in Figure~\ref{fig:toy-example-samples}. \emph{Note that, in Figures~\ref{fig:toy-example} (a - c) the lower right corner of the matrices corresponds to anomalies.}

\begin{figure}[ht]
    \centering
    \begin{subfigure}{0.95\linewidth}
        \centering
        \includegraphics[width=0.32\linewidth]{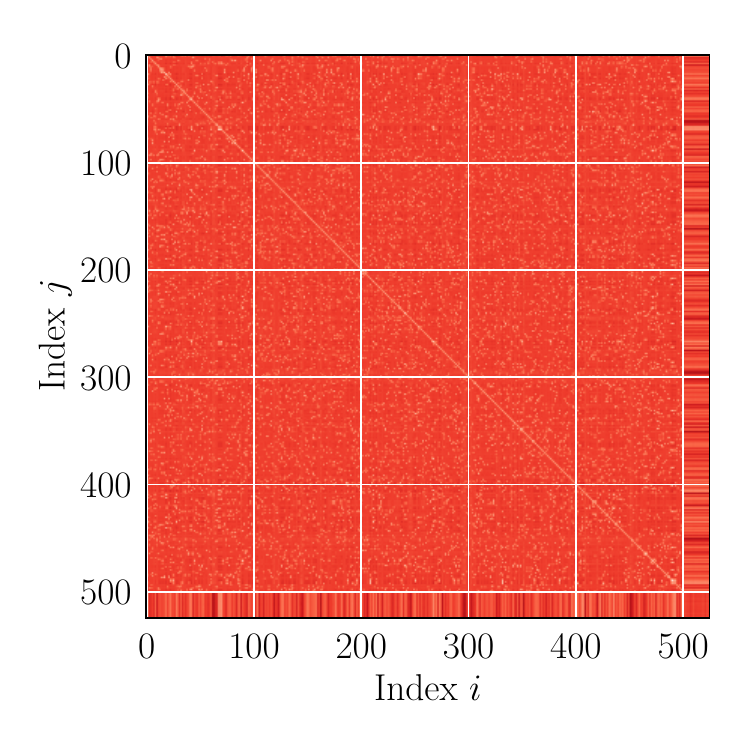}
        \includegraphics[width=0.32\linewidth]{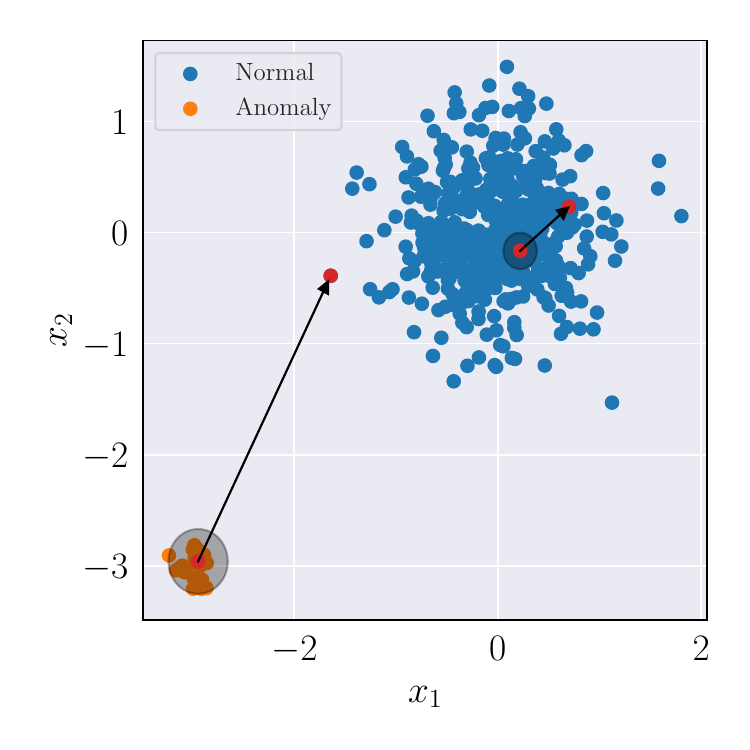}
        \includegraphics[width=0.32\linewidth]{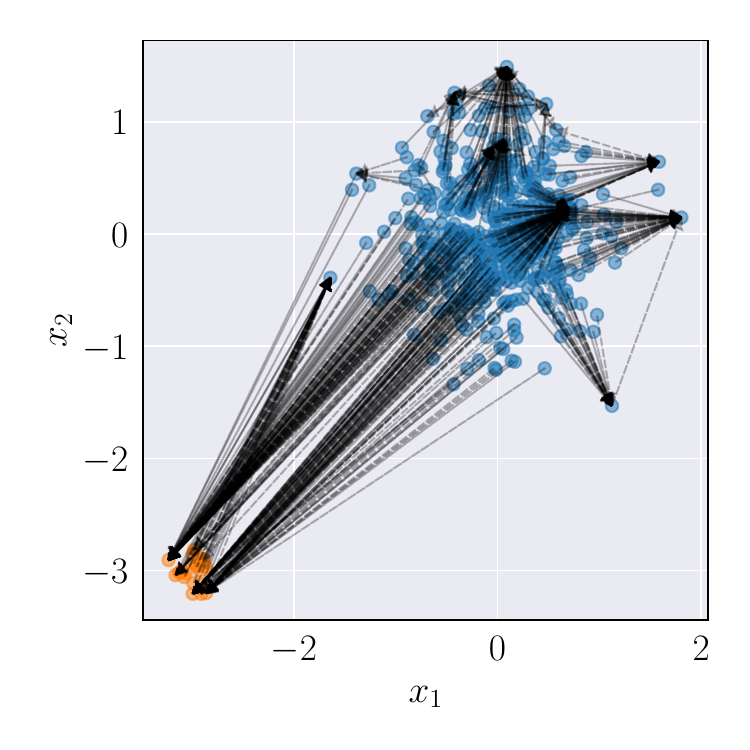}
        \caption{\gls{munot}.}
    \end{subfigure}\\
    \begin{subfigure}{0.95\linewidth}
        \centering
        \includegraphics[width=0.32\linewidth]{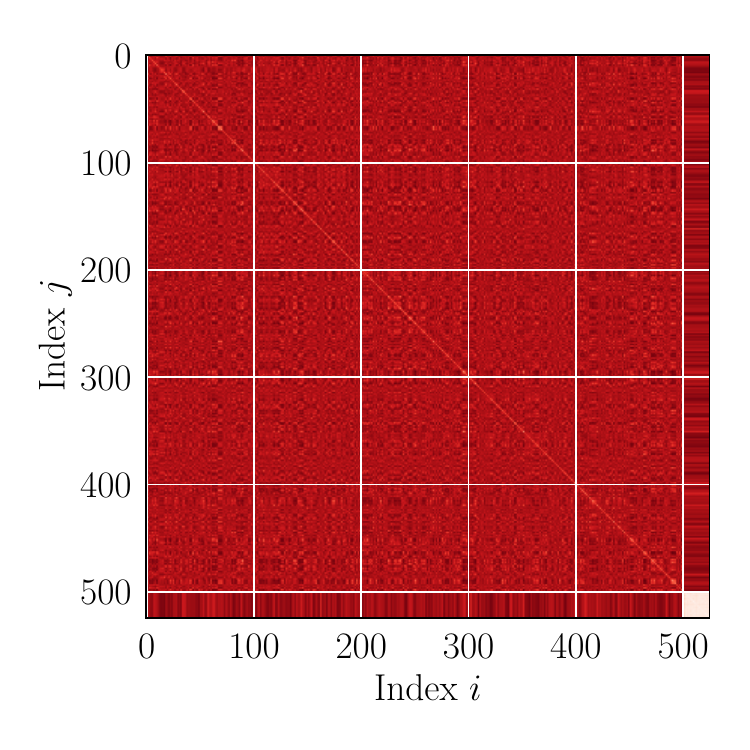}
        \includegraphics[width=0.32\linewidth]{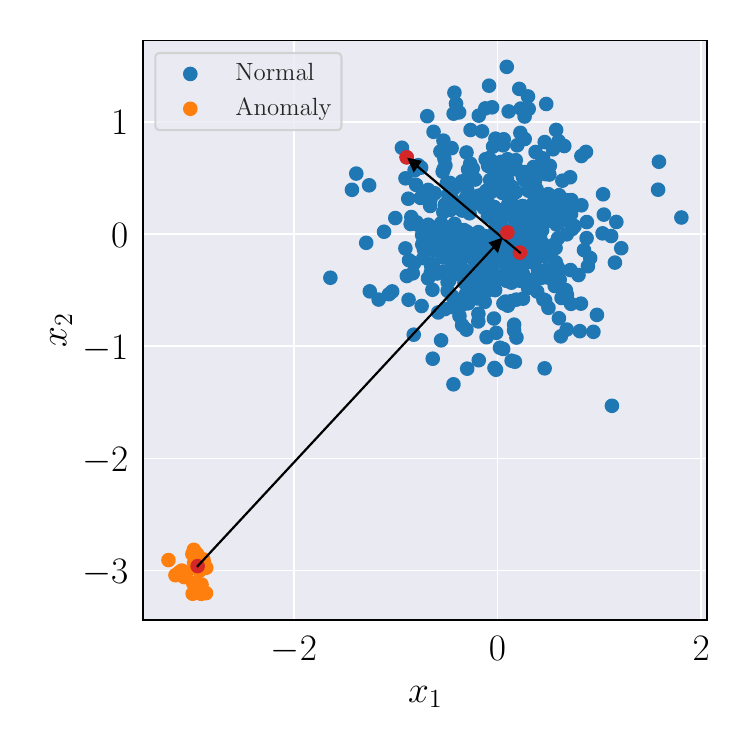}
        \includegraphics[width=0.32\linewidth]{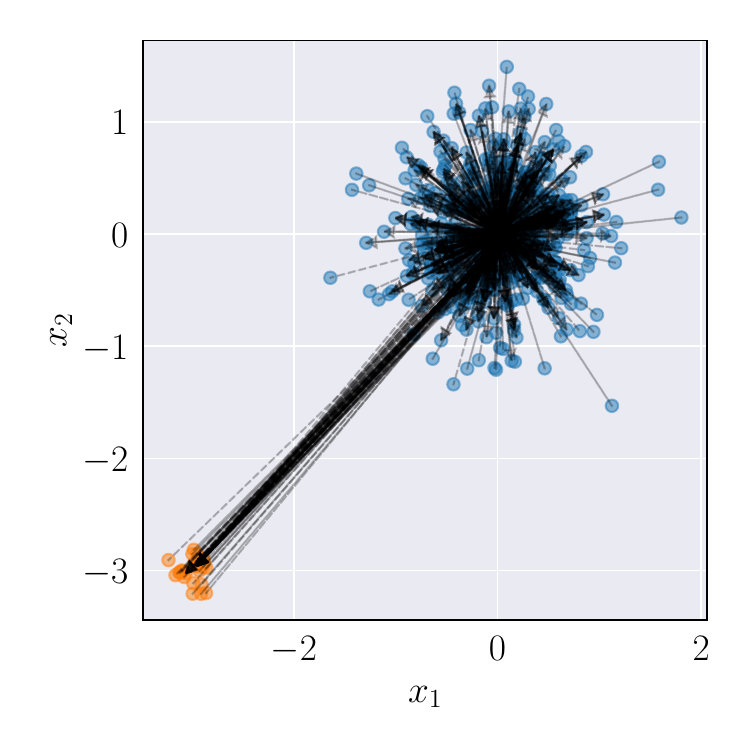}
        \caption{\gls{ot} with the regularized Coulomb cost.}
    \end{subfigure}\hfill
    \caption{In (a), we show the \gls{munot} transportation plan (left side), the matching (center) and the transport plan (right side). Likewise, in (b) we show the transportation plan and strategy for the regularized Coulomb cost. As shown in (b) left side and right side, the anomalies (lower right corner) sends and receives much less mass than other normal samples.}
    \label{fig:toy-example-transport-plans}
\end{figure}

From Figure~\ref{fig:toy-example}, \gls{ot} with the Euclidean cost results in a trivial \gls{ot} solution, because the diagonal entries $C_{ij}$ are zero. Hence, $\gamma^{\star} = \mathbf{I}_{n}$ achieves $0$ transportation cost. This is not the case for our engineered cost, and the regularized Coulomb interaction, shown in Figures~\ref{fig:toy-example} (c) and (d). To emphasize this idea, we show in Figure~\ref{fig:toy-example-transport-plans} the transportation plans acquired by \gls{munot}, and the regularized Coulomb cost. For the Coulomb cost, the anomalies send and receive less mass than normal samples. This is somewhat expected, as the anomalous samples are clustered together in a tight region of $\mathbb{R}^{2}$, resulting in a high cost. As a result, these samples are encouraged to send their mass elsewhere. This phenomenon does not happen with our engineered cost, which encourages samples to send their mass \emph{just outside} to their exclusion zone.

\begin{figure}[ht]
    \centering
    \begin{subfigure}{0.33\linewidth}
        \centering
        \includegraphics[width=\linewidth]{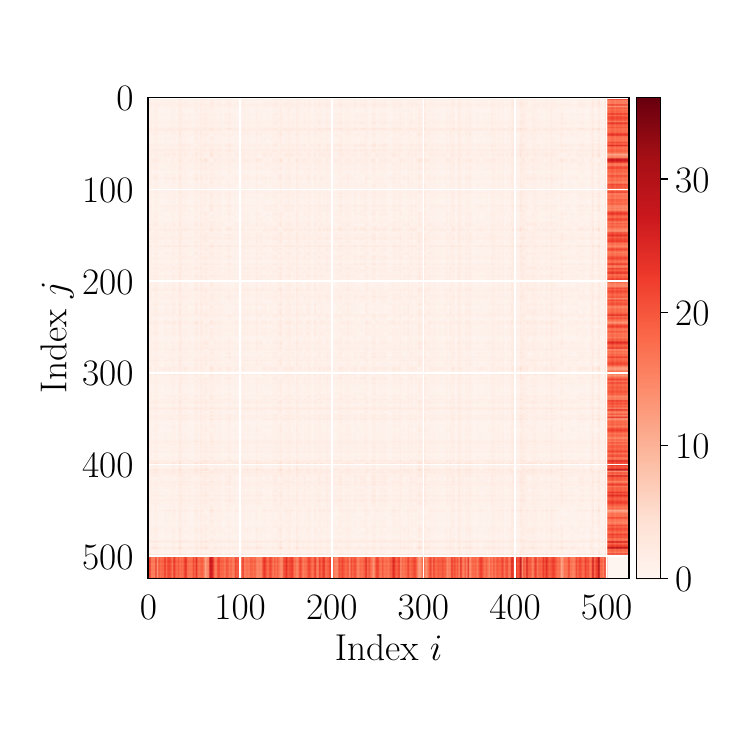}
        \caption{Euclidean cost.}
    \end{subfigure}\hfill
    \begin{subfigure}{0.33\linewidth}
        \centering
        \includegraphics[width=\linewidth]{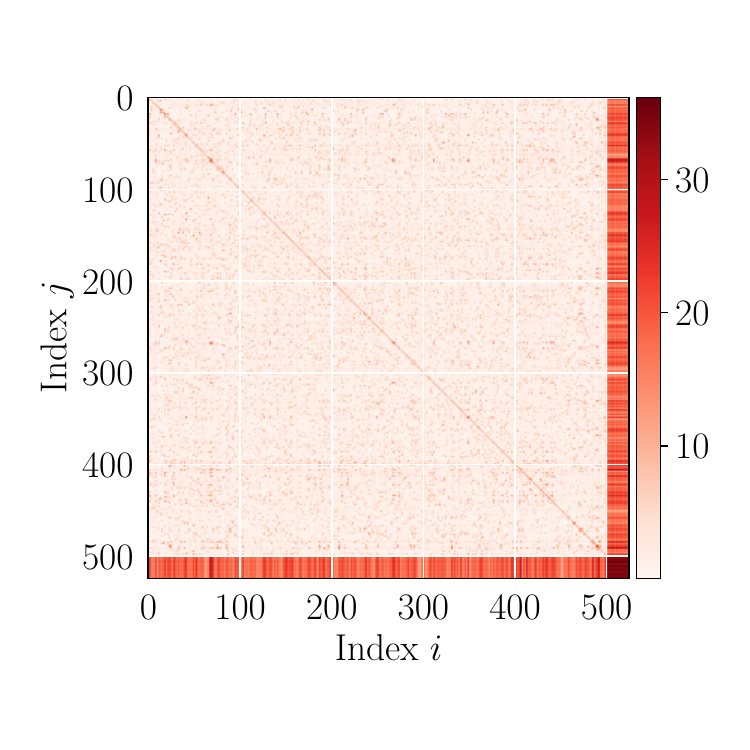}
        \caption{Engineered cost.}
    \end{subfigure}\hfill
    \begin{subfigure}{0.33\linewidth}
        \centering
        \includegraphics[width=\linewidth]{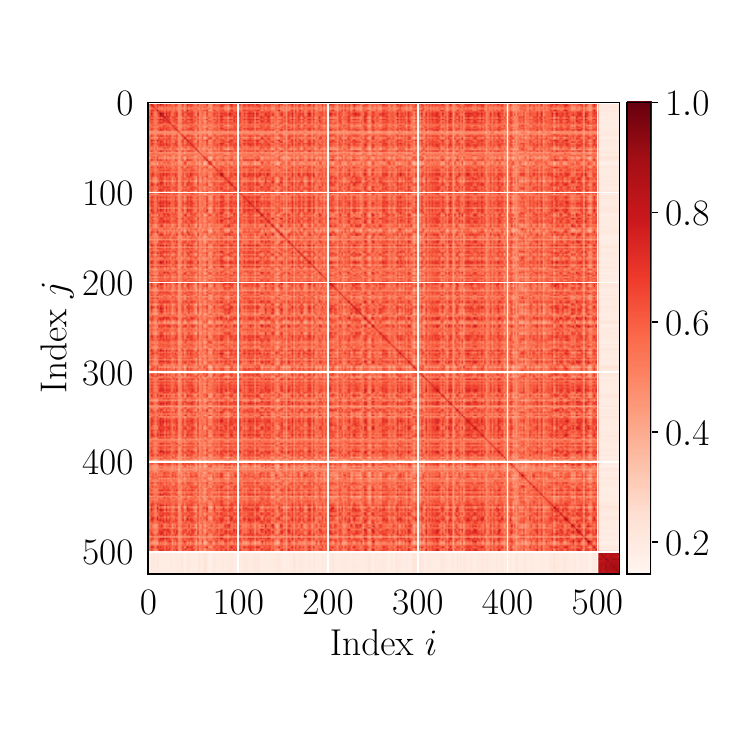}
        \caption{Coulomb cost.}
    \end{subfigure}\hfill
    \caption{In (a), (b) and (c), we show the ground-cost $C_{ij}$ between samples in Figure~\ref{fig:toy-example-samples}, using the Euclidean distance, our engineered ground-cost (equation~\ref{eq:engineered_cost}), and the regularized Coulomb interaction cost.}
    \label{fig:toy-example}
\end{figure}

Hence, there is an important distinction between our approach and using repulsive costs in \gls{ot}. Considering our last remark about the regularized Coulomb interaction, samples send their mass to distant regions of space, thus reducing $C_{ij}$. As a result, \emph{anomalies will incur in a smaller transportation cost than normal samples}. Although counterintuitive, one can still construct a detection rule out of this idea.

\begin{figure}[ht]
    \centering
    \begin{subfigure}{0.49\linewidth}
        \centering
        \includegraphics[width=0.49\linewidth]{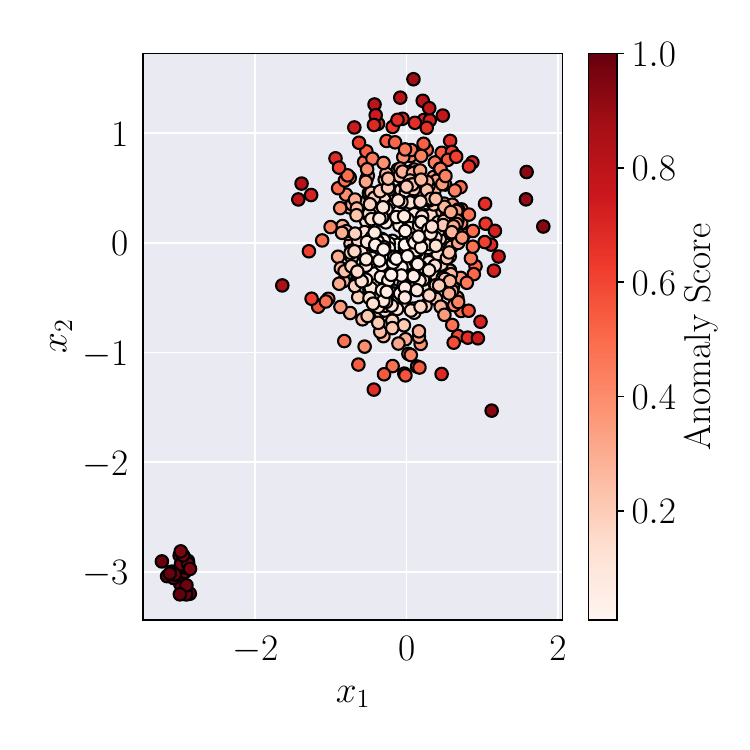}
        \includegraphics[width=0.49\linewidth]{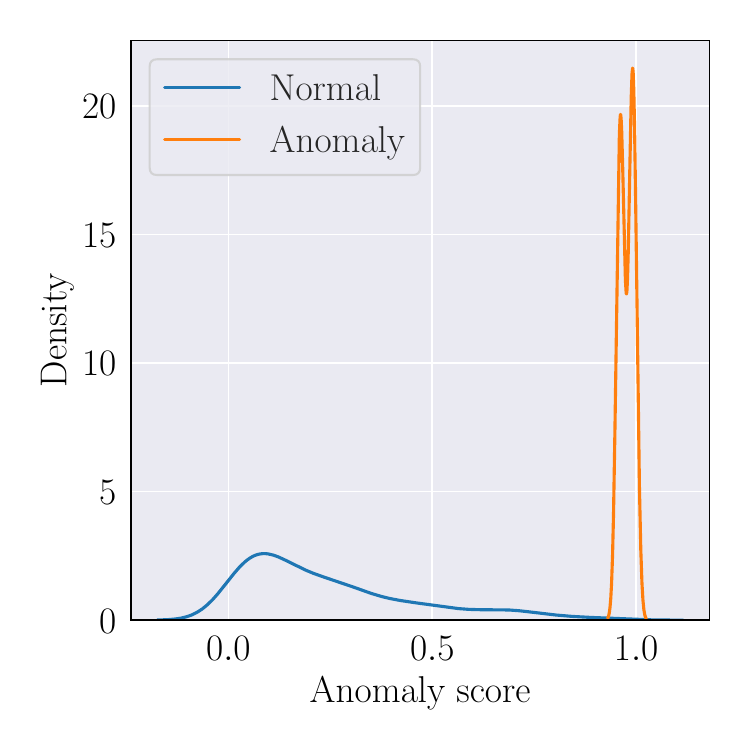}
        \caption{\gls{munot} anomaly score.}
    \end{subfigure}\hfill
    \begin{subfigure}{0.49\linewidth}
        \centering
        \includegraphics[width=0.49\linewidth]{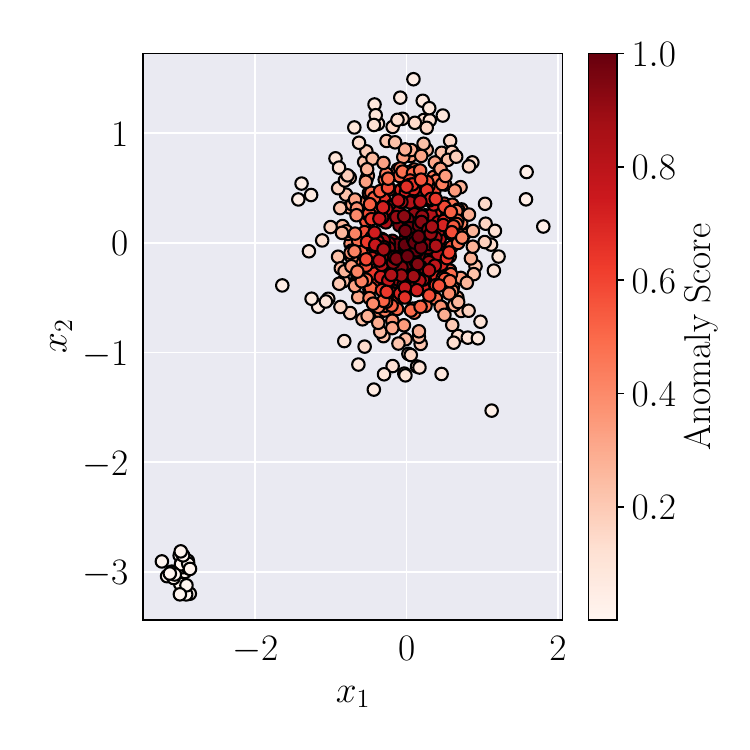}
        \includegraphics[width=0.49\linewidth]{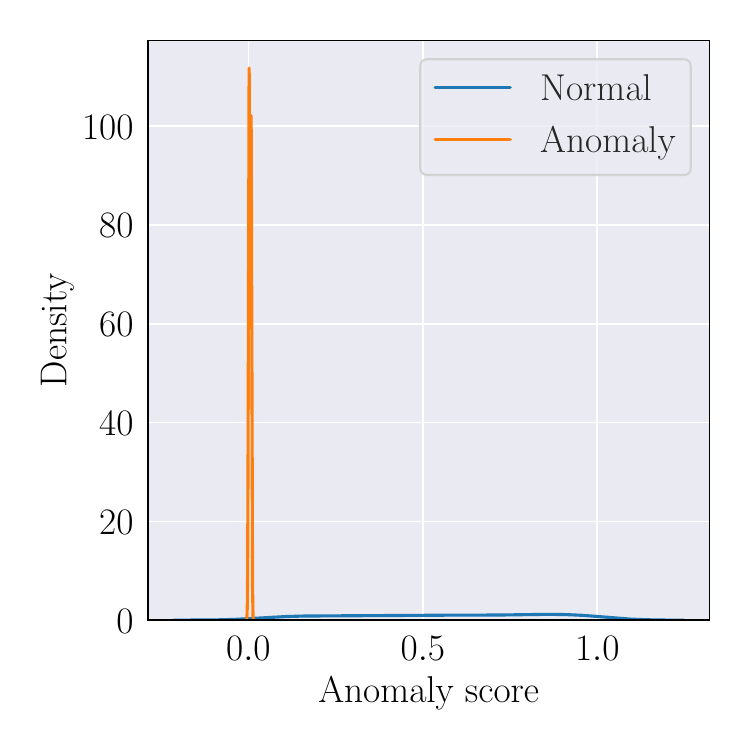}
        \caption{regularized Coulomb cost anomaly score.}
    \end{subfigure}\hfill
    \caption{Anomaly score comparison between the \gls{munot} (ours, a) and \gls{ot} with repulsive costs (b).  Scores are computed by calculating the transportation effort (Eq.~\ref{eq:score}) for each sample \(x_{i}^{(P)}\), estimating their density, and normalizing to the \([0, 1]\) range via the \gls{cdf} (c.f., section~\ref{sec:building-anomaly-score}). Overall, our strategy assigns higher scores (close to $1.0$) to anomalous samples, while assigning smaller scores to normal samples.}
    \label{fig:toy-example-transport-costs}
\end{figure}

In Figure~\ref{fig:toy-example-transport-costs}, we present the anomaly scores derived from our \gls{munot} strategy and the regularized Coulomb cost. These scores are computed by first determining the transportation effort (c.f., equation~\ref{eq:score}) for each sample $x_{i}^{(P)}$, then estimating the density of these efforts, and finally transforming the density into a $[0, 1]$ score via its \gls{cdf}, as outlined in section~\ref{sec:building-anomaly-score}. As shown in Figure~\ref{fig:toy-example-transport-costs}, the \gls{munot} approach assigns higher anomaly scores to anomalous samples compared to normal ones, simplifying the process of setting a threshold for anomaly detection. In contrast, the regularized Coulomb cost exhibits the opposite behavior. Indeed, anomalous samples send their mass to distant parts of space (i.e., to normal samples), which, due the nature of the Coulomb cost, lead to a smaller transportation effort. Nevertheless, as discussed previously, it is still possible to establish a detection rule, albeit being counterintuitive.

\begin{figure}[ht]
    \centering
    \begin{subfigure}{0.48\linewidth}
        \centering
        \includegraphics[width=0.48\linewidth]{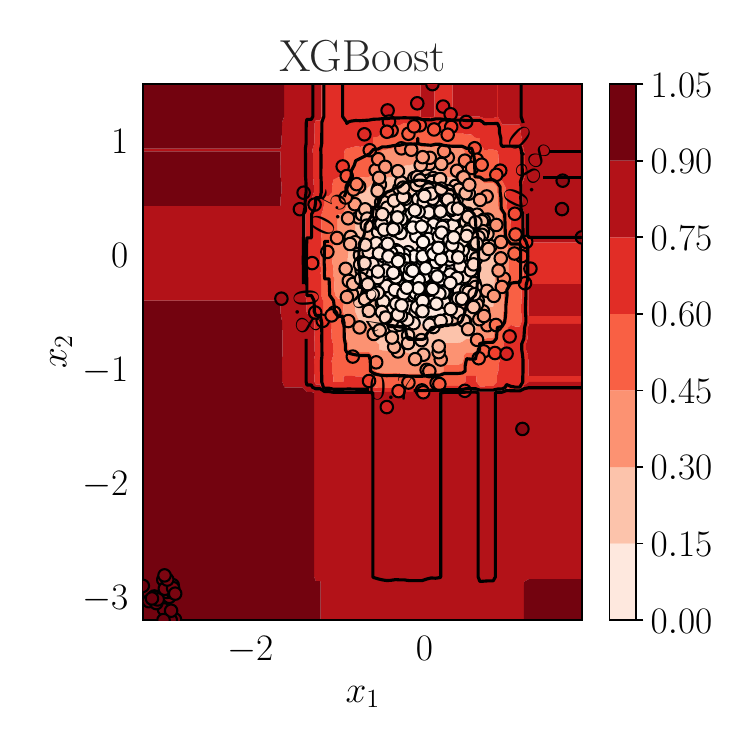}
        \includegraphics[width=0.48\linewidth]{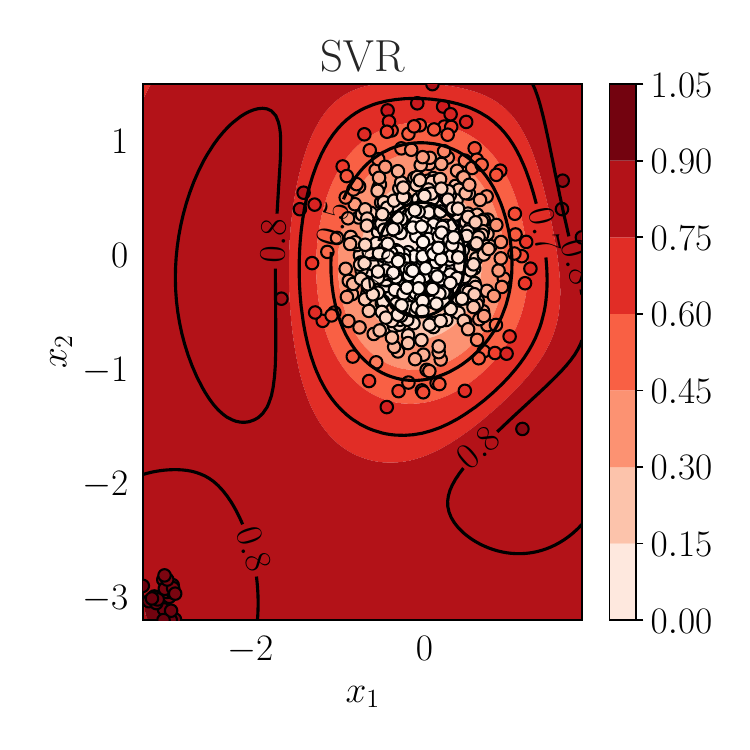}
        \caption{\gls{munot} regressed score.}
    \end{subfigure}\hfill
    \begin{subfigure}{0.48\linewidth}
        \centering
        \includegraphics[width=0.48\linewidth]{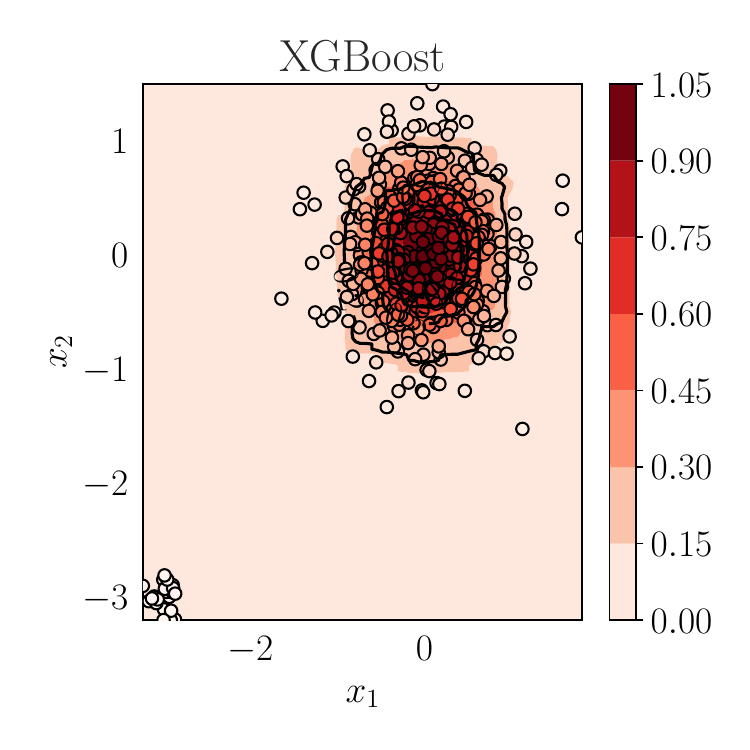}
        \includegraphics[width=0.48\linewidth]{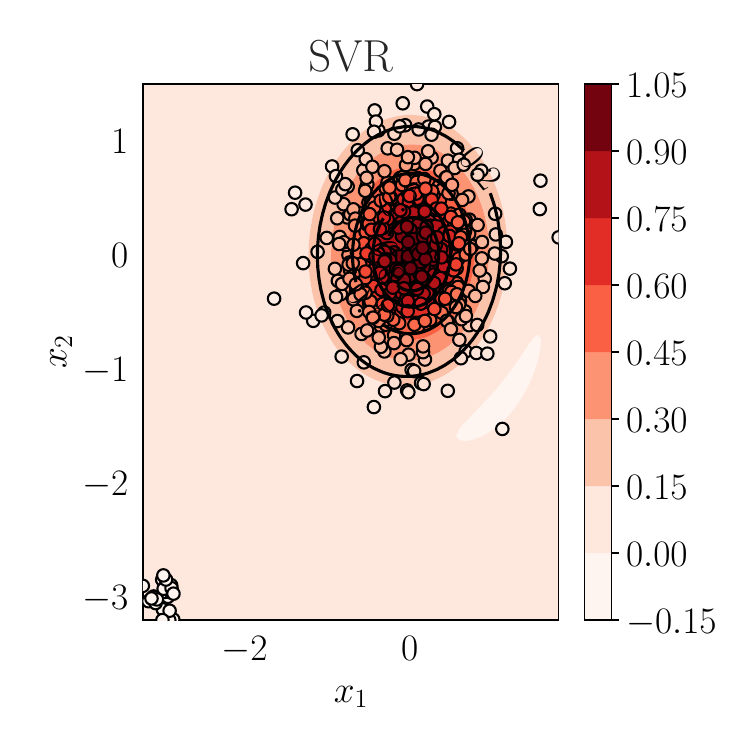}
        \caption{regularized Coulomb cost regressed score.}
    \end{subfigure}\hfill
    \caption{Regression of the anomaly score for \gls{munot} (ours, a) and \gls{ot} with repulsive costs (b). While our method attributes higher scores to anomalous regions of the space, \gls{ot} with repulsive costs does the inverse.}
    \label{fig:toy-example-regression-scores}
\end{figure}

As we discussed throughout section~\ref{sec:building-anomaly-score}, the scores in Figure~\ref{fig:toy-example-transport-costs} are only defined in the support of $\hat{P}$. We then explore the regression of this score through regression, which can be done with a variety of standard regression algorithms. In Figure~\ref{fig:toy-example-regression-scores} we show the results for \gls{xgboost}~\citep{chen2016xgboost} and \gls{svr}~\citep{smola2004tutorial}, which both define the anomaly score over the whole ambient space $\mathbb{R}^{2}$. As an important remark, the relationship between $\mathbf{x}_{i}^{(P)}$ and its score is likely non-linear. Indeed, this idea is evidenced in Figure~\ref{fig:toy-example-regression-scores}. As a result, one needs a non-linear regression algorithm.

\subsection{Comparison on AdBench}\label{ex:adbench}

AdBench~\citep{han2022adbench} is a benchmark in \gls{ad} with 57 different kinds of datasets, grouped into 47, 5 and 5 real-world, vision and \gls{nlp} datasets. In our experiments, we focus on real-world and natural-language datasets. We compare, in total, 17 methods, grouped into classical, diffusion-based and our proposed methods based on \gls{ot}. For classic methods, we consider \gls{isof}~\citep{liu2008isolation}, \gls{ocsvm}~\citep{scholkopf1999support}, \gls{knn}, \gls{lof}~\citep{breunig2000lof}, \gls{cblof}~\citep{he2003discovering}, \gls{ecod}~\citep{li2022ecod}, \gls{copod}~\citep{li2020copod}, \gls{loda}~\citep{pevny2016loda}, Feature Bagging~\citep{lazarevic2005feature}, and \gls{hbos}~\citep{goldstein2012histogram}. For diffusion based, we consider the 4 variants of \gls{dte}~\citep{livernoche2024on}. We also consider \gls{ot} with repulsive costs~\cite{di2017optimal}, and \gls{munot} (ours).

\begin{figure}[ht]
    \centering
    \begin{subfigure}{0.49\linewidth}
        \centering
        \includegraphics[width=\linewidth]{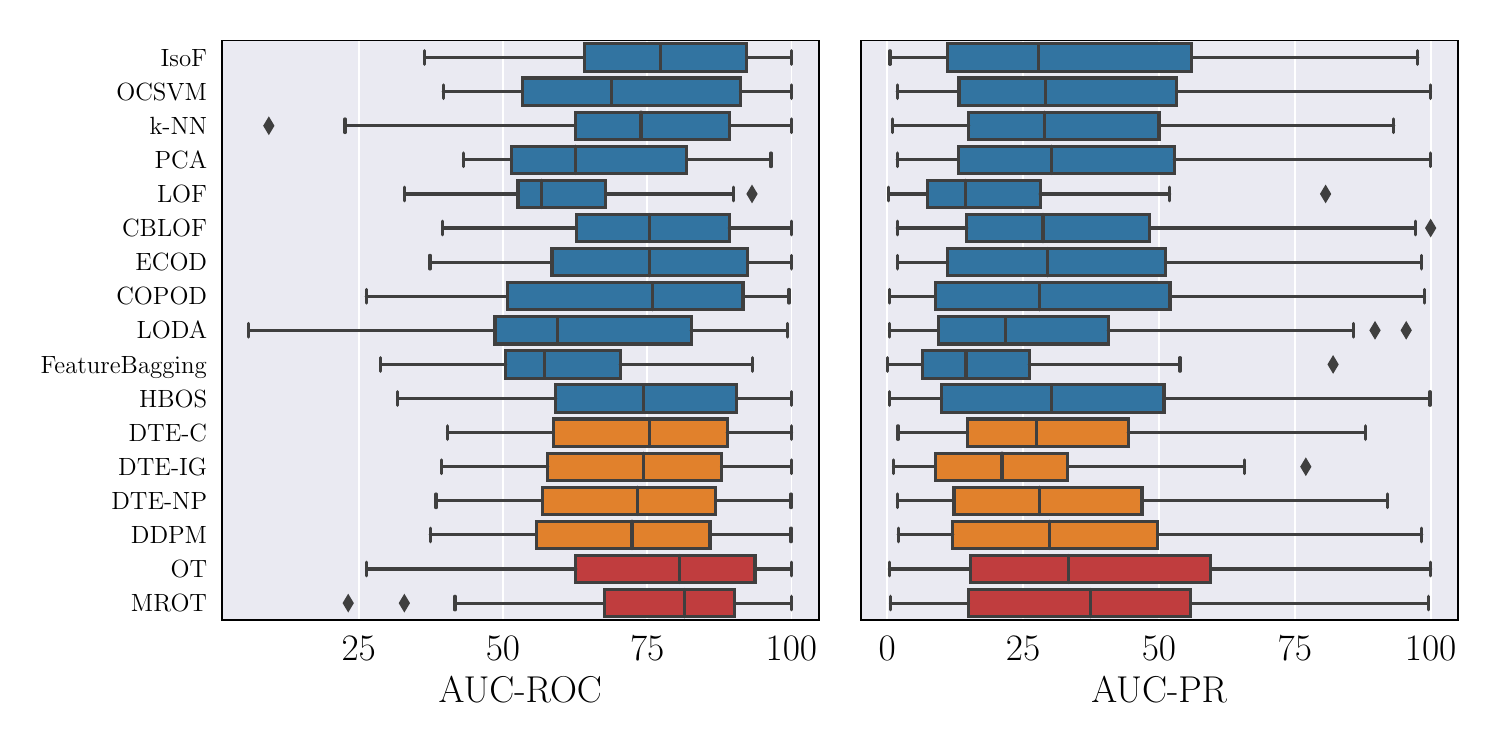}
        \caption{Real-world.}
    \end{subfigure}\hfill
    \begin{subfigure}{0.49\linewidth}
        \centering
        \includegraphics[width=\linewidth]{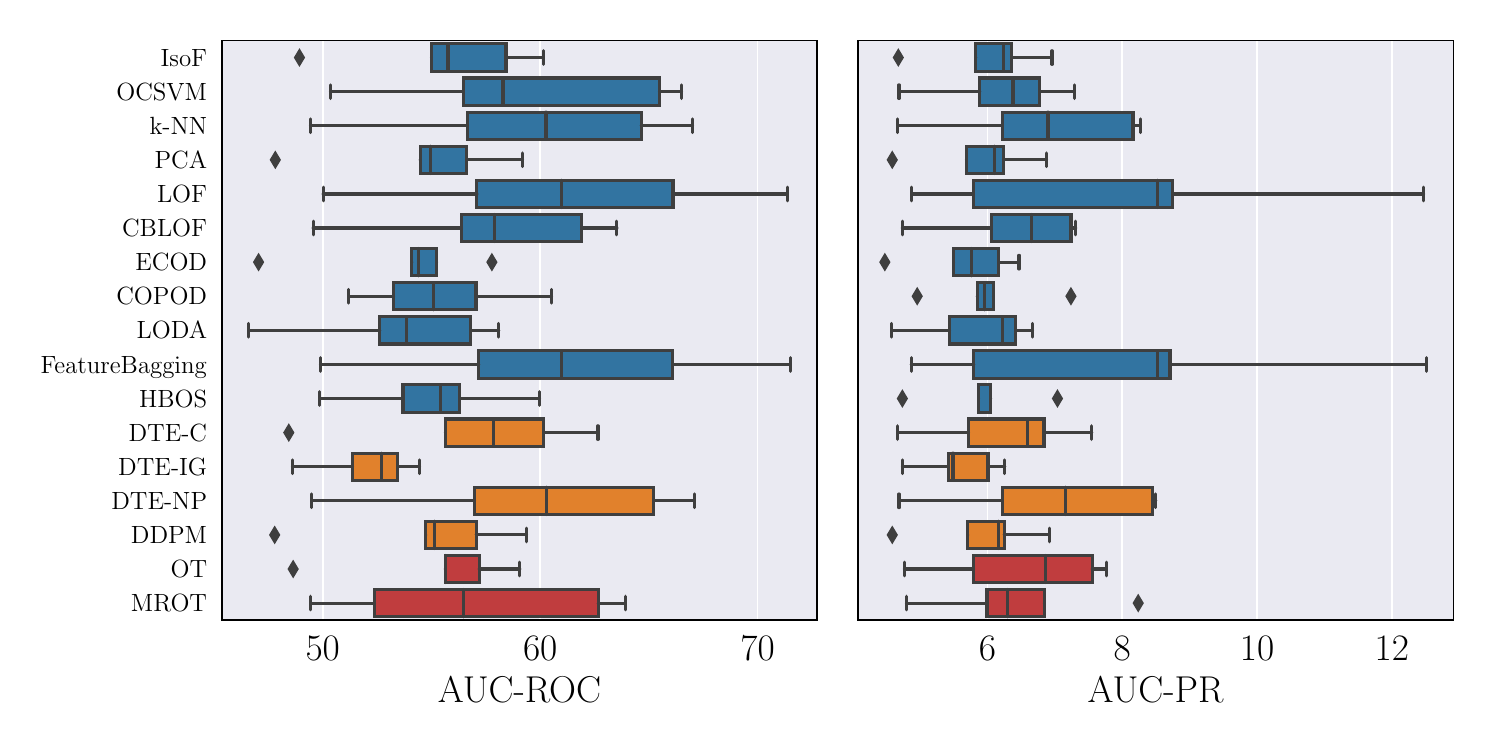}
        \caption{\gls{nlp}}
    \end{subfigure}
    \caption{AdBench result summary. AUC-ROC and AUC-PR per dataset is available in the appendix. We compare, in total, 17 algorithms over 47 real-world datasets (a) and 5 \gls{nlp} datasets (b). For real-world datasets, our \gls{munot} has state-of-the-art performance, especially when compared with recently proposed flow-based models. For higher dimensional data, our method}
    \label{fig:adbench-results}
\end{figure}

We show our summarized results in Figures~\ref{fig:adbench-results} (a) and (b), for real-world and NLP datasets, respectively. First, we note that \gls{munot} and \gls{ot} with repulsive costs have superior performance with respect other methods on real-world datasets. This result highlights the usefulness of \gls{ot} theory in the analysis of probability distributions. Furthermore, on average, our \gls{munot} has better performance than \gls{ot} (77.36\% versus 75.97\% ROC-AUC), proving the effectiveness of our cost engineering strategy. We refer readers to our appendix for detailed results per dataset.

Here, we highlight two limitations of our method. The first limitation of using \gls{munot} comes from the scalability of \gls{ot}. As a linear program, it has at least $\mathcal{O}(n^{2})$ storage, and $\mathcal{O}(n^{3}\log n)$ computational complexity, where $n$ is the number of samples. In our experiments, we limited the number of samples to $n = 20,000$, by down-sampling larger datasets. As reported in Figure~\ref{fig:adbench-results} (a), and our detailed results given in the Appendix, this process does not affect performance. Furthermore, from Figure~\ref{fig:adbench-results} (b), we see that our method struggles in high dimensional \gls{ad}, such as those in the \gls{nlp} datasets of~\cite{han2022adbench}. This limitation stems from the use of \gls{ot} in high-dimensions~\citep[Section 8.1]{montesuma2024recentadvancesoptimaltransport}.

\subsection{Tennessee Eastman Process}\label{ex:cdfd}

In this section, we focus in comparing anomaly detection methods for fault detection in control systems. In this context, as defined by~\cite{isermann2006fault}, a fault is an anomaly in at least one of the system's variables. To that end, we use the \gls{te} process benchmark~\citep{downs1993plant}, especially the simulations of~\cite{reinartz2021extended}, pre-processed by~\cite{montesuma2024benchmarking}.

This benchmark is composed by simulations of a large-scale chemical plant, from which a collection of 34 physical and chemical quantities are measured over time. There are a total of 29 states for this plant: 1 healthy state, and 28 faulty states. Furthermore, the plant can operate under 6 different modes of operation, depending on the specified production requirements. We refer readers to the aforementioned references, as well as our appendix, for detailed descriptions of the variables, faults and modes of operation.

\begin{figure}[ht]
    \centering
    \begin{subfigure}{0.31\linewidth}
        \centering
        \includegraphics[width=\linewidth]{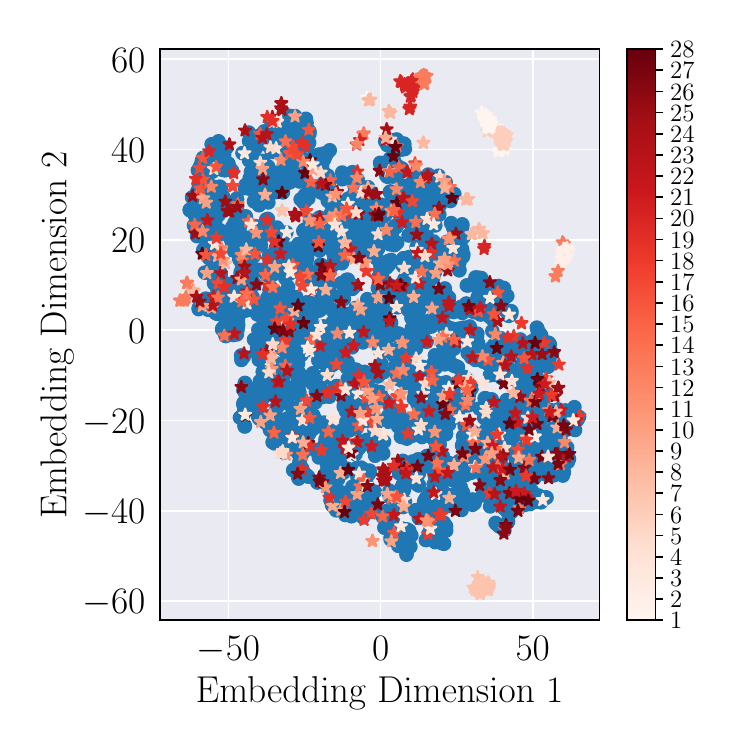}
        \caption{Mode 1.}
    \end{subfigure}\hfill
    \begin{subfigure}{0.31\linewidth}
        \centering
        \includegraphics[width=\linewidth]{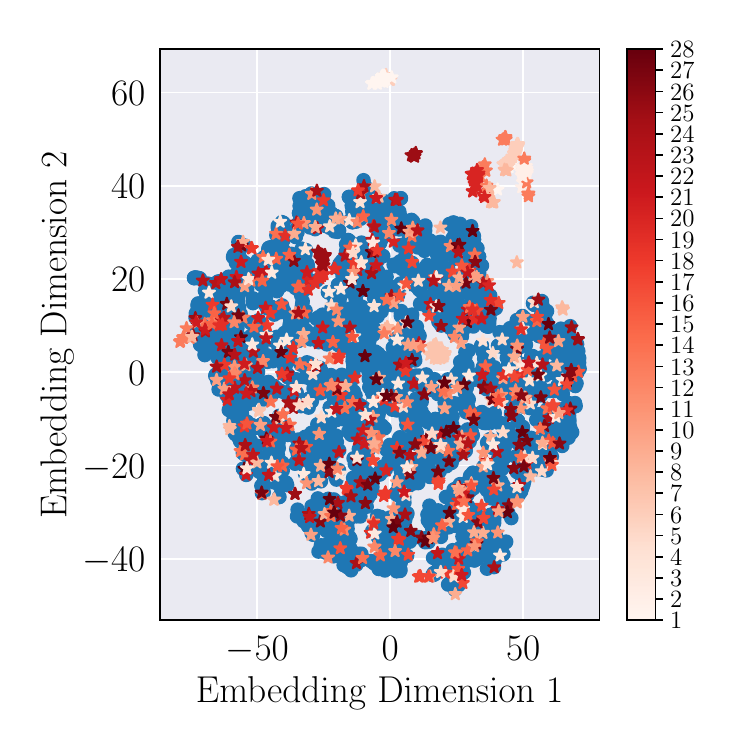}
        \caption{Mode 2.}
    \end{subfigure}\hfill
    \begin{subfigure}{0.31\linewidth}
        \centering
        \includegraphics[width=\linewidth]{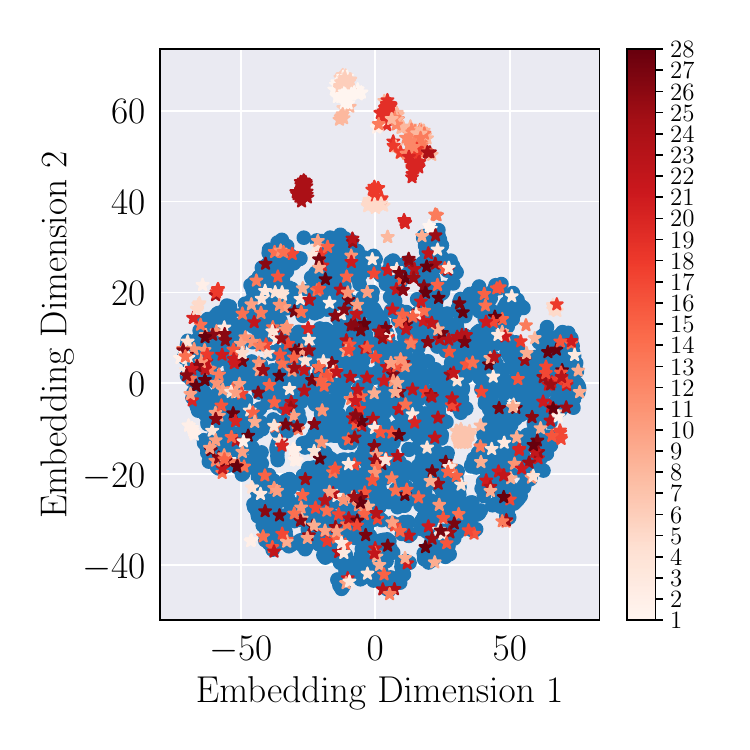}
        \caption{Mode 3.}
    \end{subfigure}\hfill
    \begin{subfigure}{0.31\linewidth}
        \centering
        \includegraphics[width=\linewidth]{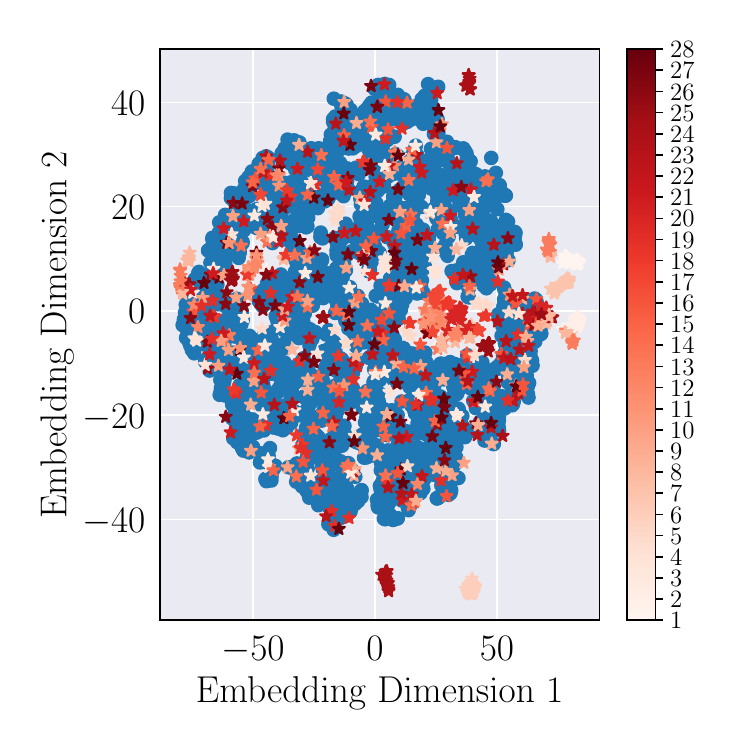}
        \caption{Mode 4.}
    \end{subfigure}\hfill
    \begin{subfigure}{0.31\linewidth}
        \centering
        \includegraphics[width=\linewidth]{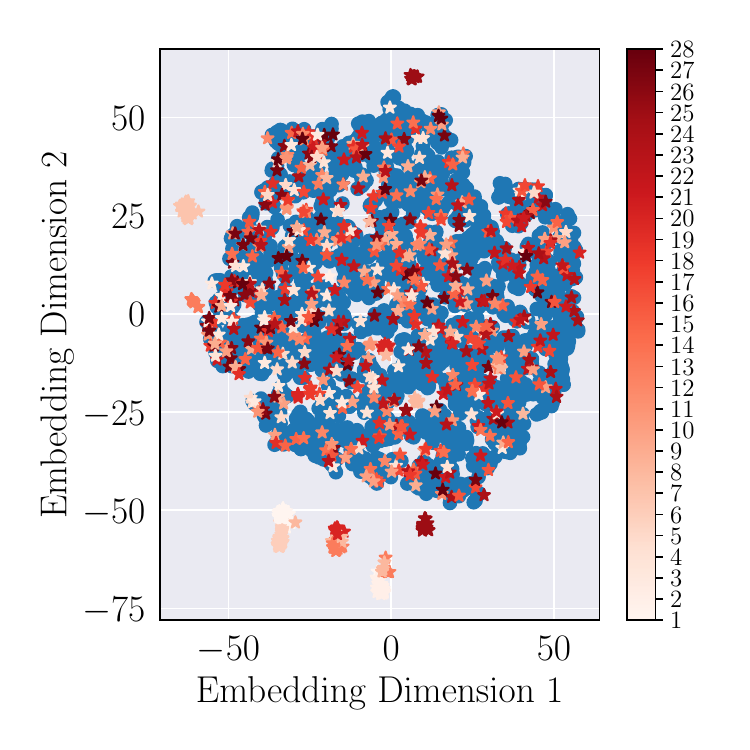}
        \caption{Mode 5.}
    \end{subfigure}\hfill
    \begin{subfigure}{0.31\linewidth}
        \centering
        \includegraphics[width=\linewidth]{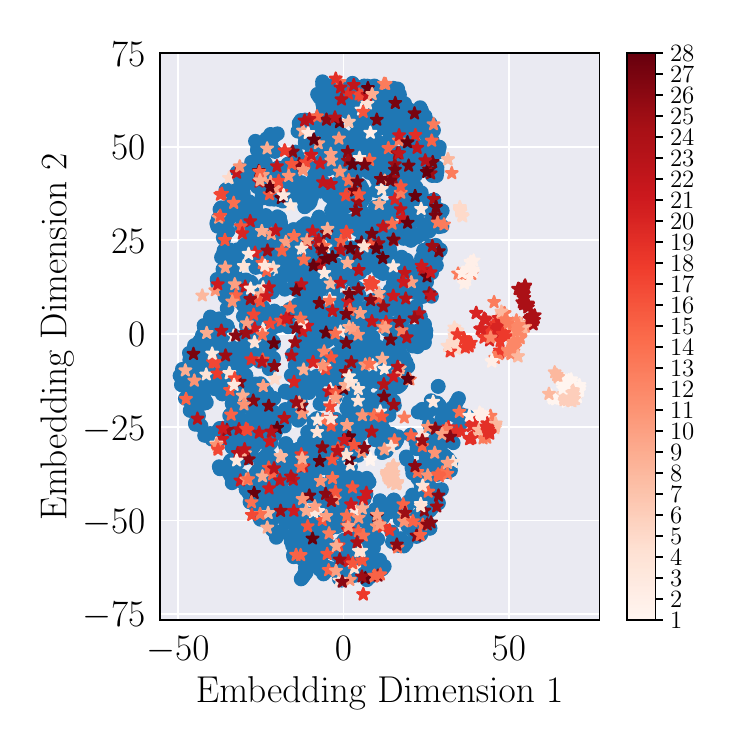}
        \caption{Mode 6.}
    \end{subfigure}\hfill
    \caption{t-SNE embeddings of the Tennessee Eastman process data per mode. In blue, we show the normal samples, whereas we show anomalous samples in shades of reds, corresponding to the actual fault category they correspond. While most anomalous samples do cluster in a region outside the non-faulty cluster, some faulty samples do not.}
    \label{fig:tep-results}
\end{figure}

From the original simulations in~\cite{reinartz2021extended}, we consider $100$ simulations for the normal state. These simulations last for $600$ hours, with a time-step of $1$ hour, and concern $34$ sensors measuring different physical and chemical properties. On top of these $100$ simulations, we take $1$ faulty simulation for each kind of fault, leading to a total of $128$ simulations. Based on this set of simulations, we extract windows of $20$ hours from the original signals. Each window is then considered a sample for anomaly detection. On each window, we compute the mean, and standard deviation of each variable, leading to vectors $\mu, \sigma \in \mathbb{R}^{34}$. We use the concatenation of these vectors as features for anomaly detection. These steps lead to $6$ datasets with $3840$ samples, $840$ ($21.875\%$) of them being anomalous, one for each mode of operation.

\begin{figure}[ht]
    \centering
    \begin{subfigure}{0.31\linewidth}
        \centering
        \includegraphics[width=\linewidth]{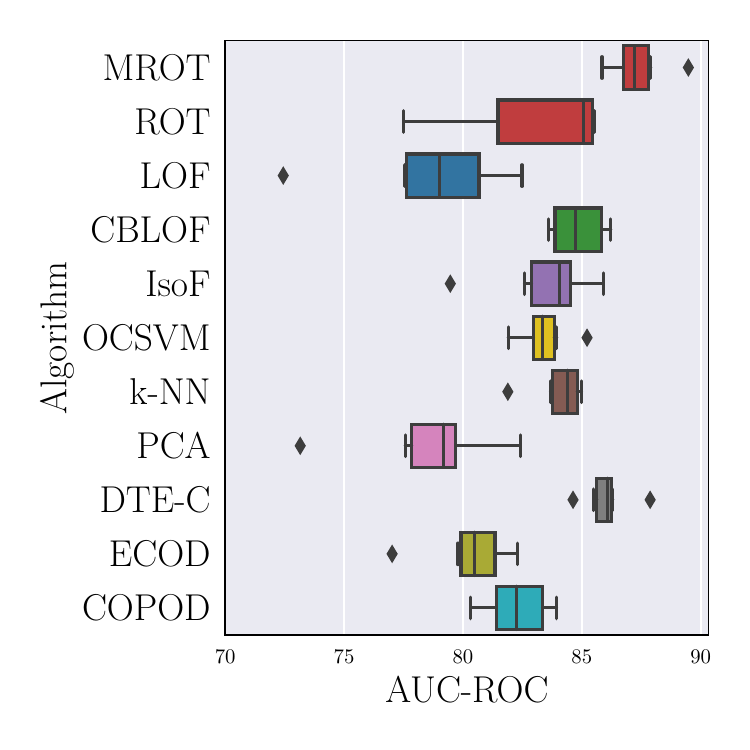}
        \caption{Mode 1.}
    \end{subfigure}\hfill
    \begin{subfigure}{0.31\linewidth}
        \centering
        \includegraphics[width=\linewidth]{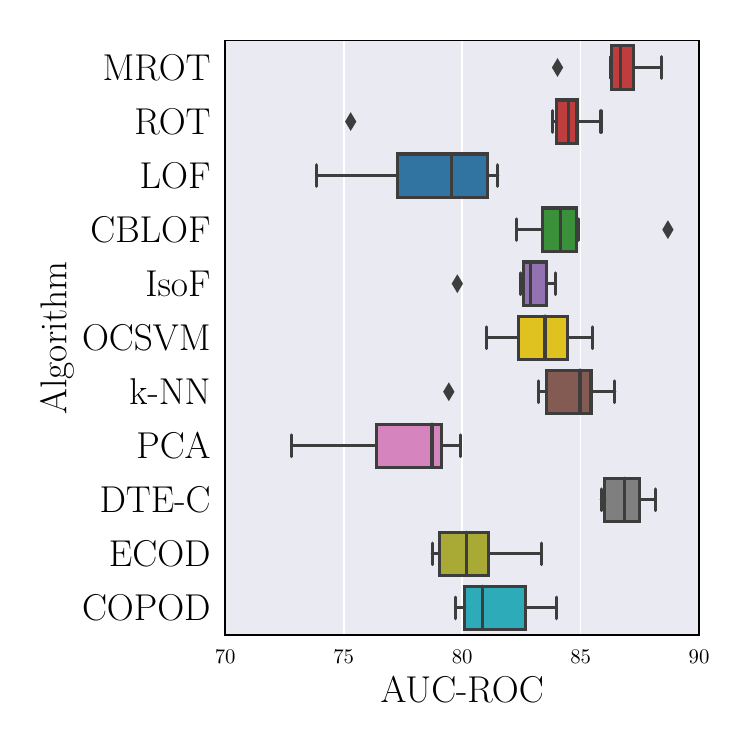}
        \caption{Mode 2.}
    \end{subfigure}\hfill
    \begin{subfigure}{0.31\linewidth}
        \centering
        \includegraphics[width=\linewidth]{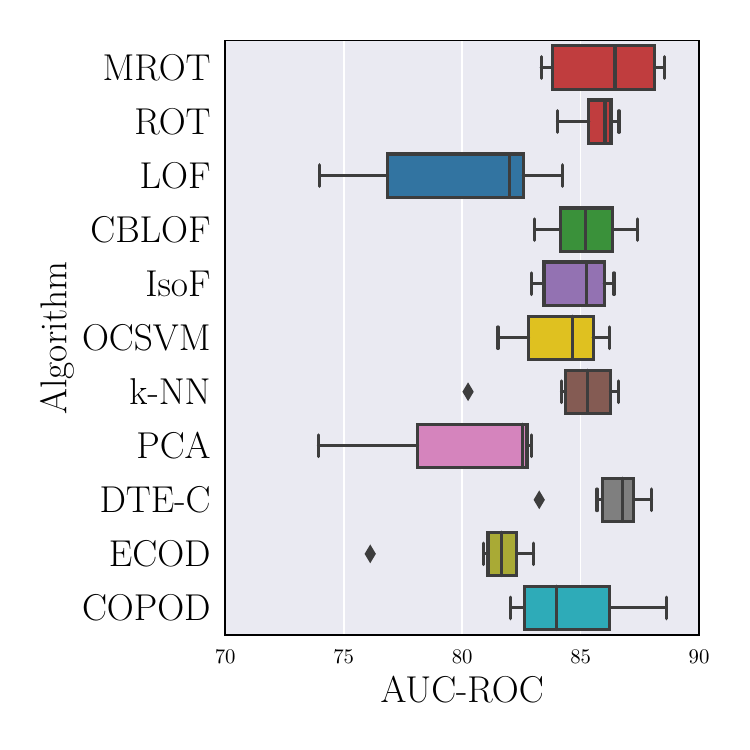}
        \caption{Mode 3.}
    \end{subfigure}\hfill
    \begin{subfigure}{0.31\linewidth}
        \centering
        \includegraphics[width=\linewidth]{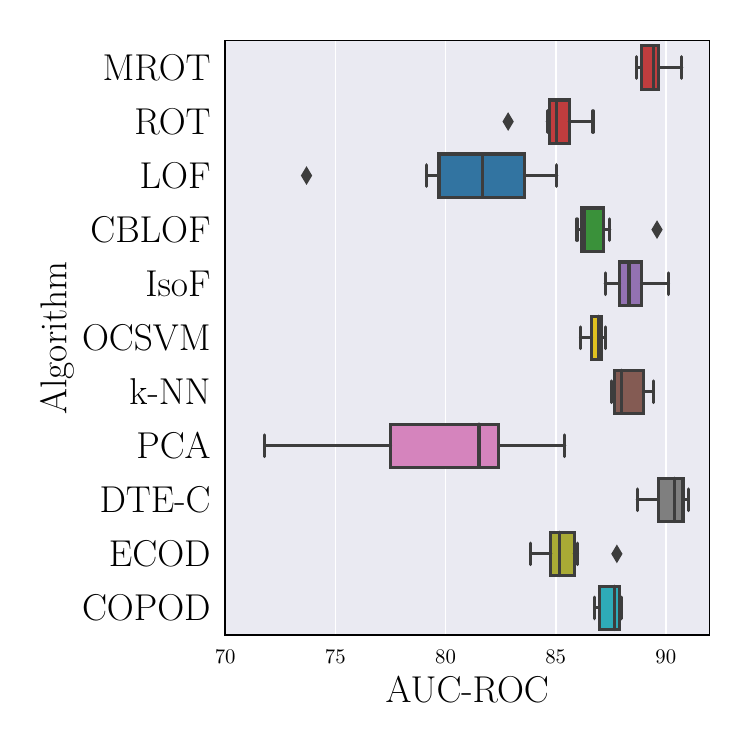}
        \caption{Mode 4.}
    \end{subfigure}\hfill
    \begin{subfigure}{0.31\linewidth}
        \centering
        \includegraphics[width=\linewidth]{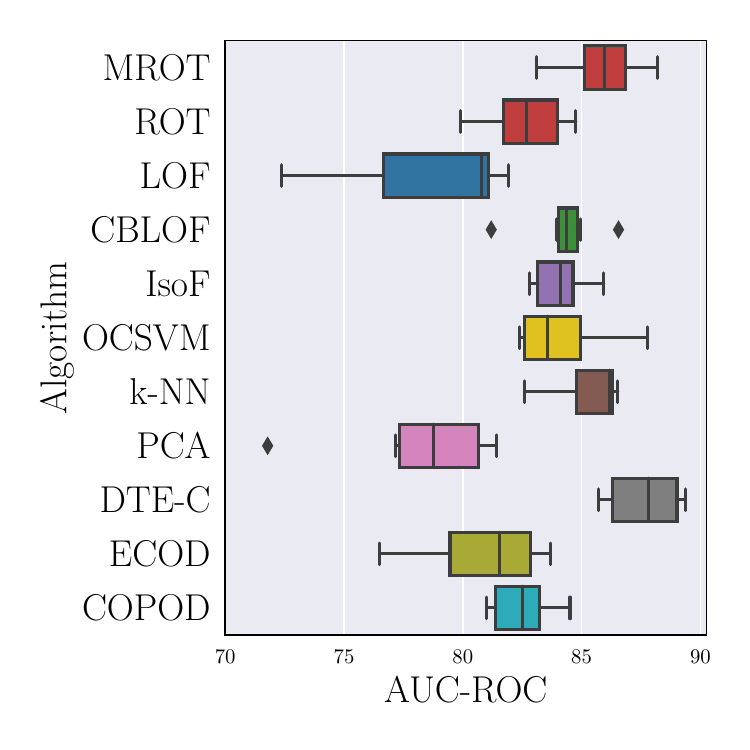}
        \caption{Mode 5.}
    \end{subfigure}\hfill
    \begin{subfigure}{0.31\linewidth}
        \centering
        \includegraphics[width=\linewidth]{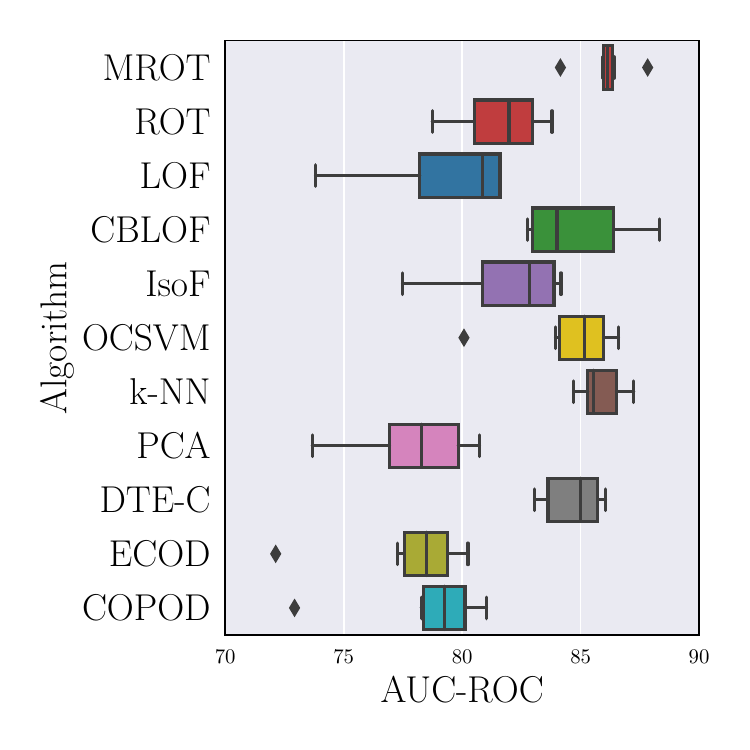}
        \caption{Mode 6.}
    \end{subfigure}\hfill
    \caption{Aggregated anomaly detection on the Tennessee Eastman data per mode of operation. First, MROT outperforms OT with repulsive costs over all modes. Second, MROT and DTE-C have state-of-the-art performance, superior to previously proposed methods.}
    \label{fig:tep-aggregated-results}
\end{figure}

We start our analysis by embedding the obtained features in $\mathbb{R}^{2}$ through \gls{tsne}~\citep{van2008visualizing}, which is shown in Figure~\ref{fig:tep-results}. We scatter the embeddings of each mode's data, showing that most faulty samples cluster outside the normal data cluster. However, some faulty samples are close to normal ones. This phenomenon is expected, as the effect of faults evolves over time. As a result, windows taken in early stages of simulation resemble those of normal samples.

Next, we benchmark the \gls{ad} performance of algorithms, and their capability to generalize to unseen that within the same mode of operation. In this experiment, we downsample the number of anomalous samples per fault category to $\{5, 10, \cdots, 30\}$. This results in a percentage of $\{4.45\%, 8.53\%, 12.28\%, 15.73\%, 18.92\%, 21.87\%\}$ of anomalous samples. In Figure~\ref{fig:tep-aggregated-results}, we report our aggregated results over all percentage of anomalies. We refer readers to our appendix for results per percentage.

The main idea of this methodology is evaluating how different algorithms perform, under a variable percentage of faults. Our results are shown in Figure~\ref{fig:tep-results} (b). Among the tested methods, \gls{munot} has a better performance and remains stable throughout the range of percentage of anomalies.

\subsection{Ablations}\label{sec:ablations}

\begin{figure}[ht]
    \centering
    \begin{subfigure}{0.8\linewidth}
        \centering
        \includegraphics[width=\linewidth]{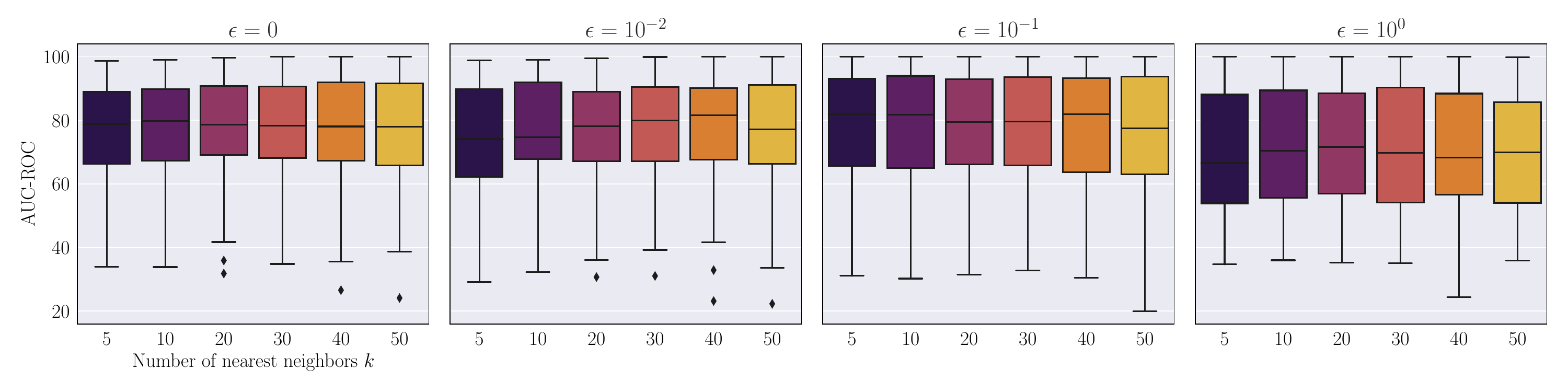}
        \caption{MROT.}
    \end{subfigure}
    \begin{subfigure}{0.18\linewidth}
        \centering
        \includegraphics[width=\linewidth]{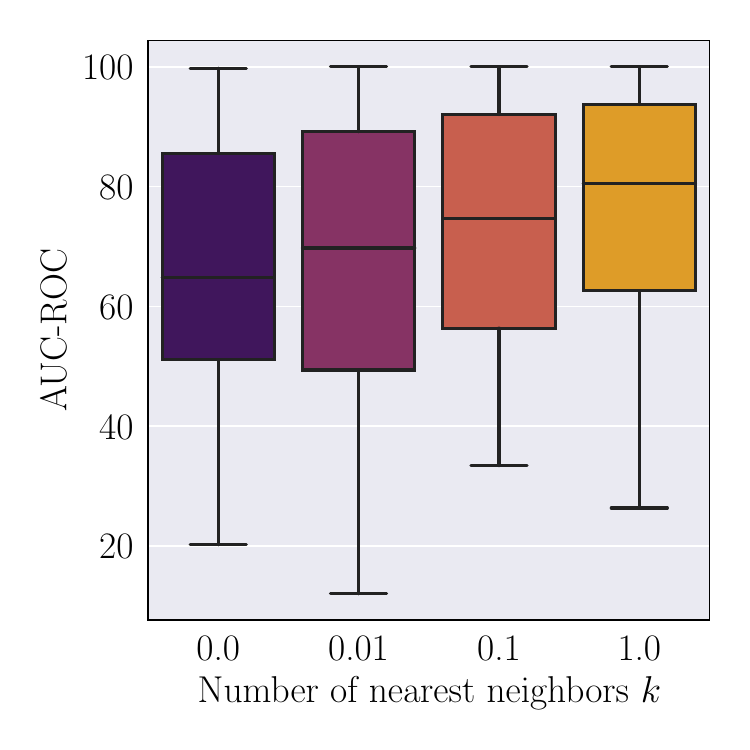}
        \caption{ROT.}
    \end{subfigure}
    \caption{Hyper-parameter sensitivity. In (a), we show the performance of \gls{munot} for a variable number of nearest neighbors $k$, and entropic regularization $\epsilon$. Overall, our method is robust to the choices of these hyper-parameters, but using a lower $\epsilon$ is generally better. In contrast, we show in (b) the performance of \gls{ot}-based \gls{ad} with the regularized Coulomb cost, for which using a higher entropic penalty $\epsilon$ improves performance.}
    \label{fig:hyper-parameters-sensitivity}
\end{figure}

\textbf{Hyper-parameter robustness.} Here, we analyze the robustness of our method with respect to the entropic regularization penalty $\epsilon$, and the number of nearest neighbors $k$ in $\mathcal{N}_{k}$. In our experiments, we evaluated our method on the values $\epsilon \in \{0, 10^{-2}, 10^{-1}, 10^{0}\}$, where $\epsilon = 0$ implies the use of exact \gls{ot}, that is, linear programming. For \gls{munot}, we use $k\in\{5, 10, 20, \cdots, 50\}$. Note that while $\epsilon$ is related to \emph{transportation plan}, $k$ is linked to the \emph{ground-cost}. We summarize our results in Figure~\ref{fig:hyper-parameters-sensitivity}, where we present box-plots over the AUC-ROC scores on each dataset in the AdBench benchmark, for each combination of hyper-parameters.

From Figure~\ref{fig:hyper-parameters-sensitivity}, we note that both \gls{munot} is robust to the choice of entropic regularization and number of nearest neighbors. As a general guideline, it is better to limit the number of nearest neighbors, as anomalous examples are likely rare. As a consequence, using a high value for $k$ may lead to the inclusion of normal points in the neighborhood of anomalous ones.

We give an additional reasoning on the choice of $\epsilon$. From the classical \gls{ot}~\citep[Chapter 4]{peyre2020computationaloptimaltransport}, using $\epsilon > 0$ leads to a nonsparse plan. This means that there are more indices $(i, j)$, for which $\gamma_{ij} > 0$. Likewise, if we let $\epsilon \rightarrow +\infty$, then $\gamma$ converges to the trivial coupling $\gamma_{ij} = p_{i}p_{j}$. Assuming uniform importance, i.e., $p_{i} = n^{-1}$, and plugging these results back into equation 10, we have $t_{i} = n^{-1}\sum_{j=1}^{n}\tilde{c}_{k}(x_{i}^{(P)}, x_{j}^{(P)})$, that is, for a large entropic penalty, we end up simply computing the average cost of each sample. The implication of this analysis is as follows: the MROT plan captures some structure in the relationship between anomalous and normal points, which contributes to a more accurate anomaly score.

\begin{figure}[ht]
    \centering
    \includegraphics[width=\linewidth]{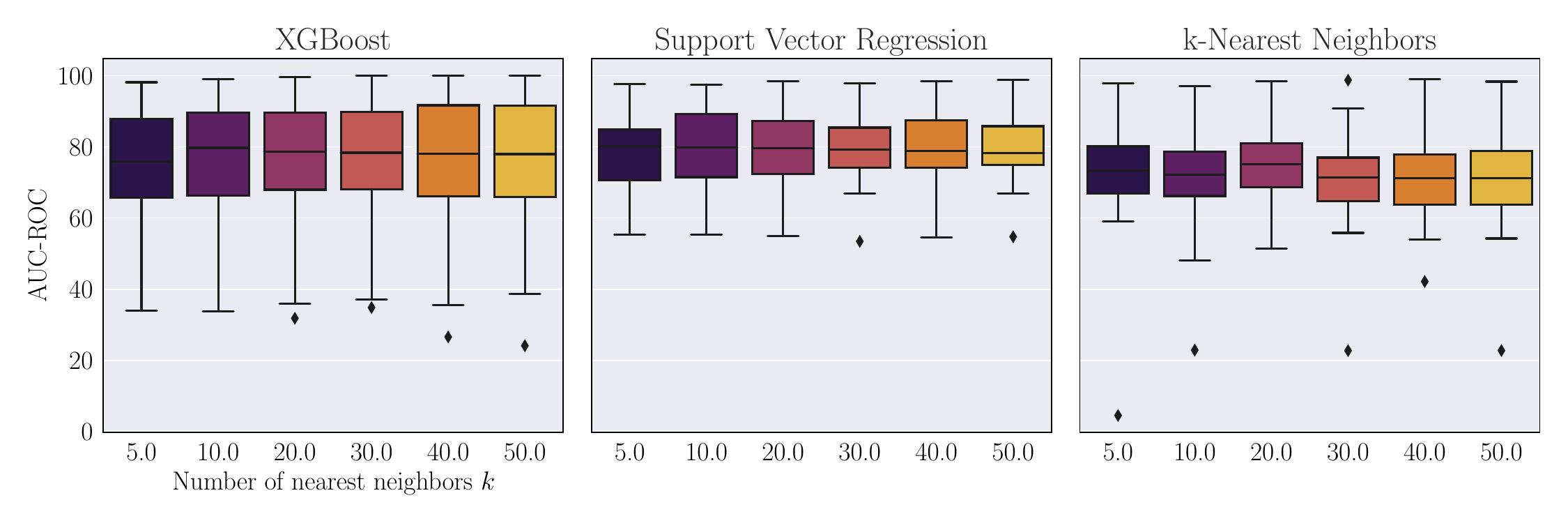}
    \caption{Ablation of \gls{munot} performance with respect regression model choice. Overall, gradient boosting and \gls{svr} have similar performance, whereas $k-$nearest neighbors has sub-optimal performance due to overfitting.}
    \label{fig:reg_sensitivity}
\end{figure}

\noindent\textbf{Robustness to regression algorithm.} As we mentioned in our toy example, it is necessary to use a nonlinear regression model for generalizing the anomaly score, as the relationship between $\mathbf{x}_{i}^{(P)}$ and $\hat{F}_{\mathcal{T}}(t_{i})$ is likely nonlinear. We ablate on the choice of regression algorithm, exploring the use of gradient boosting~\citep{friedman2002stochastic}, \gls{svr}~\citep{smola2004tutorial} and k-nearest neighbors regression. We show a summary of our results in Figure~\ref{fig:reg_sensitivity}. Note that, in general, the performance of boosting and \gls{svr} is similar and stable among different choices for $k$ in $\mathcal{N}_{k}$ in our engineered cost. The performance of $k-$nearest neighbors regression is sub-optimal, especially due to overfitting.

In addition to the pipeline shown in Algorithm~\ref{alg:mrot}, it is possible to add a model selection step, e.g., through k-fold cross-validation. The idea is to partition the training dataset $\{\mathbf{x}_{i}^{(P)},\hat{F}_{\mathcal{T}}(t_{i})\}_{i=1}^{n}$ into a training and a validation set. Then, among a pool of possible regression models, the one that has the minimum error in the validation set.

\section{Conclusion}\label{sec:conclusion}

In this paper, we introduced a novel, general purpose anomaly detection algorithm based on optimal transport theory. Our method works under the assumption that the local neighborhood of anomalous samples is likely more irregular than that of normal samples. Based on this idea, we engineer the ground-cost in optimal transport, to encourage samples to send their mass \emph{just outside} an exclusion zone, defined through its $k-$nearest neighbors. While this idea bears some similarity to optimal transport with repulsive costs~\citep{di2017optimal}, the defined ground-cost is not repulsive, and, as we show in our experiments, it leads to better anomaly detectors. We thoroughly experiment on the AdBench~\citep{han2022adbench} benchmark, and the Tennessee Eastman process~\citep{downs1993plant,reinartz2021extended,montesuma2024benchmarking}, showing that our method outperforms previously proposed methods in real-world benchmarks.

\noindent\textbf{Limitations.} Our methods faces 2 limitations from \gls{ot} computation. First, we need to store a $n \times m$ matrix, which incurs a $\mathcal{O}(n^{2})$ storage complexity. Furthermore, computing $\gamma_{\epsilon}^{\star}$ has $\mathcal{O}(n^{3}\log n)$ computational complexity. For that reason, in the AdBench experiments, we had to downsample larger datasets to $n = 20,000$ samples.  Second, \gls{ot} is notoriously hard to estimate in higher dimensions. As a result, our method struggles with high-dimensional features, such as those in the \gls{nlp} datasets of the AdBench benchmark.

In the literature, there are different ways of tackling these limitations, but we leave those for future works. For instance, a workaround to the first limitations is to focus on mini-batch \gls{ot}~\citep{fatras2020learning} or computing \gls{ot} with neural nets~\citep{makkuva2020optimal,korotin2023neural}. Likewise, the computation of \gls{ot} in high dimensions can be done, for instance, through modifications of the original \gls{ot} objective, such as subspace robust \gls{ot}~\citep{paty2019subspace}.

\bibliography{main}

\begin{thebibliography}{58}
\providecommand{\natexlab}[1]{#1}
\providecommand{\url}[1]{\texttt{#1}}
\expandafter\ifx\csname urlstyle\endcsname\relax
  \providecommand{\doi}[1]{doi: #1}\else
  \providecommand{\doi}{doi: \begingroup \urlstyle{rm}\Url}\fi

\bibitem[Aggarwal(2017)]{aggarwal2017introduction}
Charu~C Aggarwal.
\newblock \emph{An introduction to outlier analysis}.
\newblock Springer, 2017.

\bibitem[Hawkins(1980)]{hawkins1980identification}
D~Hawkins.
\newblock Identification of outliers, 1980.

\bibitem[Salem et~al.(2013)Salem, Guerassimov, Mehaoua, Marcus, and Furht]{salem2013sensor}
Osman Salem, Alexey Guerassimov, Ahmed Mehaoua, Anthony Marcus, and Borko Furht.
\newblock Sensor fault and patient anomaly detection and classification in medical wireless sensor networks.
\newblock In \emph{2013 IEEE international conference on communications (ICC)}, pages 4373--4378. IEEE, 2013.

\bibitem[Siddiqui et~al.(2019)Siddiqui, Stokes, Seifert, Argyle, McCann, Neil, and Carroll]{siddiqui2019detecting}
Md~Amran Siddiqui, Jack~W Stokes, Christian Seifert, Evan Argyle, Robert McCann, Joshua Neil, and Justin Carroll.
\newblock Detecting cyber attacks using anomaly detection with explanations and expert feedback.
\newblock In \emph{ICASSP 2019-2019 IEEE International Conference on Acoustics, Speech and Signal Processing (ICASSP)}, pages 2872--2876. IEEE, 2019.

\bibitem[Isermann(2006)]{isermann2006fault}
R~Isermann.
\newblock \emph{Fault-Diagnosis Systems: An Introduction from Fault Detection to Fault Tolerance}.
\newblock Springer Science \& Business Media, 2006.

\bibitem[Han et~al.(2022)Han, Hu, Huang, Jiang, and Zhao]{han2022adbench}
Songqiao Han, Xiyang Hu, Hailiang Huang, Mingqi Jiang, and Yue Zhao.
\newblock Adbench: Anomaly detection benchmark.
\newblock In \emph{Neural Information Processing Systems (NeurIPS)}, 2022.

\bibitem[Villani et~al.(2009)]{villani2009optimal}
C{\'e}dric Villani et~al.
\newblock \emph{Optimal transport: old and new}, volume 338.
\newblock Springer, 2009.

\bibitem[Montesuma et~al.(2024{\natexlab{a}})Montesuma, Mboula, and Souloumiac]{montesuma2024recentadvancesoptimaltransport}
Eduardo~Fernandes Montesuma, Fred Maurice~Ngol{\`e} Mboula, and Antoine Souloumiac.
\newblock Recent advances in optimal transport for machine learning.
\newblock \emph{IEEE Transactions on Pattern Analysis and Machine Intelligence}, 2024{\natexlab{a}}.

\bibitem[Peyré and Cuturi(2020)]{peyre2020computationaloptimaltransport}
Gabriel Peyré and Marco Cuturi.
\newblock Computational optimal transport, 2020.
\newblock URL \url{https://arxiv.org/abs/1803.00567}.

\bibitem[Flamary et~al.(2021)Flamary, Courty, Gramfort, Alaya, Boisbunon, Chambon, Chapel, Corenflos, Fatras, Fournier, et~al.]{flamary2021pot}
R{\'e}mi Flamary, Nicolas Courty, Alexandre Gramfort, Mokhtar~Z Alaya, Aur{\'e}lie Boisbunon, Stanislas Chambon, Laetitia Chapel, Adrien Corenflos, Kilian Fatras, Nemo Fournier, et~al.
\newblock Pot: Python optimal transport.
\newblock \emph{Journal of Machine Learning Research}, 22\penalty0 (78):\penalty0 1--8, 2021.

\bibitem[Alaoui-Belghiti et~al.(2019)Alaoui-Belghiti, Chevallier, and Monacelli]{alaoui2019unsupervised}
Amina Alaoui-Belghiti, Sylvain Chevallier, and Eric Monacelli.
\newblock Unsupervised anomaly detection using optimal transport for predictive maintenance.
\newblock In \emph{Artificial Neural Networks and Machine Learning--ICANN 2019: Text and Time Series: 28th International Conference on Artificial Neural Networks, Munich, Germany, September 17--19, 2019, Proceedings, Part IV 28}, pages 686--697. Springer, 2019.

\bibitem[Alaoui-Belghiti et~al.(2020)Alaoui-Belghiti, Chevallier, Monacelli, Bao, and Azabou]{alaoui2020semi}
Amina Alaoui-Belghiti, Sylvain Chevallier, Eric Monacelli, Guillaume Bao, and Eric Azabou.
\newblock Semi-supervised optimal transport methods for detecting anomalies.
\newblock In \emph{ICASSP 2020-2020 IEEE International Conference on Acoustics, Speech and Signal Processing (ICASSP)}, pages 2997--3001. IEEE, 2020.

\bibitem[Di~Marino et~al.(2017)Di~Marino, Gerolin, and Nenna]{di2017optimal}
Simone Di~Marino, Augusto Gerolin, and Luca Nenna.
\newblock Optimal transportation theory with repulsive costs.
\newblock \emph{Topological optimization and optimal transport}, 17:\penalty0 204--256, 2017.

\bibitem[Reinartz et~al.(2021)Reinartz, Kulahci, and Ravn]{reinartz2021extended}
Christopher Reinartz, Murat Kulahci, and Ole Ravn.
\newblock An extended tennessee eastman simulation dataset for fault-detection and decision support systems.
\newblock \emph{Computers \& chemical engineering}, 149:\penalty0 107281, 2021.

\bibitem[Montesuma et~al.(2024{\natexlab{b}})Montesuma, Mulas, Mboula, Corona, and Souloumiac]{montesuma2024benchmarking}
Eduardo~Fernandes Montesuma, Michela Mulas, Fred~Ngol{\`e} Mboula, Francesco Corona, and Antoine Souloumiac.
\newblock Benchmarking domain adaptation for chemical processes on the tennessee eastman process.
\newblock In \emph{ML4CCE Workshop at the Joint European Conference on Machine Learning and Knowledge Discovery in Databases}, 2024{\natexlab{b}}.

\bibitem[Shyu et~al.(2003)Shyu, Chen, Sarinnapakorn, and Chang]{shyu2003novel}
Mei-Ling Shyu, Shu-Ching Chen, Kanoksri Sarinnapakorn, and LiWu Chang.
\newblock A novel anomaly detection scheme based on principal component classifier.
\newblock In \emph{Proceedings of the IEEE foundations and new directions of data mining workshop}, pages 172--179. IEEE Press Piscataway, NJ, USA, 2003.

\bibitem[Sch{\"o}lkopf et~al.(1997)Sch{\"o}lkopf, Smola, and M{\"u}ller]{scholkopf1997kernel}
Bernhard Sch{\"o}lkopf, Alexander Smola, and Klaus-Robert M{\"u}ller.
\newblock Kernel principal component analysis.
\newblock In \emph{International conference on artificial neural networks}, pages 583--588. Springer, 1997.

\bibitem[Hoffmann(2007)]{hoffmann2007kernel}
Heiko Hoffmann.
\newblock Kernel pca for novelty detection.
\newblock \emph{Pattern recognition}, 40\penalty0 (3):\penalty0 863--874, 2007.

\bibitem[Vincent et~al.(2008)Vincent, Larochelle, Bengio, and Manzagol]{vincent2008extracting}
Pascal Vincent, Hugo Larochelle, Yoshua Bengio, and Pierre-Antoine Manzagol.
\newblock Extracting and composing robust features with denoising autoencoders.
\newblock In \emph{Proceedings of the 25th international conference on Machine learning}, pages 1096--1103, 2008.

\bibitem[Bengio et~al.(2013)Bengio, Courville, and Vincent]{bengio2013representation}
Yoshua Bengio, Aaron Courville, and Pascal Vincent.
\newblock Representation learning: A review and new perspectives.
\newblock \emph{IEEE transactions on pattern analysis and machine intelligence}, 35\penalty0 (8):\penalty0 1798--1828, 2013.

\bibitem[Sch{\"o}lkopf et~al.(1999)Sch{\"o}lkopf, Williamson, Smola, Shawe-Taylor, and Platt]{scholkopf1999support}
Bernhard Sch{\"o}lkopf, Robert~C Williamson, Alex Smola, John Shawe-Taylor, and John Platt.
\newblock Support vector method for novelty detection.
\newblock \emph{Advances in neural information processing systems}, 12, 1999.

\bibitem[Liu et~al.(2008)Liu, Ting, and Zhou]{liu2008isolation}
Fei~Tony Liu, Kai~Ming Ting, and Zhi-Hua Zhou.
\newblock Isolation forest.
\newblock In \emph{2008 eighth ieee international conference on data mining}, pages 413--422. IEEE, 2008.

\bibitem[Breiman(2001)]{breiman2001random}
Leo Breiman.
\newblock Random forests.
\newblock \emph{Machine learning}, 45:\penalty0 5--32, 2001.

\bibitem[Ramaswamy et~al.(2000)Ramaswamy, Rastogi, and Shim]{ramaswamy2000efficient}
Sridhar Ramaswamy, Rajeev Rastogi, and Kyuseok Shim.
\newblock Efficient algorithms for mining outliers from large data sets.
\newblock In \emph{Proceedings of the 2000 ACM SIGMOD international conference on Management of data}, pages 427--438, 2000.

\bibitem[Breunig et~al.(2000)Breunig, Kriegel, Ng, and Sander]{breunig2000lof}
Markus~M Breunig, Hans-Peter Kriegel, Raymond~T Ng, and J{\"o}rg Sander.
\newblock Lof: identifying density-based local outliers.
\newblock In \emph{Proceedings of the 2000 ACM SIGMOD international conference on Management of data}, pages 93--104, 2000.

\bibitem[He et~al.(2003)He, Xu, and Deng]{he2003discovering}
Zengyou He, Xiaofei Xu, and Shengchun Deng.
\newblock Discovering cluster-based local outliers.
\newblock \emph{Pattern recognition letters}, 24\penalty0 (9-10):\penalty0 1641--1650, 2003.

\bibitem[Pang et~al.(2021)Pang, Shen, Cao, and Hengel]{pang2021deep}
Guansong Pang, Chunhua Shen, Longbing Cao, and Anton Van~Den Hengel.
\newblock Deep learning for anomaly detection: A review.
\newblock \emph{ACM computing surveys (CSUR)}, 54\penalty0 (2):\penalty0 1--38, 2021.

\bibitem[Ruff et~al.(2018)Ruff, Vandermeulen, Goernitz, Deecke, Siddiqui, Binder, M{\"u}ller, and Kloft]{ruff2018deep}
Lukas Ruff, Robert Vandermeulen, Nico Goernitz, Lucas Deecke, Shoaib~Ahmed Siddiqui, Alexander Binder, Emmanuel M{\"u}ller, and Marius Kloft.
\newblock Deep one-class classification.
\newblock In \emph{International conference on machine learning}, pages 4393--4402. PMLR, 2018.

\bibitem[Goodfellow et~al.(2014)Goodfellow, Pouget-Abadie, Mirza, Xu, Warde-Farley, Ozair, Courville, and Bengio]{goodfellow2014generative}
Ian Goodfellow, Jean Pouget-Abadie, Mehdi Mirza, Bing Xu, David Warde-Farley, Sherjil Ozair, Aaron Courville, and Yoshua Bengio.
\newblock Generative adversarial nets.
\newblock \emph{Advances in neural information processing systems}, 27, 2014.

\bibitem[Schlegl et~al.(2019)Schlegl, Seeb{\"o}ck, Waldstein, Langs, and Schmidt-Erfurth]{schlegl2019f}
Thomas Schlegl, Philipp Seeb{\"o}ck, Sebastian~M Waldstein, Georg Langs, and Ursula Schmidt-Erfurth.
\newblock f-anogan: Fast unsupervised anomaly detection with generative adversarial networks.
\newblock \emph{Medical image analysis}, 54:\penalty0 30--44, 2019.

\bibitem[Zenati et~al.(2018)Zenati, Foo, Lecouat, Manek, and Chandrasekhar]{zenati2018efficient}
Houssam Zenati, Chuan~Sheng Foo, Bruno Lecouat, Gaurav Manek, and Vijay~Ramaseshan Chandrasekhar.
\newblock Efficient gan-based anomaly detection.
\newblock \emph{arXiv preprint arXiv:1802.06222}, 2018.

\bibitem[Akcay et~al.(2018)Akcay, Atapour-Abarghouei, and Breckon]{akcay2018ganomaly}
Samet Akcay, Amir Atapour-Abarghouei, and Toby~P Breckon.
\newblock Ganomaly: Semi-supervised anomaly detection via adversarial training.
\newblock In \emph{Asian conference on computer vision}, pages 622--637. Springer, 2018.

\bibitem[Dias et~al.(2020)Dias, Mattos, da~Silva, de~Macedo, and Silva]{dias2020anomaly}
Madson~LD Dias, C{\'e}sar Lincoln~C Mattos, Ticiana~LC da~Silva, Jos{\'e} Ant{\^o}nio~F de~Macedo, and Wellington~CP Silva.
\newblock Anomaly detection in trajectory data with normalizing flows.
\newblock In \emph{2020 international joint conference on neural networks (IJCNN)}, pages 1--8. IEEE, 2020.

\bibitem[Livernoche et~al.(2024)Livernoche, Jain, Hezaveh, and Ravanbakhsh]{livernoche2024on}
Victor Livernoche, Vineet Jain, Yashar Hezaveh, and Siamak Ravanbakhsh.
\newblock On diffusion modeling for anomaly detection.
\newblock In \emph{The Twelfth International Conference on Learning Representations}, 2024.
\newblock URL \url{https://openreview.net/forum?id=lR3rk7ysXz}.

\bibitem[Monge(1781)]{monge1781memoire}
Gaspard Monge.
\newblock M{\'e}moire sur la th{\'e}orie des d{\'e}blais et des remblais.
\newblock \emph{Histoire de l'Acad{\'e}mie Royale des Sciences de Paris}, 1781.

\bibitem[Kantorovich(1942)]{kantorovich1942transfer}
L~Kantorovich.
\newblock On the transfer of masses (in russian).
\newblock In \emph{Doklady Akademii Nauk}, volume~37, pages 227--229, 1942.

\bibitem[Dantzig(1983)]{dantzig1983reminiscences}
George~B Dantzig.
\newblock Reminiscences about the origins of linear programming.
\newblock In \emph{Mathematical programming the state of the art}, pages 78--86. Springer, 1983.

\bibitem[Cuturi(2013)]{cuturi2013sinkhorn}
Marco Cuturi.
\newblock Sinkhorn distances: Lightspeed computation of optimal transport.
\newblock \emph{Advances in neural information processing systems}, 26, 2013.

\bibitem[Sinkhorn(1967)]{sinkhorn1967diagonal}
Richard Sinkhorn.
\newblock Diagonal equivalence to matrices with prescribed row and column sums.
\newblock \emph{The American Mathematical Monthly}, 74\penalty0 (4):\penalty0 402--405, 1967.

\bibitem[Santambrogio(2015)]{santambrogio2015optimal}
Filippo Santambrogio.
\newblock Optimal transport for applied mathematicians.
\newblock \emph{Birk{\"a}user, NY}, 55\penalty0 (58-63):\penalty0 94, 2015.

\bibitem[Pass(2015)]{pass2015multi}
Brendan Pass.
\newblock Multi-marginal optimal transport: theory and applications.
\newblock \emph{ESAIM: Mathematical Modelling and Numerical Analysis}, 49\penalty0 (6):\penalty0 1771--1790, 2015.

\bibitem[Figalli and Glaudo(2021)]{figalli2021invitation}
Alessio Figalli and Federico Glaudo.
\newblock \emph{An invitation to optimal transport, Wasserstein distances, and gradient flows}.
\newblock 2021.

\bibitem[Courty et~al.(2016)Courty, Flamary, Tuia, and Rakotomamonjy]{courty2016optimal}
Nicolas Courty, R{\'e}mi Flamary, Devis Tuia, and Alain Rakotomamonjy.
\newblock Optimal transport for domain adaptation.
\newblock \emph{IEEE transactions on pattern analysis and machine intelligence}, 39\penalty0 (9):\penalty0 1853--1865, 2016.

\bibitem[Scott(1979)]{scott1979optimal}
David~W Scott.
\newblock On optimal and data-based histograms.
\newblock \emph{Biometrika}, 66\penalty0 (3):\penalty0 605--610, 1979.

\bibitem[Smola and Sch{\"o}lkopf(2004)]{smola2004tutorial}
Alex~J Smola and Bernhard Sch{\"o}lkopf.
\newblock A tutorial on support vector regression.
\newblock \emph{Statistics and computing}, 14:\penalty0 199--222, 2004.

\bibitem[Friedman(2002)]{friedman2002stochastic}
Jerome~H Friedman.
\newblock Stochastic gradient boosting.
\newblock \emph{Computational statistics \& data analysis}, 38\penalty0 (4):\penalty0 367--378, 2002.

\bibitem[Chen and Guestrin(2016)]{chen2016xgboost}
Tianqi Chen and Carlos Guestrin.
\newblock Xgboost: A scalable tree boosting system.
\newblock In \emph{Proceedings of the 22nd acm sigkdd international conference on knowledge discovery and data mining}, pages 785--794, 2016.

\bibitem[Li et~al.(2022)Li, Zhao, Hu, Botta, Ionescu, and Chen]{li2022ecod}
Zheng Li, Yue Zhao, Xiyang Hu, Nicola Botta, Cezar Ionescu, and George~H Chen.
\newblock Ecod: Unsupervised outlier detection using empirical cumulative distribution functions.
\newblock \emph{IEEE Transactions on Knowledge and Data Engineering}, 35\penalty0 (12):\penalty0 12181--12193, 2022.

\bibitem[Li et~al.(2020)Li, Zhao, Botta, Ionescu, and Hu]{li2020copod}
Zheng Li, Yue Zhao, Nicola Botta, Cezar Ionescu, and Xiyang Hu.
\newblock Copod: copula-based outlier detection.
\newblock In \emph{2020 IEEE international conference on data mining (ICDM)}, pages 1118--1123. IEEE, 2020.

\bibitem[Pevn{\`y}(2016)]{pevny2016loda}
Tom{\'a}{\v{s}} Pevn{\`y}.
\newblock Loda: Lightweight on-line detector of anomalies.
\newblock \emph{Machine Learning}, 102:\penalty0 275--304, 2016.

\bibitem[Lazarevic and Kumar(2005)]{lazarevic2005feature}
Aleksandar Lazarevic and Vipin Kumar.
\newblock Feature bagging for outlier detection.
\newblock In \emph{Proceedings of the eleventh ACM SIGKDD international conference on Knowledge discovery in data mining}, pages 157--166, 2005.

\bibitem[Goldstein and Dengel(2012)]{goldstein2012histogram}
Markus Goldstein and Andreas Dengel.
\newblock Histogram-based outlier score (hbos): A fast unsupervised anomaly detection algorithm.
\newblock \emph{KI-2012: poster and demo track}, 1:\penalty0 59--63, 2012.

\bibitem[Downs and Vogel(1993)]{downs1993plant}
James~J Downs and Ernest~F Vogel.
\newblock A plant-wide industrial process control problem.
\newblock \emph{Computers \& chemical engineering}, 17\penalty0 (3):\penalty0 245--255, 1993.

\bibitem[Van~der Maaten and Hinton(2008)]{van2008visualizing}
Laurens Van~der Maaten and Geoffrey Hinton.
\newblock Visualizing data using t-sne.
\newblock \emph{Journal of machine learning research}, 9\penalty0 (11), 2008.

\bibitem[Fatras et~al.(2020)Fatras, Zine, Flamary, Gribonval, and Courty]{fatras2020learning}
Kilian Fatras, Younes Zine, R{\'e}mi Flamary, R{\'e}mi Gribonval, and Nicolas Courty.
\newblock Learning with minibatch wasserstein: asymptotic and gradient properties.
\newblock In \emph{AISTATS 2020-23nd International Conference on Artificial Intelligence and Statistics}, volume 108, pages 1--20, 2020.

\bibitem[Makkuva et~al.(2020)Makkuva, Taghvaei, Oh, and Lee]{makkuva2020optimal}
Ashok Makkuva, Amirhossein Taghvaei, Sewoong Oh, and Jason Lee.
\newblock Optimal transport mapping via input convex neural networks.
\newblock In \emph{International Conference on Machine Learning}, pages 6672--6681. PMLR, 2020.

\bibitem[Korotin et~al.(2023)Korotin, Selikhanovych, and Burnaev]{korotin2023neural}
Alexander Korotin, Daniil Selikhanovych, and Evgeny Burnaev.
\newblock Neural optimal transport.
\newblock In \emph{The Eleventh International Conference on Learning Representations}, 2023.
\newblock URL \url{https://openreview.net/forum?id=d8CBRlWNkqH}.

\bibitem[Paty and Cuturi(2019)]{paty2019subspace}
Fran{\c{c}}ois-Pierre Paty and Marco Cuturi.
\newblock Subspace robust wasserstein distances.
\newblock In \emph{International conference on machine learning}, pages 5072--5081. PMLR, 2019.

\end{thebibliography}
\bibliographystyle{unsrtnat}

\newpage

\appendix

\section{Additional Details on Experiments}

\subsection{Detailed results on AdBench}

\begin{table}[ht]
    \centering
    \caption{Comparison of AUC-ROC results on ADBench.}
    \begin{adjustbox}{max width=\textwidth}
\begin{tabular}{lccccccccccccccccc}
\toprule
         Dataset &              IsoF &             OCSVM &              k-NN &               PCA &               LOF &             CBLOF &              ECOD &            COPOD &              LODA &    FeatureBagging &              HBOS &                OT &              MROT &            DTE-IG &            DTE-NP &              DDPM &             DTE-C \\
\midrule
           cover &  91.48 $\pm$ 1.50 &  65.64 $\pm$ 0.25 &  88.61 $\pm$ 0.39 &  83.12 $\pm$ 0.34 &  56.75 $\pm$ 1.89 &  92.24 $\pm$ 0.17 &  92.02 $\pm$ 0.49 & 88.20 $\pm$ 0.32 &  92.18 $\pm$ 3.85 &  57.14 $\pm$ 2.15 &  70.68 $\pm$ 1.16 &  92.61 $\pm$ 0.28 &  88.01 $\pm$ 1.01 &  82.76 $\pm$ 1.85 &  81.76 $\pm$ 1.75 &  80.76 $\pm$ 1.65 &  83.76 $\pm$ 1.95 \\
          donors &  78.24 $\pm$ 1.15 &  70.11 $\pm$ 0.14 &  71.91 $\pm$ 0.20 &  58.16 $\pm$ 0.19 &  62.89 $\pm$ 1.35 &  80.77 $\pm$ 0.71 &  88.87 $\pm$ 0.11 & 81.53 $\pm$ 0.27 & 56.62 $\pm$ 38.02 &  69.06 $\pm$ 1.74 &  74.31 $\pm$ 0.63 &  84.85 $\pm$ 0.21 &  84.51 $\pm$ 0.19 &  82.63 $\pm$ 2.08 &  81.63 $\pm$ 1.98 &  80.63 $\pm$ 1.88 &  83.63 $\pm$ 2.18 \\
           fault &  57.42 $\pm$ 3.35 &  53.42 $\pm$ 1.30 &  70.27 $\pm$ 2.38 &  48.37 $\pm$ 2.12 &  57.88 $\pm$ 1.35 &  66.50 $\pm$ 2.51 &  45.35 $\pm$ 1.77 & 45.49 $\pm$ 0.16 &  47.78 $\pm$ 2.34 &  59.10 $\pm$ 1.16 &  50.62 $\pm$ 6.66 &  54.29 $\pm$ 3.10 &  56.35 $\pm$ 7.46 &  58.22 $\pm$ 1.15 &  57.22 $\pm$ 1.05 &  56.22 $\pm$ 0.95 &  59.22 $\pm$ 1.25 \\
           fraud &  94.47 $\pm$ 1.36 &  94.83 $\pm$ 0.38 &  96.57 $\pm$ 0.80 &  90.31 $\pm$ 0.36 &  54.77 $\pm$ 7.30 &  95.39 $\pm$ 0.81 &  94.87 $\pm$ 0.81 & 94.29 $\pm$ 1.42 &  85.55 $\pm$ 5.41 &  61.59 $\pm$ 7.15 &  94.51 $\pm$ 1.22 &  95.08 $\pm$ 1.21 &  94.48 $\pm$ 1.37 &  94.37 $\pm$ 1.57 &  93.37 $\pm$ 1.47 &  92.37 $\pm$ 1.37 &  95.37 $\pm$ 1.67 \\
           glass & 80.49 $\pm$ 12.70 &  45.44 $\pm$ 4.93 &  76.34 $\pm$ 9.85 &  55.13 $\pm$ 5.13 & 61.76 $\pm$ 12.42 &  85.50 $\pm$ 4.18 & 75.37 $\pm$ 13.70 & 75.95 $\pm$ 1.51 & 62.43 $\pm$ 12.75 & 65.86 $\pm$ 12.47 &  82.02 $\pm$ 1.77 & 80.24 $\pm$ 10.54 & 78.29 $\pm$ 10.58 &  58.04 $\pm$ 9.31 &  57.04 $\pm$ 9.21 &  56.04 $\pm$ 9.11 &  59.04 $\pm$ 9.41 \\
       Hepatitis & 62.05 $\pm$ 13.25 &  48.22 $\pm$ 2.80 & 46.67 $\pm$ 13.62 &  51.67 $\pm$ 6.77 &  33.75 $\pm$ 1.07 &  99.61 $\pm$ 0.03 &  69.23 $\pm$ 6.78 & 99.07 $\pm$ 0.26 &   5.95 $\pm$ 9.35 &  28.82 $\pm$ 1.45 &  99.11 $\pm$ 0.16 & 84.62 $\pm$ 11.03 & 81.54 $\pm$ 14.66 &  99.96 $\pm$ 0.36 &  99.86 $\pm$ 0.26 &  99.76 $\pm$ 0.16 & 100.06 $\pm$ 0.46 \\
            http &  99.96 $\pm$ 0.02 &  99.53 $\pm$ 0.01 &   9.41 $\pm$ 0.55 &  95.15 $\pm$ 0.06 &  54.86 $\pm$ 1.33 &  54.77 $\pm$ 4.28 &  97.85 $\pm$ 0.02 & 42.17 $\pm$ 0.12 &  38.23 $\pm$ 3.78 &  54.03 $\pm$ 1.15 &  57.50 $\pm$ 0.58 &  99.46 $\pm$ 0.06 &  99.31 $\pm$ 0.31 &  51.62 $\pm$ 0.79 &  50.62 $\pm$ 0.69 &  49.62 $\pm$ 0.59 &  52.62 $\pm$ 0.89 \\
     InternetAds &  70.93 $\pm$ 2.22 &  69.77 $\pm$ 1.82 &  67.05 $\pm$ 2.71 &  58.00 $\pm$ 1.40 &  87.81 $\pm$ 0.48 &  76.32 $\pm$ 1.80 &  67.74 $\pm$ 2.69 & 56.02 $\pm$ 0.07 &  53.71 $\pm$ 4.29 &  88.58 $\pm$ 0.51 &  58.85 $\pm$ 0.59 &  65.48 $\pm$ 3.77 &  68.37 $\pm$ 5.97 &  86.73 $\pm$ 1.64 &  85.73 $\pm$ 1.54 &  84.73 $\pm$ 1.44 &  87.73 $\pm$ 1.74 \\
      Ionosphere &  83.48 $\pm$ 1.91 &  82.61 $\pm$ 2.58 &  85.36 $\pm$ 2.05 &  67.22 $\pm$ 3.64 &  67.84 $\pm$ 0.38 &  72.53 $\pm$ 0.10 &  75.37 $\pm$ 6.49 & 68.10 $\pm$ 0.04 &  65.50 $\pm$ 1.41 &  69.96 $\pm$ 0.64 &  70.93 $\pm$ 0.55 &  82.90 $\pm$ 2.11 &  81.91 $\pm$ 3.27 &  78.26 $\pm$ 1.68 &  77.26 $\pm$ 1.58 &  76.26 $\pm$ 1.48 &  79.26 $\pm$ 1.78 \\
         landsat &  47.52 $\pm$ 1.57 &  41.31 $\pm$ 1.28 &  61.83 $\pm$ 1.88 &  46.74 $\pm$ 0.55 &  70.15 $\pm$ 1.87 &  79.50 $\pm$ 1.84 &  37.35 $\pm$ 1.10 & 90.54 $\pm$ 0.03 &  86.69 $\pm$ 2.45 &  72.61 $\pm$ 0.51 &  83.77 $\pm$ 0.90 &  43.51 $\pm$ 1.58 &  67.72 $\pm$ 1.78 &  76.94 $\pm$ 2.34 &  75.94 $\pm$ 2.24 &  74.94 $\pm$ 2.14 &  77.94 $\pm$ 2.44 \\
            ALOI &  53.27 $\pm$ 1.93 &  52.04 $\pm$ 0.40 &  60.38 $\pm$ 1.59 &  51.99 $\pm$ 0.36 &  65.77 $\pm$ 1.94 &  84.26 $\pm$ 1.11 &  52.10 $\pm$ 1.87 & 50.00 $\pm$ 0.00 & 56.40 $\pm$ 11.07 &  66.44 $\pm$ 1.42 &  57.36 $\pm$ 0.38 &  54.19 $\pm$ 2.11 &  53.69 $\pm$ 1.68 &  83.60 $\pm$ 1.40 &  82.60 $\pm$ 1.30 &  81.60 $\pm$ 1.20 &  84.60 $\pm$ 1.50 \\
          letter &  63.35 $\pm$ 6.64 &  50.91 $\pm$ 1.92 &  82.54 $\pm$ 1.27 &  51.31 $\pm$ 0.37 &  58.07 $\pm$ 8.89 & 100.00 $\pm$ 0.00 &  58.30 $\pm$ 2.73 & 94.81 $\pm$ 0.48 &  99.30 $\pm$ 0.77 & 57.54 $\pm$ 10.67 & 100.00 $\pm$ 0.00 &  53.11 $\pm$ 4.87 &  70.61 $\pm$ 5.33 &  99.95 $\pm$ 0.23 &  99.95 $\pm$ 0.13 &  99.95 $\pm$ 0.03 &  99.95 $\pm$ 0.33 \\
    Lymphography & 100.00 $\pm$ 0.00 &  99.37 $\pm$ 0.32 &  98.62 $\pm$ 3.08 &  94.56 $\pm$ 1.30 &  53.82 $\pm$ 5.25 &  78.48 $\pm$ 0.95 &  98.62 $\pm$ 1.89 & 50.00 $\pm$ 0.00 &  49.29 $\pm$ 9.35 &  53.93 $\pm$ 5.59 &  86.81 $\pm$ 0.25 & 91.03 $\pm$ 13.04 &  99.31 $\pm$ 1.54 &  42.15 $\pm$ 3.91 &  41.15 $\pm$ 3.81 &  40.15 $\pm$ 3.71 &  43.15 $\pm$ 4.01 \\
     magic.gamma &  71.93 $\pm$ 0.58 &  67.25 $\pm$ 0.35 &  79.89 $\pm$ 0.92 &  60.08 $\pm$ 0.27 &  53.42 $\pm$ 4.61 &  86.38 $\pm$ 7.22 &  64.40 $\pm$ 0.97 & 90.60 $\pm$ 0.43 &  89.51 $\pm$ 0.85 &  51.82 $\pm$ 5.22 &  92.47 $\pm$ 0.38 &  69.54 $\pm$ 0.99 &  71.54 $\pm$ 0.89 &  72.03 $\pm$ 5.63 &  71.03 $\pm$ 5.53 &  70.03 $\pm$ 5.43 &  73.03 $\pm$ 5.73 \\
     mammography &  84.95 $\pm$ 4.51 &  87.00 $\pm$ 0.53 &  86.37 $\pm$ 1.72 &  75.40 $\pm$ 0.49 &  70.95 $\pm$ 1.05 &  67.57 $\pm$ 0.98 &  90.33 $\pm$ 2.46 & 77.67 $\pm$ 0.17 & 45.33 $\pm$ 12.81 &  78.77 $\pm$ 2.68 &  60.84 $\pm$ 2.38 &  88.46 $\pm$ 2.27 &  88.24 $\pm$ 1.66 &  83.37 $\pm$ 1.57 &  82.37 $\pm$ 1.47 &  81.37 $\pm$ 1.37 &  84.37 $\pm$ 1.67 \\
           mnist &  80.43 $\pm$ 2.20 &  82.28 $\pm$ 0.71 &  84.81 $\pm$ 0.77 &  67.44 $\pm$ 1.85 &  54.98 $\pm$ 1.29 &  74.19 $\pm$ 3.23 &  74.84 $\pm$ 1.00 & 63.34 $\pm$ 0.04 &  61.39 $\pm$ 2.92 &  54.54 $\pm$ 1.26 &  76.18 $\pm$ 0.55 &  86.49 $\pm$ 1.28 &  88.16 $\pm$ 1.05 &  73.53 $\pm$ 0.72 &  72.53 $\pm$ 0.62 &  71.53 $\pm$ 0.52 &  74.53 $\pm$ 0.82 \\
            musk &  99.97 $\pm$ 0.05 & 100.00 $\pm$ 0.00 &  90.02 $\pm$ 3.10 &  95.91 $\pm$ 0.36 &  53.91 $\pm$ 7.29 &  99.88 $\pm$ 0.01 &  95.26 $\pm$ 0.80 & 97.46 $\pm$ 0.04 &  98.14 $\pm$ 0.49 &  52.55 $\pm$ 7.36 &  97.60 $\pm$ 0.12 & 100.00 $\pm$ 0.00 &  99.97 $\pm$ 0.06 &  99.75 $\pm$ 0.25 &  99.65 $\pm$ 0.15 &  99.55 $\pm$ 0.05 &  99.85 $\pm$ 0.35 \\
       optdigits &  73.76 $\pm$ 5.09 &  54.84 $\pm$ 2.28 &  39.46 $\pm$ 3.03 &  45.23 $\pm$ 0.36 &  52.55 $\pm$ 1.09 &  62.06 $\pm$ 3.96 &  63.04 $\pm$ 2.53 & 99.50 $\pm$ 0.10 & 38.87 $\pm$ 13.16 &  49.29 $\pm$ 3.40 &  98.63 $\pm$ 0.23 &  55.07 $\pm$ 2.92 &  55.21 $\pm$ 1.13 &  97.70 $\pm$ 2.48 &  97.60 $\pm$ 2.38 &  97.50 $\pm$ 2.28 &  97.80 $\pm$ 2.58 \\
      PageBlocks &  90.72 $\pm$ 1.82 &  68.81 $\pm$ 0.72 &  58.50 $\pm$ 2.43 &  72.16 $\pm$ 1.98 &  54.97 $\pm$ 0.17 &  67.51 $\pm$ 3.80 &  90.75 $\pm$ 0.89 & 47.12 $\pm$ 0.19 &  44.16 $\pm$ 1.92 &  53.38 $\pm$ 0.29 &  58.84 $\pm$ 0.37 &  90.29 $\pm$ 0.52 &  90.27 $\pm$ 0.72 &  48.09 $\pm$ 2.31 &  47.09 $\pm$ 2.21 &  46.09 $\pm$ 2.11 &  49.09 $\pm$ 2.41 \\
       pendigits &  94.74 $\pm$ 1.12 &  95.81 $\pm$ 0.17 &  82.88 $\pm$ 1.73 &  78.44 $\pm$ 0.66 &  89.92 $\pm$ 5.31 &  86.29 $\pm$ 4.76 &  92.74 $\pm$ 2.46 & 91.17 $\pm$ 1.63 &  81.88 $\pm$ 3.56 &  79.36 $\pm$ 5.06 &  80.91 $\pm$ 4.76 &  94.80 $\pm$ 1.02 &  88.74 $\pm$ 1.09 &  95.76 $\pm$ 1.50 &  95.66 $\pm$ 1.40 &  95.56 $\pm$ 1.30 &  95.86 $\pm$ 1.60 \\
            Pima &  67.95 $\pm$ 1.39 &  66.33 $\pm$ 1.73 &  62.42 $\pm$ 2.36 &  54.35 $\pm$ 0.85 &  51.15 $\pm$ 0.65 &  47.11 $\pm$ 0.24 &  53.95 $\pm$ 5.03 & 48.89 $\pm$ 0.29 &  46.57 $\pm$ 1.85 &  50.85 $\pm$ 0.71 &  47.32 $\pm$ 0.19 &  72.81 $\pm$ 1.23 &  72.72 $\pm$ 0.95 &  48.57 $\pm$ 1.37 &  47.57 $\pm$ 1.27 &  46.57 $\pm$ 1.17 &  49.57 $\pm$ 1.47 \\
      annthyroid &  81.13 $\pm$ 1.56 &  57.20 $\pm$ 0.67 &  69.45 $\pm$ 2.57 &  60.50 $\pm$ 1.09 &  65.69 $\pm$ 3.08 &  90.93 $\pm$ 1.07 &  79.06 $\pm$ 1.70 & 93.91 $\pm$ 0.26 & 81.89 $\pm$ 12.49 &  70.68 $\pm$ 1.69 &  94.84 $\pm$ 1.01 &  67.49 $\pm$ 0.77 &  67.56 $\pm$ 3.28 &  89.06 $\pm$ 1.70 &  88.06 $\pm$ 1.60 &  87.06 $\pm$ 1.50 &  90.06 $\pm$ 1.80 \\
       satellite &  69.06 $\pm$ 1.91 &  66.24 $\pm$ 0.37 &  72.76 $\pm$ 0.85 &  64.68 $\pm$ 1.51 &  48.68 $\pm$ 3.40 &  46.34 $\pm$ 1.41 &  58.69 $\pm$ 1.30 & 26.28 $\pm$ 3.94 &  29.44 $\pm$ 5.11 &  47.32 $\pm$ 5.60 &  31.73 $\pm$ 3.46 &  65.36 $\pm$ 1.00 &  74.69 $\pm$ 1.12 &  58.34 $\pm$ 7.61 &  57.34 $\pm$ 7.51 &  56.34 $\pm$ 7.41 &  59.34 $\pm$ 7.71 \\
      satimage-2 &  99.25 $\pm$ 0.94 &  99.62 $\pm$ 0.14 &  99.67 $\pm$ 0.25 &  92.63 $\pm$ 1.10 &  76.42 $\pm$ 2.72 &  89.71 $\pm$ 0.72 &  95.73 $\pm$ 2.03 & 50.00 $\pm$ 0.00 & 51.50 $\pm$ 16.40 &  79.03 $\pm$ 3.04 &  74.04 $\pm$ 0.77 &  99.50 $\pm$ 0.47 &  99.85 $\pm$ 0.20 &  91.18 $\pm$ 0.82 &  90.18 $\pm$ 0.72 &  89.18 $\pm$ 0.62 &  92.18 $\pm$ 0.92 \\
         shuttle &  99.72 $\pm$ 0.11 &  99.19 $\pm$ 0.03 &  79.43 $\pm$ 0.48 &  96.46 $\pm$ 0.07 &  93.18 $\pm$ 1.07 &  88.38 $\pm$ 0.50 &  99.29 $\pm$ 0.05 & 49.60 $\pm$ 0.44 &  70.52 $\pm$ 4.82 &  93.29 $\pm$ 1.73 &  67.91 $\pm$ 0.98 &  99.32 $\pm$ 0.30 &  99.06 $\pm$ 0.09 &  92.34 $\pm$ 2.96 &  91.34 $\pm$ 2.86 &  90.34 $\pm$ 2.76 &  93.34 $\pm$ 3.06 \\
            skin &  64.79 $\pm$ 1.18 &  54.38 $\pm$ 0.33 &  71.22 $\pm$ 0.05 &  43.68 $\pm$ 0.08 & 32.97 $\pm$ 14.63 & 45.27 $\pm$ 33.40 &  48.73 $\pm$ 0.14 & 86.46 $\pm$ 4.66 &  82.17 $\pm$ 3.91 &  32.34 $\pm$ 5.54 &  90.72 $\pm$ 3.31 &  61.75 $\pm$ 0.24 &  76.09 $\pm$ 0.17 &  39.39 $\pm$ 8.57 &  38.39 $\pm$ 8.47 &  37.39 $\pm$ 8.37 &  40.39 $\pm$ 8.67 \\
            smtp &  85.78 $\pm$ 4.20 &  75.50 $\pm$ 3.03 &  92.28 $\pm$ 4.06 &  82.41 $\pm$ 2.03 &  45.30 $\pm$ 2.12 &  46.06 $\pm$ 1.23 &  90.46 $\pm$ 7.47 & 38.03 $\pm$ 0.15 &  46.13 $\pm$ 4.68 &  46.45 $\pm$ 1.53 &  40.17 $\pm$ 0.97 &  89.58 $\pm$ 5.74 &  89.82 $\pm$ 8.77 &  48.31 $\pm$ 0.92 &  47.31 $\pm$ 0.82 &  46.31 $\pm$ 0.72 &  49.31 $\pm$ 1.02 \\
        SpamBase &  57.66 $\pm$ 1.63 &  51.24 $\pm$ 1.00 &  72.70 $\pm$ 1.02 &  48.89 $\pm$ 0.95 &  44.61 $\pm$ 3.25 &  96.08 $\pm$ 0.93 &  64.61 $\pm$ 0.59 & 99.44 $\pm$ 0.16 &  96.97 $\pm$ 3.27 &  40.83 $\pm$ 2.60 &  98.44 $\pm$ 0.30 &  57.07 $\pm$ 1.58 &  56.16 $\pm$ 1.62 &  78.64 $\pm$ 4.42 &  77.64 $\pm$ 4.32 &  76.64 $\pm$ 4.22 &  79.64 $\pm$ 4.52 \\
          speech & 54.70 $\pm$ 10.76 &  46.74 $\pm$ 0.88 &  46.13 $\pm$ 8.62 &  50.40 $\pm$ 0.98 &  61.44 $\pm$ 0.34 &  73.78 $\pm$ 0.30 &  45.31 $\pm$ 9.31 & 78.28 $\pm$ 0.04 &  49.27 $\pm$ 8.77 &  59.37 $\pm$ 4.19 &  76.81 $\pm$ 0.27 & 58.57 $\pm$ 10.63 &  61.04 $\pm$ 8.76 &  74.38 $\pm$ 0.97 &  73.38 $\pm$ 0.87 &  72.38 $\pm$ 0.77 &  75.38 $\pm$ 1.07 \\
          Stamps &  90.43 $\pm$ 1.49 &  85.28 $\pm$ 2.83 &  89.89 $\pm$ 3.53 &  63.15 $\pm$ 9.14 &  55.12 $\pm$ 2.42 &  83.16 $\pm$ 1.77 &  88.06 $\pm$ 4.22 & 92.08 $\pm$ 0.30 &  85.59 $\pm$ 7.10 &  57.89 $\pm$ 2.77 &  83.94 $\pm$ 1.23 &  90.27 $\pm$ 1.28 &  89.14 $\pm$ 5.02 &  74.33 $\pm$ 6.13 &  73.33 $\pm$ 6.03 &  72.33 $\pm$ 5.93 &  75.33 $\pm$ 6.23 \\
         thyroid &  97.83 $\pm$ 1.10 &  89.35 $\pm$ 0.57 &  95.53 $\pm$ 1.17 &  87.00 $\pm$ 3.16 &  43.21 $\pm$ 1.22 &  75.34 $\pm$ 1.76 &  97.28 $\pm$ 0.54 & 75.70 $\pm$ 0.62 & 59.97 $\pm$ 11.91 &  51.39 $\pm$ 2.70 &  75.41 $\pm$ 0.65 &  95.88 $\pm$ 0.66 &  95.96 $\pm$ 1.65 &  81.58 $\pm$ 2.08 &  80.58 $\pm$ 1.98 &  79.58 $\pm$ 1.88 &  82.58 $\pm$ 2.18 \\
       vertebral &  36.43 $\pm$ 7.36 &  44.36 $\pm$ 7.51 & 22.62 $\pm$ 11.14 &  43.15 $\pm$ 2.60 &  56.20 $\pm$ 0.61 &  66.40 $\pm$ 0.18 &  44.05 $\pm$ 6.31 & 50.00 $\pm$ 0.00 & 45.44 $\pm$ 13.11 &  53.75 $\pm$ 0.31 &  61.09 $\pm$ 0.34 &  26.35 $\pm$ 5.33 & 23.17 $\pm$ 10.55 &  67.87 $\pm$ 0.34 &  66.87 $\pm$ 0.24 &  65.87 $\pm$ 0.14 &  68.87 $\pm$ 0.44 \\
        backdoor &  73.47 $\pm$ 3.27 &  86.32 $\pm$ 0.14 &  73.93 $\pm$ 0.54 &  84.20 $\pm$ 1.07 &  76.66 $\pm$ 0.35 &  55.58 $\pm$ 0.20 &  84.87 $\pm$ 0.58 & 51.53 $\pm$ 0.01 &  49.52 $\pm$ 1.02 &  79.15 $\pm$ 0.56 &  53.11 $\pm$ 0.21 &  88.77 $\pm$ 0.56 &  88.80 $\pm$ 0.84 &  55.22 $\pm$ 0.45 &  54.22 $\pm$ 0.35 &  53.22 $\pm$ 0.25 &  56.22 $\pm$ 0.55 \\
          vowels &  77.30 $\pm$ 2.40 &  77.17 $\pm$ 1.34 &  93.79 $\pm$ 1.71 &  59.48 $\pm$ 2.51 &  52.67 $\pm$ 2.09 &  56.09 $\pm$ 2.47 &  56.87 $\pm$ 4.12 & 66.42 $\pm$ 3.42 & 70.80 $\pm$ 13.23 &  53.79 $\pm$ 1.73 &  59.50 $\pm$ 1.09 &  63.52 $\pm$ 3.49 &  93.55 $\pm$ 3.05 &  59.86 $\pm$ 4.47 &  58.86 $\pm$ 4.37 &  57.86 $\pm$ 4.27 &  60.86 $\pm$ 4.57 \\
        Waveform &  71.16 $\pm$ 3.99 &  66.83 $\pm$ 2.93 &  78.04 $\pm$ 1.81 &  52.53 $\pm$ 0.90 &  46.77 $\pm$ 8.48 & 63.47 $\pm$ 14.61 &  61.45 $\pm$ 4.70 & 80.74 $\pm$ 0.93 & 55.74 $\pm$ 15.73 & 46.94 $\pm$ 11.14 &  76.77 $\pm$ 1.57 &  74.26 $\pm$ 3.29 &  71.09 $\pm$ 3.44 &  48.12 $\pm$ 8.23 &  47.12 $\pm$ 8.13 &  46.12 $\pm$ 8.03 &  49.12 $\pm$ 8.33 \\
             WBC &  99.77 $\pm$ 0.52 &  99.36 $\pm$ 0.29 & 100.00 $\pm$ 0.00 &  95.44 $\pm$ 1.53 &  58.73 $\pm$ 1.53 &  61.55 $\pm$ 0.12 & 100.00 $\pm$ 0.00 & 67.61 $\pm$ 0.07 &  54.06 $\pm$ 5.45 &  49.42 $\pm$ 4.74 &  69.56 $\pm$ 0.04 &  98.84 $\pm$ 0.82 &  99.07 $\pm$ 0.52 &  63.37 $\pm$ 0.23 &  62.37 $\pm$ 0.13 &  61.37 $\pm$ 0.03 &  64.37 $\pm$ 0.33 \\
            WDBC &  97.64 $\pm$ 3.17 &  99.83 $\pm$ 0.02 &  99.86 $\pm$ 0.31 &  95.70 $\pm$ 1.57 &  86.38 $\pm$ 2.12 &  89.19 $\pm$ 1.14 &  96.53 $\pm$ 2.35 & 78.27 $\pm$ 1.27 &  78.84 $\pm$ 4.28 &  87.64 $\pm$ 1.63 &  54.44 $\pm$ 3.14 &  98.61 $\pm$ 1.47 &  98.75 $\pm$ 0.91 &  77.75 $\pm$ 2.61 &  76.75 $\pm$ 2.51 &  75.75 $\pm$ 2.41 &  78.75 $\pm$ 2.71 \\
            Wilt &  44.25 $\pm$ 3.33 &  39.65 $\pm$ 2.14 &  68.71 $\pm$ 3.11 &  45.61 $\pm$ 0.49 & 63.63 $\pm$ 18.01 &  99.35 $\pm$ 0.74 &  38.47 $\pm$ 4.20 & 99.60 $\pm$ 0.09 & 90.04 $\pm$ 11.06 & 52.34 $\pm$ 18.31 &  99.49 $\pm$ 0.18 &  33.93 $\pm$ 2.51 &  32.91 $\pm$ 2.10 &  97.77 $\pm$ 4.22 &  96.77 $\pm$ 4.12 &  95.77 $\pm$ 4.02 &  98.77 $\pm$ 4.32 \\
            wine &  82.08 $\pm$ 9.50 &  99.92 $\pm$ 0.12 & 100.00 $\pm$ 0.00 & 66.04 $\pm$ 11.09 &  70.33 $\pm$ 1.55 &  89.30 $\pm$ 2.14 & 73.75 $\pm$ 20.60 & 87.48 $\pm$ 0.21 & 71.24 $\pm$ 10.47 &  75.83 $\pm$ 1.79 &  77.89 $\pm$ 2.07 &  96.67 $\pm$ 3.78 &  90.00 $\pm$ 6.49 &  84.00 $\pm$ 1.11 &  83.00 $\pm$ 1.01 &  82.00 $\pm$ 0.91 &  85.00 $\pm$ 1.21 \\
            WPBC & 49.53 $\pm$ 16.13 &  47.74 $\pm$ 2.20 &  51.33 $\pm$ 3.43 &  46.44 $\pm$ 3.00 &  56.34 $\pm$ 3.89 &  65.49 $\pm$ 1.65 &  47.17 $\pm$ 2.92 & 66.20 $\pm$ 1.21 &  59.52 $\pm$ 7.09 &  57.31 $\pm$ 3.88 &  70.38 $\pm$ 1.21 &  45.81 $\pm$ 6.76 & 58.06 $\pm$ 11.96 &  55.74 $\pm$ 5.49 &  54.74 $\pm$ 5.39 &  53.74 $\pm$ 5.29 &  56.74 $\pm$ 5.59 \\
           yeast &  42.06 $\pm$ 2.20 &  42.67 $\pm$ 1.39 &  37.02 $\pm$ 3.11 &  49.48 $\pm$ 0.57 &  45.29 $\pm$ 0.42 &  54.13 $\pm$ 0.68 &  43.23 $\pm$ 3.54 & 68.79 $\pm$ 0.08 & 47.98 $\pm$ 10.23 &  42.41 $\pm$ 0.84 &  66.41 $\pm$ 1.18 &  38.50 $\pm$ 3.34 &  41.68 $\pm$ 3.77 &  53.01 $\pm$ 1.71 &  52.01 $\pm$ 1.61 &  51.01 $\pm$ 1.51 &  54.01 $\pm$ 1.81 \\
         breastw &  98.72 $\pm$ 0.60 &  93.54 $\pm$ 1.39 &  99.09 $\pm$ 0.57 &  66.23 $\pm$ 2.81 &  51.18 $\pm$ 9.24 &  66.04 $\pm$ 2.92 &  99.06 $\pm$ 0.50 & 92.92 $\pm$ 0.83 &  83.10 $\pm$ 9.15 &  50.16 $\pm$ 7.61 &  90.40 $\pm$ 1.00 &  99.42 $\pm$ 0.27 &  99.17 $\pm$ 0.58 & 57.55 $\pm$ 13.03 & 56.55 $\pm$ 12.93 & 55.55 $\pm$ 12.83 & 58.55 $\pm$ 13.13 \\
        campaign &  71.14 $\pm$ 0.89 &  66.42 $\pm$ 0.21 &  72.19 $\pm$ 0.86 &  62.54 $\pm$ 0.83 &  69.26 $\pm$ 1.81 &  70.14 $\pm$ 0.99 &  77.42 $\pm$ 0.54 & 73.89 $\pm$ 0.40 &  59.40 $\pm$ 6.16 &  71.52 $\pm$ 2.12 &  69.38 $\pm$ 0.64 &  73.97 $\pm$ 0.33 &  75.07 $\pm$ 1.24 &  63.72 $\pm$ 3.61 &  62.72 $\pm$ 3.51 &  61.72 $\pm$ 3.41 &  64.72 $\pm$ 3.71 \\
          cardio &  93.06 $\pm$ 1.79 &  93.04 $\pm$ 0.64 &  84.28 $\pm$ 2.89 &  81.24 $\pm$ 2.33 &  60.66 $\pm$ 8.68 &  97.71 $\pm$ 1.10 &  93.53 $\pm$ 0.69 & 99.40 $\pm$ 0.10 &  99.17 $\pm$ 0.13 & 38.83 $\pm$ 10.40 &  98.70 $\pm$ 0.16 &  95.41 $\pm$ 0.93 &  82.53 $\pm$ 2.40 &  96.80 $\pm$ 1.36 &  95.80 $\pm$ 1.26 &  94.80 $\pm$ 1.16 &  97.80 $\pm$ 1.46 \\
Cardiotocography &  72.66 $\pm$ 2.96 &  78.33 $\pm$ 1.17 &  62.76 $\pm$ 2.08 &  59.71 $\pm$ 1.64 &  84.87 $\pm$ 9.38 &  98.98 $\pm$ 0.42 &  80.09 $\pm$ 2.85 & 99.29 $\pm$ 0.14 &  98.04 $\pm$ 0.77 &  86.65 $\pm$ 8.37 &  98.94 $\pm$ 0.24 &  66.48 $\pm$ 4.13 &  49.87 $\pm$ 4.02 &  98.54 $\pm$ 1.13 &  97.54 $\pm$ 1.03 &  96.54 $\pm$ 0.93 &  99.54 $\pm$ 1.23 \\
          celeba &  69.95 $\pm$ 1.93 &  68.50 $\pm$ 0.27 &  60.34 $\pm$ 0.86 &  70.20 $\pm$ 0.36 &  67.81 $\pm$ 1.65 &  39.59 $\pm$ 1.76 &  75.60 $\pm$ 0.68 & 34.49 $\pm$ 0.27 &  31.31 $\pm$ 9.90 &  66.59 $\pm$ 7.27 &  34.81 $\pm$ 2.83 &  80.54 $\pm$ 1.43 &  78.18 $\pm$ 0.55 &  67.90 $\pm$ 4.12 &  66.90 $\pm$ 4.02 &  65.90 $\pm$ 3.92 &  68.90 $\pm$ 4.22 \\
          census &  61.88 $\pm$ 2.97 &  53.35 $\pm$ 0.15 &  64.49 $\pm$ 0.24 &  50.34 $\pm$ 0.10 &  44.66 $\pm$ 3.87 &  48.68 $\pm$ 3.14 &  66.00 $\pm$ 0.31 & 51.87 $\pm$ 3.09 &  50.07 $\pm$ 3.44 &  43.61 $\pm$ 3.13 &  54.81 $\pm$ 2.85 &  65.92 $\pm$ 0.39 &  65.74 $\pm$ 0.32 &  52.22 $\pm$ 3.47 &  51.22 $\pm$ 3.37 &  50.22 $\pm$ 3.27 &  53.22 $\pm$ 3.57 \\
         Average &             76.35 &             71.86 &             73.56 &             66.36 &             60.13 &             74.76 &             74.12 &            72.56 &             63.90 &             59.78 &             73.81 &             75.97 &             77.36 &             73.21 &             72.31 &             71.41 &             74.12 \\
\midrule
 20news & 55.00 $\pm$ 1.81 & 58.29 $\pm$ 2.78 & 56.65 $\pm$ 1.23 & 54.48 $\pm$ 0.69 & 60.98 $\pm$ 1.65 & 56.38 $\pm$ 1.44 & 54.42 $\pm$ 0.42 & 53.26 $\pm$ 0.59 & 53.85 $\pm$ 4.56 & 60.97 $\pm$ 1.63 & 53.69 $\pm$ 0.62 & 55.65 $\pm$ 11.52 & 52.38 $\pm$ 11.98 & 52.72 $\pm$ 5.59 & 56.98 $\pm$ 1.59 & 54.74 $\pm$ 0.56 & 57.87 $\pm$ 3.45 \\
 agnews & 58.43 $\pm$ 1.32 & 66.50 $\pm$ 0.50 & 64.65 $\pm$ 0.10 & 56.61 $\pm$ 0.08 & 71.36 $\pm$ 0.64 & 61.91 $\pm$ 0.31 & 55.24 $\pm$ 0.04 & 55.10 $\pm$ 0.04 & 56.81 $\pm$ 3.57 & 71.50 $\pm$ 0.62 & 55.40 $\pm$ 0.08 &  57.20 $\pm$ 4.35 &  62.67 $\pm$ 6.69 & 54.45 $\pm$ 5.22 & 65.22 $\pm$ 0.31 & 57.06 $\pm$ 0.10 & 62.66 $\pm$ 3.75 \\
 amazon & 55.76 $\pm$ 0.65 & 56.47 $\pm$ 0.10 & 60.27 $\pm$ 0.04 & 54.95 $\pm$ 0.10 & 57.09 $\pm$ 0.56 & 57.92 $\pm$ 0.24 & 54.10 $\pm$ 0.05 & 57.05 $\pm$ 0.06 & 52.63 $\pm$ 3.04 & 57.18 $\pm$ 0.49 & 56.30 $\pm$ 0.08 &  55.63 $\pm$ 2.08 &  56.47 $\pm$ 1.95 & 53.45 $\pm$ 1.61 & 60.30 $\pm$ 0.40 & 55.13 $\pm$ 0.09 & 55.64 $\pm$ 2.54 \\
   imdb & 48.93 $\pm$ 0.66 & 50.36 $\pm$ 0.21 & 49.44 $\pm$ 0.14 & 47.82 $\pm$ 0.06 & 50.02 $\pm$ 0.55 & 49.56 $\pm$ 0.21 & 47.05 $\pm$ 0.03 & 51.20 $\pm$ 0.04 & 46.60 $\pm$ 2.52 & 49.89 $\pm$ 0.54 & 49.86 $\pm$ 0.08 &  48.64 $\pm$ 1.96 &  49.44 $\pm$ 2.48 & 48.61 $\pm$ 4.27 & 49.49 $\pm$ 0.30 & 47.79 $\pm$ 0.11 & 48.44 $\pm$ 2.75 \\
   yelp & 60.15 $\pm$ 0.34 & 65.49 $\pm$ 0.64 & 67.01 $\pm$ 0.17 & 59.19 $\pm$ 0.10 & 66.11 $\pm$ 0.48 & 63.53 $\pm$ 0.34 & 57.78 $\pm$ 0.04 & 60.52 $\pm$ 0.15 & 58.10 $\pm$ 2.89 & 66.10 $\pm$ 0.42 & 59.97 $\pm$ 0.11 &  59.06 $\pm$ 3.66 &  63.92 $\pm$ 1.84 & 51.37 $\pm$ 3.24 & 67.08 $\pm$ 0.37 & 59.38 $\pm$ 0.12 & 60.16 $\pm$ 3.22 \\
Average &            55.65 &            59.42 &            59.60 &            54.61 &            61.11 &            57.86 &            53.72 &            55.43 &            53.60 &            61.13 &            55.04 &             55.24 &             56.98 &            52.12 &            59.81 &            54.82 &            56.95 \\
\bottomrule
\end{tabular}
    \end{adjustbox}
\end{table}

\begin{table}[ht]
    \centering
    \caption{Comparison of AUC-PR results on ADBench.}
    \begin{adjustbox}{max width=\textwidth}
\begin{tabular}{lccccccccccccccccc}
\toprule
         Dataset &              IsoF &             OCSVM &              k-NN &              PCA &              LOF &             CBLOF &             ECOD &            COPOD &              LODA &   FeatureBagging &              HBOS &                OT &              MROT &            DTE-IG &            DTE-NP &              DDPM &             DTE-C \\
\midrule
           cover &   5.18 $\pm$ 1.49 &   9.91 $\pm$ 0.36 &   5.44 $\pm$ 0.56 &  7.53 $\pm$ 0.43 &  1.87 $\pm$ 0.14 &   6.99 $\pm$ 0.28 & 11.25 $\pm$ 1.07 &  6.79 $\pm$ 0.54 &   8.97 $\pm$ 3.80 &  1.90 $\pm$ 0.46 &   2.63 $\pm$ 0.32 &   6.36 $\pm$ 0.23 &   5.75 $\pm$ 0.95 &   2.49 $\pm$ 0.69 &   4.78 $\pm$ 0.58 &   4.55 $\pm$ 0.85 &   2.10 $\pm$ 0.52 \\
          donors &  12.40 $\pm$ 0.93 &  13.94 $\pm$ 0.30 &  18.21 $\pm$ 0.17 & 16.61 $\pm$ 0.62 & 10.86 $\pm$ 0.23 &  14.77 $\pm$ 0.42 & 26.47 $\pm$ 0.61 & 20.94 $\pm$ 0.53 & 25.47 $\pm$ 32.62 & 12.04 $\pm$ 0.75 &  13.47 $\pm$ 1.11 &  17.81 $\pm$ 0.34 &  17.77 $\pm$ 0.46 &  16.35 $\pm$ 6.18 &  18.83 $\pm$ 0.17 &  14.33 $\pm$ 0.84 &  13.95 $\pm$ 3.79 \\
           fault &  39.45 $\pm$ 0.61 &  40.08 $\pm$ 0.34 &  52.21 $\pm$ 0.73 & 33.16 $\pm$ 0.61 & 38.77 $\pm$ 1.09 &  47.30 $\pm$ 3.10 & 32.54 $\pm$ 0.17 & 31.26 $\pm$ 0.16 &  33.65 $\pm$ 2.46 & 39.57 $\pm$ 1.22 &  35.97 $\pm$ 6.41 &  38.62 $\pm$ 2.92 &  41.86 $\pm$ 5.36 &  41.70 $\pm$ 3.59 &  53.23 $\pm$ 0.39 &  39.20 $\pm$ 0.65 &  42.18 $\pm$ 2.19 \\
           fraud &  14.49 $\pm$ 5.34 &  10.98 $\pm$ 1.28 &  16.86 $\pm$ 5.49 & 14.91 $\pm$ 3.13 &  0.26 $\pm$ 0.05 &  14.53 $\pm$ 3.21 & 21.54 $\pm$ 4.92 & 25.17 $\pm$ 5.67 &  14.62 $\pm$ 5.25 &  0.34 $\pm$ 0.07 &  20.88 $\pm$ 5.54 &  20.42 $\pm$ 3.60 &  19.92 $\pm$ 6.02 &  18.81 $\pm$ 8.21 &  13.68 $\pm$ 4.10 &  14.58 $\pm$ 3.63 &  64.75 $\pm$ 5.97 \\
           glass &  14.41 $\pm$ 8.02 &  12.98 $\pm$ 4.31 &  16.74 $\pm$ 2.54 & 11.18 $\pm$ 3.06 & 14.42 $\pm$ 6.70 &  14.36 $\pm$ 3.18 & 18.33 $\pm$ 6.17 & 11.05 $\pm$ 2.35 &   8.99 $\pm$ 3.19 & 15.09 $\pm$ 5.90 &  16.07 $\pm$ 4.53 & 25.51 $\pm$ 13.16 & 18.36 $\pm$ 10.36 &  13.54 $\pm$ 7.27 &  20.57 $\pm$ 6.59 &   7.29 $\pm$ 1.13 &  16.82 $\pm$ 4.63 \\
       hepatitis &  24.31 $\pm$ 2.23 &  27.70 $\pm$ 3.25 &  25.17 $\pm$ 5.46 & 33.91 $\pm$ 5.86 & 21.39 $\pm$ 6.04 & 30.36 $\pm$ 15.57 & 29.47 $\pm$ 2.74 & 38.88 $\pm$ 3.25 &  27.47 $\pm$ 8.72 & 22.49 $\pm$ 8.34 &  32.80 $\pm$ 4.31 &  46.70 $\pm$ 2.15 &  53.62 $\pm$ 2.63 &  21.49 $\pm$ 8.14 &  23.82 $\pm$ 2.31 &  16.49 $\pm$ 2.95 &  25.73 $\pm$ 7.50 \\
            http & 88.63 $\pm$ 15.26 &  35.59 $\pm$ 2.55 &   0.98 $\pm$ 0.69 & 49.99 $\pm$ 2.19 &  4.95 $\pm$ 2.10 &  46.43 $\pm$ 3.33 & 14.47 $\pm$ 0.73 & 28.02 $\pm$ 4.37 &   0.41 $\pm$ 0.07 &  4.69 $\pm$ 1.83 &  30.19 $\pm$ 3.04 &  19.28 $\pm$ 0.54 &  28.89 $\pm$ 1.60 & 29.53 $\pm$ 19.62 &   2.41 $\pm$ 0.99 & 64.22 $\pm$ 20.75 & 44.03 $\pm$ 16.11 \\
     internetads &  48.62 $\pm$ 4.28 &  29.09 $\pm$ 0.14 &  29.64 $\pm$ 0.09 & 27.56 $\pm$ 0.76 & 23.20 $\pm$ 1.20 &  29.65 $\pm$ 0.08 & 50.54 $\pm$ 0.20 & 50.47 $\pm$ 0.20 &  24.18 $\pm$ 3.77 & 18.19 $\pm$ 1.90 &  52.27 $\pm$ 0.30 &   7.12 $\pm$ 0.91 &  12.17 $\pm$ 3.04 &  27.53 $\pm$ 2.43 &  29.01 $\pm$ 0.46 &  29.46 $\pm$ 0.05 &  30.18 $\pm$ 2.54 \\
      ionosphere &  77.92 $\pm$ 2.90 &  82.91 $\pm$ 0.84 &  91.09 $\pm$ 0.79 & 72.08 $\pm$ 2.42 & 80.67 $\pm$ 4.01 &  88.10 $\pm$ 2.92 & 63.34 $\pm$ 2.30 & 66.28 $\pm$ 3.20 &  74.07 $\pm$ 1.71 & 82.05 $\pm$ 3.00 &  35.26 $\pm$ 2.03 &  62.08 $\pm$ 1.12 &  66.43 $\pm$ 0.98 & 60.98 $\pm$ 10.63 &  92.04 $\pm$ 1.18 &  63.29 $\pm$ 3.49 &  87.96 $\pm$ 2.24 \\
         landsat &  19.37 $\pm$ 0.64 &  17.50 $\pm$ 0.05 &  25.75 $\pm$ 0.22 & 16.33 $\pm$ 0.13 & 24.99 $\pm$ 0.55 &  21.23 $\pm$ 1.78 & 16.37 $\pm$ 0.05 & 17.60 $\pm$ 0.05 &  18.29 $\pm$ 3.54 & 24.63 $\pm$ 0.49 &  23.07 $\pm$ 0.25 &  24.53 $\pm$ 7.77 &  19.14 $\pm$ 4.58 &  20.27 $\pm$ 3.23 &  25.45 $\pm$ 0.27 &  19.99 $\pm$ 0.35 &  22.34 $\pm$ 0.89 \\
            aloi &   3.39 $\pm$ 0.03 &   3.92 $\pm$ 0.14 &   4.76 $\pm$ 0.02 &  3.72 $\pm$ 0.03 &  9.69 $\pm$ 0.28 &   3.74 $\pm$ 0.07 &  3.29 $\pm$ 0.00 &  3.13 $\pm$ 0.00 &   3.27 $\pm$ 0.29 & 10.36 $\pm$ 0.45 &   3.38 $\pm$ 0.03 &  37.03 $\pm$ 2.19 &  41.98 $\pm$ 2.05 &   3.95 $\pm$ 0.11 &   5.55 $\pm$ 0.02 &   3.59 $\pm$ 0.02 &   3.28 $\pm$ 0.07 \\
          letter &   8.59 $\pm$ 0.16 &  11.27 $\pm$ 0.27 &  20.31 $\pm$ 0.72 &  7.62 $\pm$ 0.12 & 43.32 $\pm$ 2.85 &  16.64 $\pm$ 1.01 &  7.71 $\pm$ 0.07 &  6.84 $\pm$ 0.03 &   8.26 $\pm$ 1.17 & 44.53 $\pm$ 3.22 &   7.79 $\pm$ 0.21 & 100.00 $\pm$ 0.00 &  99.54 $\pm$ 1.04 &  18.09 $\pm$ 2.29 &  25.53 $\pm$ 1.41 &  36.69 $\pm$ 1.64 &  25.65 $\pm$ 1.60 \\
    lymphography &  97.22 $\pm$ 1.71 &  88.48 $\pm$ 6.11 &  89.44 $\pm$ 6.58 & 93.51 $\pm$ 4.78 & 13.52 $\pm$ 9.57 &  91.49 $\pm$ 6.57 & 89.39 $\pm$ 2.04 & 90.69 $\pm$ 2.49 & 49.05 $\pm$ 38.67 &  9.00 $\pm$ 7.08 &  91.91 $\pm$ 3.02 &   3.00 $\pm$ 0.19 &   3.01 $\pm$ 0.08 & 38.80 $\pm$ 19.67 &  80.51 $\pm$ 9.21 & 73.10 $\pm$ 20.13 & 38.13 $\pm$ 14.88 \\
     magic.gamma &  63.77 $\pm$ 0.37 &  62.51 $\pm$ 0.10 &  72.35 $\pm$ 0.14 & 58.88 $\pm$ 0.09 & 51.98 $\pm$ 0.44 &  66.61 $\pm$ 0.05 & 53.34 $\pm$ 0.05 & 58.80 $\pm$ 0.04 &  57.87 $\pm$ 1.31 & 53.87 $\pm$ 0.79 &  61.74 $\pm$ 0.15 &  33.44 $\pm$ 4.76 &  12.15 $\pm$ 2.46 &  65.74 $\pm$ 2.99 &  72.98 $\pm$ 0.13 &  65.14 $\pm$ 1.91 &  66.40 $\pm$ 0.97 \\
     mammography &  21.78 $\pm$ 3.74 &  18.69 $\pm$ 0.74 &  18.06 $\pm$ 0.92 & 20.44 $\pm$ 1.39 &  8.48 $\pm$ 0.72 &  13.95 $\pm$ 2.79 & 43.54 $\pm$ 0.39 & 43.02 $\pm$ 0.41 &  21.76 $\pm$ 4.58 &  7.01 $\pm$ 0.99 &  13.24 $\pm$ 1.35 &  19.44 $\pm$ 2.11 &  19.48 $\pm$ 4.25 &   8.20 $\pm$ 2.52 &  17.45 $\pm$ 0.99 &   9.89 $\pm$ 2.26 &  17.02 $\pm$ 1.42 \\
           mnist &  29.03 $\pm$ 4.81 &  38.54 $\pm$ 0.33 &  40.87 $\pm$ 0.50 & 38.14 $\pm$ 0.94 & 23.34 $\pm$ 1.51 &  38.61 $\pm$ 1.75 &  9.21 $\pm$ 0.00 &  9.21 $\pm$ 0.00 &  16.97 $\pm$ 6.79 & 24.11 $\pm$ 1.05 &  10.91 $\pm$ 0.12 &  67.00 $\pm$ 1.18 &  59.18 $\pm$ 1.53 &  27.63 $\pm$ 5.40 &  39.99 $\pm$ 0.75 &  37.38 $\pm$ 1.09 &  36.76 $\pm$ 2.05 \\
            musk &  94.47 $\pm$ 9.05 & 100.00 $\pm$ 0.00 & 70.81 $\pm$ 10.35 & 99.95 $\pm$ 0.02 & 11.77 $\pm$ 5.21 & 100.00 $\pm$ 0.00 & 47.47 $\pm$ 1.53 & 36.91 $\pm$ 4.05 & 84.15 $\pm$ 17.56 & 13.95 $\pm$ 7.85 &  99.87 $\pm$ 0.08 &  93.40 $\pm$ 4.12 &  96.91 $\pm$ 3.39 &  13.68 $\pm$ 4.74 &  43.36 $\pm$ 3.14 &  98.38 $\pm$ 1.16 & 55.30 $\pm$ 21.58 \\
       optdigits &   4.61 $\pm$ 0.81 &   2.65 $\pm$ 0.08 &   2.18 $\pm$ 0.09 &  2.70 $\pm$ 0.03 &  3.53 $\pm$ 0.69 &   5.92 $\pm$ 0.26 &  2.88 $\pm$ 0.00 &  2.88 $\pm$ 0.00 &   2.90 $\pm$ 0.95 &  3.62 $\pm$ 0.78 &  19.18 $\pm$ 1.06 &  96.77 $\pm$ 0.64 &  95.43 $\pm$ 0.80 &   2.82 $\pm$ 0.35 &   2.14 $\pm$ 0.09 &   2.24 $\pm$ 0.14 &   2.75 $\pm$ 0.38 \\
      pageblocks &  46.37 $\pm$ 1.42 &  53.07 $\pm$ 0.57 &  55.58 $\pm$ 0.65 & 52.46 $\pm$ 1.66 & 29.16 $\pm$ 2.08 &  54.67 $\pm$ 5.46 & 51.96 $\pm$ 0.39 & 37.03 $\pm$ 0.41 &  40.95 $\pm$ 8.60 & 34.10 $\pm$ 2.79 &  31.88 $\pm$ 4.58 &  23.05 $\pm$ 0.09 &  31.65 $\pm$ 0.14 &  50.72 $\pm$ 9.01 &  52.96 $\pm$ 0.97 &  49.27 $\pm$ 2.07 &  55.49 $\pm$ 4.05 \\
       pendigits &  26.01 $\pm$ 4.72 &  22.57 $\pm$ 1.29 &   9.95 $\pm$ 2.61 & 21.86 $\pm$ 0.32 &  4.01 $\pm$ 0.53 & 19.17 $\pm$ 10.42 & 26.96 $\pm$ 0.57 & 17.71 $\pm$ 1.05 &  18.56 $\pm$ 5.48 &  4.83 $\pm$ 0.78 &  24.73 $\pm$ 0.80 &   0.41 $\pm$ 0.13 &   0.59 $\pm$ 0.22 &   4.40 $\pm$ 0.91 &   8.87 $\pm$ 1.47 &   5.61 $\pm$ 0.64 &   4.36 $\pm$ 1.06 \\
            pima &  50.96 $\pm$ 4.11 &  47.74 $\pm$ 2.79 &  52.99 $\pm$ 3.09 & 49.19 $\pm$ 4.08 & 40.63 $\pm$ 2.05 &  48.38 $\pm$ 3.73 & 48.38 $\pm$ 2.46 & 53.62 $\pm$ 2.38 &  40.39 $\pm$ 5.19 & 41.22 $\pm$ 2.23 &  57.73 $\pm$ 2.72 &   3.62 $\pm$ 1.94 &   3.91 $\pm$ 1.22 &  43.74 $\pm$ 3.29 &  52.83 $\pm$ 2.87 &  40.02 $\pm$ 2.72 &  44.68 $\pm$ 2.50 \\
      annthyroid &  31.23 $\pm$ 3.56 &  18.75 $\pm$ 0.28 &  22.41 $\pm$ 0.47 & 19.55 $\pm$ 1.07 & 16.33 $\pm$ 0.53 &  16.94 $\pm$ 0.78 & 27.21 $\pm$ 0.44 & 17.43 $\pm$ 0.19 &   9.80 $\pm$ 2.71 & 20.55 $\pm$ 4.61 &  22.79 $\pm$ 0.86 &  30.35 $\pm$ 3.41 &  41.37 $\pm$ 5.07 &  38.03 $\pm$ 6.20 &  22.82 $\pm$ 0.34 &  29.74 $\pm$ 2.36 &  67.01 $\pm$ 0.84 \\
       satellite &  64.88 $\pm$ 1.51 &  65.44 $\pm$ 0.16 &  58.16 $\pm$ 0.35 & 60.61 $\pm$ 0.17 & 38.10 $\pm$ 0.70 &  65.64 $\pm$ 6.27 & 52.62 $\pm$ 0.10 & 57.04 $\pm$ 0.08 &  61.27 $\pm$ 4.30 & 37.77 $\pm$ 0.72 &  68.78 $\pm$ 0.47 &   9.70 $\pm$ 0.70 &   9.39 $\pm$ 1.14 &  37.96 $\pm$ 2.66 &  56.29 $\pm$ 0.46 &  66.16 $\pm$ 0.76 &  52.91 $\pm$ 3.46 \\
      satimage-2 &  91.75 $\pm$ 0.85 &  96.53 $\pm$ 0.02 & 68.98 $\pm$ 15.78 & 87.19 $\pm$ 0.10 &  4.08 $\pm$ 2.50 &  97.21 $\pm$ 0.03 & 66.62 $\pm$ 1.58 & 79.70 $\pm$ 0.94 &  85.74 $\pm$ 7.48 &  4.23 $\pm$ 2.71 &  76.00 $\pm$ 1.14 &  51.57 $\pm$ 2.15 &  53.07 $\pm$ 2.37 &   9.52 $\pm$ 5.66 &  50.73 $\pm$ 8.98 &  78.25 $\pm$ 5.90 &  13.84 $\pm$ 3.38 \\
         shuttle &  97.62 $\pm$ 0.41 &  90.72 $\pm$ 0.06 &  19.31 $\pm$ 0.46 & 91.33 $\pm$ 0.15 & 10.93 $\pm$ 0.17 &  18.38 $\pm$ 2.14 & 90.50 $\pm$ 0.14 & 96.22 $\pm$ 0.17 & 16.83 $\pm$ 19.06 &  8.08 $\pm$ 1.88 &  96.47 $\pm$ 0.15 &   7.11 $\pm$ 1.21 & 41.90 $\pm$ 10.47 & 24.72 $\pm$ 13.77 &  18.65 $\pm$ 0.26 &  77.88 $\pm$ 7.67 &  62.60 $\pm$ 9.55 \\
            skin &  25.36 $\pm$ 0.43 &  22.01 $\pm$ 0.19 &  29.00 $\pm$ 0.18 & 17.24 $\pm$ 0.15 & 22.10 $\pm$ 0.17 &  28.86 $\pm$ 3.15 & 18.27 $\pm$ 0.10 & 17.86 $\pm$ 0.09 &  18.03 $\pm$ 0.48 & 20.68 $\pm$ 0.24 &  23.20 $\pm$ 0.19 & 84.00 $\pm$ 15.88 &  66.86 $\pm$ 6.23 &  31.57 $\pm$ 3.14 &  28.99 $\pm$ 0.20 &  17.54 $\pm$ 0.58 &  30.24 $\pm$ 1.74 \\
            smtp &   0.53 $\pm$ 0.08 &  38.25 $\pm$ 8.36 &  41.54 $\pm$ 5.59 & 38.24 $\pm$ 5.87 &  2.23 $\pm$ 1.39 &  40.32 $\pm$ 5.33 & 58.85 $\pm$ 4.72 &  0.50 $\pm$ 0.05 & 31.21 $\pm$ 10.41 &  0.13 $\pm$ 0.02 &   0.50 $\pm$ 0.05 &  31.40 $\pm$ 2.19 &  31.59 $\pm$ 2.30 &   1.16 $\pm$ 2.20 &  41.07 $\pm$ 5.45 &  50.23 $\pm$ 9.75 &  42.15 $\pm$ 3.73 \\
        spambase &  48.75 $\pm$ 1.64 &  40.21 $\pm$ 0.07 &  41.53 $\pm$ 0.17 & 40.93 $\pm$ 0.51 & 35.95 $\pm$ 0.33 &  40.23 $\pm$ 0.63 & 51.82 $\pm$ 0.17 & 54.37 $\pm$ 0.16 &  38.65 $\pm$ 5.96 & 34.39 $\pm$ 0.60 &  51.77 $\pm$ 1.22 &  98.92 $\pm$ 0.47 &  98.14 $\pm$ 1.41 &  39.86 $\pm$ 3.19 &  40.69 $\pm$ 0.22 &  38.37 $\pm$ 0.71 &  40.04 $\pm$ 1.52 \\
          speech &   2.05 $\pm$ 0.34 &   1.85 $\pm$ 0.03 &   1.85 $\pm$ 0.02 &  1.84 $\pm$ 0.00 &  2.16 $\pm$ 0.15 &   1.87 $\pm$ 0.02 &  1.96 $\pm$ 0.01 &  1.88 $\pm$ 0.07 &   1.61 $\pm$ 0.20 &  2.18 $\pm$ 0.15 &   2.29 $\pm$ 0.14 &  30.27 $\pm$ 0.43 &  31.64 $\pm$ 2.19 &   1.90 $\pm$ 0.21 &   1.88 $\pm$ 0.15 &   2.04 $\pm$ 0.43 &   2.00 $\pm$ 0.33 \\
          stamps &  34.72 $\pm$ 4.50 &  31.76 $\pm$ 4.47 &  31.69 $\pm$ 3.92 & 36.40 $\pm$ 6.13 & 15.27 $\pm$ 4.40 &  21.06 $\pm$ 2.78 & 32.35 $\pm$ 3.22 & 39.78 $\pm$ 4.75 &  27.97 $\pm$ 8.26 & 14.26 $\pm$ 4.10 &  33.18 $\pm$ 3.90 &  71.58 $\pm$ 6.37 &  46.30 $\pm$ 6.00 & 23.48 $\pm$ 11.01 &  27.25 $\pm$ 4.34 &  14.26 $\pm$ 4.14 &  22.63 $\pm$ 4.68 \\
         thyroid &  56.22 $\pm$ 8.46 &  32.89 $\pm$ 2.07 &  39.22 $\pm$ 2.16 & 35.57 $\pm$ 3.87 &  7.73 $\pm$ 2.47 &  27.17 $\pm$ 0.59 & 47.18 $\pm$ 2.25 & 17.94 $\pm$ 0.90 &  18.90 $\pm$ 8.97 &  6.93 $\pm$ 2.88 &  50.12 $\pm$ 2.65 &  12.82 $\pm$ 0.36 &   9.26 $\pm$ 0.56 &  11.80 $\pm$ 6.26 &  35.98 $\pm$ 2.15 &  32.48 $\pm$ 4.30 &  70.49 $\pm$ 4.14 \\
       vertebral &   9.68 $\pm$ 1.00 &  10.68 $\pm$ 1.32 &   9.51 $\pm$ 1.18 &  9.93 $\pm$ 0.89 & 12.95 $\pm$ 3.05 &  12.34 $\pm$ 0.98 & 10.97 $\pm$ 0.72 &  8.50 $\pm$ 1.20 &   8.88 $\pm$ 1.14 & 12.37 $\pm$ 3.08 &   9.12 $\pm$ 1.03 &   8.42 $\pm$ 0.10 &   8.43 $\pm$ 0.11 &  13.31 $\pm$ 5.54 &   9.82 $\pm$ 1.03 &  14.97 $\pm$ 4.55 &  11.92 $\pm$ 1.70 \\
        backdoor &   4.54 $\pm$ 0.72 &  53.38 $\pm$ 1.03 &  47.92 $\pm$ 1.45 & 53.14 $\pm$ 1.28 & 35.80 $\pm$ 2.43 &  54.65 $\pm$ 1.42 &  2.48 $\pm$ 0.05 &  2.48 $\pm$ 0.05 &  10.08 $\pm$ 7.77 & 21.68 $\pm$ 6.06 &   5.15 $\pm$ 0.09 & 66.40 $\pm$ 20.45 & 58.08 $\pm$ 27.93 &  43.84 $\pm$ 3.25 &  47.29 $\pm$ 1.45 &  52.01 $\pm$ 0.96 &  48.07 $\pm$ 1.28 \\
          vowels &  16.23 $\pm$ 6.18 &  19.58 $\pm$ 1.16 &  44.32 $\pm$ 0.55 &  6.87 $\pm$ 0.26 & 32.58 $\pm$ 5.97 &  16.61 $\pm$ 1.03 &  8.28 $\pm$ 0.54 &  3.43 $\pm$ 0.05 &  12.72 $\pm$ 3.85 & 31.42 $\pm$ 8.14 &   7.83 $\pm$ 0.89 &  46.75 $\pm$ 2.79 &  49.03 $\pm$ 6.90 &  16.57 $\pm$ 4.43 &  50.44 $\pm$ 3.18 &  31.06 $\pm$ 4.37 & 41.70 $\pm$ 12.32 \\
        waveform &   5.63 $\pm$ 0.92 &   5.23 $\pm$ 0.11 &  13.28 $\pm$ 0.76 &  4.41 $\pm$ 0.02 &  7.09 $\pm$ 0.90 &  12.23 $\pm$ 1.76 &  4.04 $\pm$ 0.03 &  5.69 $\pm$ 0.14 &   4.02 $\pm$ 0.78 &  7.84 $\pm$ 1.43 &   4.83 $\pm$ 0.11 &  74.55 $\pm$ 3.91 &  74.10 $\pm$ 4.67 &   3.73 $\pm$ 0.96 &  10.93 $\pm$ 1.18 &   4.98 $\pm$ 0.61 &   4.28 $\pm$ 0.59 \\
             wbc &  94.84 $\pm$ 2.02 & 81.27 $\pm$ 11.50 &  74.27 $\pm$ 6.66 & 91.33 $\pm$ 4.96 &  7.72 $\pm$ 1.47 & 69.07 $\pm$ 11.79 & 88.19 $\pm$ 2.42 & 88.33 $\pm$ 2.34 &  89.76 $\pm$ 2.29 &  3.72 $\pm$ 0.48 &  72.83 $\pm$ 6.35 &   3.42 $\pm$ 0.22 &   3.40 $\pm$ 0.22 & 34.84 $\pm$ 17.79 & 72.17 $\pm$ 13.60 &  75.78 $\pm$ 9.25 &  19.35 $\pm$ 3.20 \\
            wdbc &  70.18 $\pm$ 4.66 &  53.89 $\pm$ 7.79 &  52.13 $\pm$ 4.05 & 61.28 $\pm$ 3.38 & 12.80 $\pm$ 7.89 &  68.81 $\pm$ 9.09 & 49.27 $\pm$ 4.01 & 76.04 $\pm$ 3.54 & 52.69 $\pm$ 13.22 & 15.45 $\pm$ 9.61 &  76.14 $\pm$ 4.83 & 57.00 $\pm$ 41.77 & 90.00 $\pm$ 22.36 &   7.41 $\pm$ 7.03 &  46.51 $\pm$ 7.89 & 48.27 $\pm$ 10.80 &  15.65 $\pm$ 7.73 \\
            wilt &   4.40 $\pm$ 0.25 &   3.54 $\pm$ 0.01 &   4.92 $\pm$ 0.07 &  3.22 $\pm$ 0.01 &  8.31 $\pm$ 0.34 &   4.01 $\pm$ 0.12 &  4.17 $\pm$ 0.00 &  3.70 $\pm$ 0.01 &   3.60 $\pm$ 0.48 &  8.05 $\pm$ 2.16 &   3.94 $\pm$ 0.15 &  51.32 $\pm$ 1.18 &  51.12 $\pm$ 4.69 &  21.14 $\pm$ 6.69 &   5.35 $\pm$ 0.07 &   7.62 $\pm$ 0.85 &  16.29 $\pm$ 1.47 \\
            wine &  20.69 $\pm$ 4.89 &  13.48 $\pm$ 2.11 &   8.05 $\pm$ 0.89 & 26.39 $\pm$ 5.02 &  6.42 $\pm$ 1.66 & 17.04 $\pm$ 22.72 & 19.45 $\pm$ 3.20 & 36.39 $\pm$ 6.24 &  24.99 $\pm$ 9.90 &  6.06 $\pm$ 0.50 & 41.21 $\pm$ 10.01 &  55.00 $\pm$ 3.12 &  51.96 $\pm$ 2.29 &   6.39 $\pm$ 1.93 &   7.37 $\pm$ 1.21 &   7.45 $\pm$ 2.14 &  10.27 $\pm$ 3.41 \\
            wpbc &  23.73 $\pm$ 1.92 &  22.15 $\pm$ 1.31 &  23.44 $\pm$ 1.40 & 22.86 $\pm$ 1.58 & 20.98 $\pm$ 1.73 &  22.74 $\pm$ 2.24 & 21.66 $\pm$ 1.22 & 23.37 $\pm$ 1.68 &  22.65 $\pm$ 1.74 & 20.57 $\pm$ 1.44 &  24.10 $\pm$ 1.66 &  42.16 $\pm$ 1.82 &  42.42 $\pm$ 1.09 &  23.80 $\pm$ 2.07 &  22.72 $\pm$ 1.72 &  23.80 $\pm$ 3.11 &  23.14 $\pm$ 3.11 \\
           yeast &  30.39 $\pm$ 0.49 &  30.33 $\pm$ 0.38 &  29.36 $\pm$ 0.48 & 30.17 $\pm$ 0.20 & 31.51 $\pm$ 0.78 &  31.39 $\pm$ 0.56 & 33.19 $\pm$ 0.18 & 30.79 $\pm$ 0.15 &  33.01 $\pm$ 2.79 & 32.55 $\pm$ 0.99 &  32.79 $\pm$ 0.50 &  37.40 $\pm$ 7.46 &  34.67 $\pm$ 9.43 &  30.64 $\pm$ 1.82 &  29.45 $\pm$ 0.66 &  32.04 $\pm$ 0.49 &  30.61 $\pm$ 1.70 \\
         breastw &  95.64 $\pm$ 1.34 &  89.69 $\pm$ 1.55 &  93.20 $\pm$ 1.85 & 94.55 $\pm$ 0.90 & 29.65 $\pm$ 2.09 &  88.99 $\pm$ 3.32 & 98.24 $\pm$ 0.39 & 98.87 $\pm$ 0.33 &  95.50 $\pm$ 3.15 & 28.44 $\pm$ 1.29 &  95.44 $\pm$ 1.00 &   7.20 $\pm$ 2.74 &   9.31 $\pm$ 2.00 &  77.03 $\pm$ 2.76 &  92.09 $\pm$ 1.62 &  53.68 $\pm$ 5.28 &  71.52 $\pm$ 3.80 \\
        campaign &  27.91 $\pm$ 1.24 &  28.33 $\pm$ 0.08 &  28.91 $\pm$ 0.14 & 28.40 $\pm$ 0.32 & 15.80 $\pm$ 0.13 &  28.68 $\pm$ 0.21 & 35.44 $\pm$ 0.07 & 36.84 $\pm$ 0.06 &  13.05 $\pm$ 4.47 & 14.51 $\pm$ 2.47 &  35.21 $\pm$ 0.32 &  85.00 $\pm$ 9.13 &  86.67 $\pm$ 7.45 &  23.68 $\pm$ 2.47 &  28.05 $\pm$ 0.19 &  29.90 $\pm$ 0.96 &  32.12 $\pm$ 1.10 \\
          cardio &  55.88 $\pm$ 4.43 &  53.57 $\pm$ 0.67 &  40.17 $\pm$ 1.51 & 60.87 $\pm$ 0.73 & 15.89 $\pm$ 1.81 &  48.23 $\pm$ 1.68 & 56.68 $\pm$ 0.74 & 57.59 $\pm$ 0.51 & 42.78 $\pm$ 10.47 & 16.09 $\pm$ 1.04 &  45.80 $\pm$ 0.86 & 77.33 $\pm$ 24.28 & 70.67 $\pm$ 17.97 &  18.35 $\pm$ 6.15 &  37.62 $\pm$ 0.74 &  27.84 $\pm$ 5.61 &  26.80 $\pm$ 1.87 \\
cardiotocography &  43.62 $\pm$ 2.11 &  40.83 $\pm$ 0.26 &  32.37 $\pm$ 0.29 & 46.20 $\pm$ 1.18 & 27.15 $\pm$ 0.91 &  33.53 $\pm$ 5.06 & 50.23 $\pm$ 0.37 & 40.29 $\pm$ 2.63 & 46.28 $\pm$ 12.59 & 27.64 $\pm$ 0.53 &  36.10 $\pm$ 0.67 &   3.74 $\pm$ 0.13 &   3.67 $\pm$ 0.11 &  25.00 $\pm$ 3.68 &  31.16 $\pm$ 0.49 &  33.84 $\pm$ 3.20 &  27.55 $\pm$ 1.23 \\
          celeba &   6.26 $\pm$ 0.41 &  10.28 $\pm$ 0.48 &   6.07 $\pm$ 0.27 & 11.19 $\pm$ 0.62 &  1.81 $\pm$ 0.02 &   6.88 $\pm$ 2.06 &  9.53 $\pm$ 0.55 &  9.28 $\pm$ 0.59 &   4.65 $\pm$ 3.19 &  2.37 $\pm$ 0.28 &   8.95 $\pm$ 0.56 &  25.77 $\pm$ 4.08 &  29.41 $\pm$ 6.38 &   5.77 $\pm$ 1.59 &   5.19 $\pm$ 0.23 &   9.25 $\pm$ 1.47 &   7.68 $\pm$ 0.83 \\
          census &   7.30 $\pm$ 0.49 &   8.52 $\pm$ 0.23 &   8.82 $\pm$ 0.09 &  8.66 $\pm$ 0.23 &  6.87 $\pm$ 0.23 &   8.75 $\pm$ 0.28 &  6.23 $\pm$ 0.16 &  6.23 $\pm$ 0.16 &   6.52 $\pm$ 2.72 &  6.11 $\pm$ 0.18 &   7.30 $\pm$ 0.19 &  46.56 $\pm$ 5.17 &  37.43 $\pm$ 5.18 &   8.34 $\pm$ 0.92 &   9.00 $\pm$ 0.09 &   8.56 $\pm$ 0.23 &   8.09 $\pm$ 0.30 \\
         Average &             37.47 &             36.03 &             33.83 &            36.60 &            18.92 &             35.20 &            34.34 &            33.34 &             28.97 &            18.55 &             34.40 &             39.60 &             39.95 &             23.62 &             32.24 &             33.25 &             31.89 \\
\midrule
 20news & 6.24 $\pm$ 0.36 & 6.38 $\pm$ 0.44 & 6.90 $\pm$ 0.36 & 6.24 $\pm$ 0.23 &  8.75 $\pm$ 0.88 & 6.66 $\pm$ 0.42 & 6.17 $\pm$ 0.12 & 6.09 $\pm$ 0.29 & 6.23 $\pm$ 1.07 &  8.71 $\pm$ 0.87 & 6.05 $\pm$ 0.21 & 6.01 $\pm$ 1.38 & 7.16 $\pm$ 0.49 & 6.25 $\pm$ 0.21 & 6.84 $\pm$ 0.87 & 6.86 $\pm$ 3.01 & 8.24 $\pm$ 4.48 \\
 agnews & 6.36 $\pm$ 0.22 & 6.78 $\pm$ 0.07 & 8.16 $\pm$ 0.03 & 6.11 $\pm$ 0.02 & 12.47 $\pm$ 0.61 & 7.24 $\pm$ 0.07 & 5.76 $\pm$ 0.01 & 5.85 $\pm$ 0.01 & 6.42 $\pm$ 0.58 & 12.51 $\pm$ 0.62 & 5.87 $\pm$ 0.01 & 6.26 $\pm$ 1.28 & 8.45 $\pm$ 0.06 & 6.17 $\pm$ 0.01 & 7.55 $\pm$ 1.04 & 7.77 $\pm$ 2.43 & 6.29 $\pm$ 1.21 \\
 amazon & 5.83 $\pm$ 0.09 & 5.89 $\pm$ 0.01 & 6.22 $\pm$ 0.01 & 5.69 $\pm$ 0.02 &  5.79 $\pm$ 0.13 & 6.06 $\pm$ 0.06 & 5.50 $\pm$ 0.01 & 5.96 $\pm$ 0.01 & 5.44 $\pm$ 0.43 &  5.80 $\pm$ 0.11 & 5.87 $\pm$ 0.02 & 5.49 $\pm$ 0.40 & 6.22 $\pm$ 0.08 & 5.71 $\pm$ 0.01 & 5.72 $\pm$ 0.54 & 5.80 $\pm$ 0.34 & 5.99 $\pm$ 0.61 \\
   imdb & 4.68 $\pm$ 0.04 & 4.69 $\pm$ 0.09 & 4.67 $\pm$ 0.02 & 4.59 $\pm$ 0.01 &  4.88 $\pm$ 0.06 & 4.74 $\pm$ 0.03 & 4.48 $\pm$ 0.01 & 4.96 $\pm$ 0.03 & 4.58 $\pm$ 0.28 &  4.87 $\pm$ 0.04 & 4.74 $\pm$ 0.01 & 4.74 $\pm$ 0.52 & 4.69 $\pm$ 0.03 & 4.59 $\pm$ 0.01 & 4.67 $\pm$ 0.32 & 4.77 $\pm$ 0.34 & 4.80 $\pm$ 0.34 \\
   yelp & 6.96 $\pm$ 0.05 & 7.29 $\pm$ 0.01 & 8.27 $\pm$ 0.05 & 6.88 $\pm$ 0.03 &  8.52 $\pm$ 0.18 & 7.31 $\pm$ 0.17 & 6.47 $\pm$ 0.01 & 7.24 $\pm$ 0.03 & 6.67 $\pm$ 0.60 &  8.52 $\pm$ 0.19 & 7.04 $\pm$ 0.01 & 5.42 $\pm$ 0.93 & 8.50 $\pm$ 0.13 & 6.92 $\pm$ 0.01 & 6.59 $\pm$ 0.78 & 7.56 $\pm$ 0.39 & 6.84 $\pm$ 0.92 \\
Average &            6.01 &            6.21 &            6.84 &            5.90 &             8.08 &            6.40 &            5.68 &            6.02 &            5.87 &             8.08 &            5.91 &            5.58 &            7.00 &            5.93 &            6.27 &            6.55 &            6.43 \\
\bottomrule
\end{tabular}
    \end{adjustbox}
\end{table}

\newpage

\subsection{Tennessee Eastman Process}

The \gls{te} process was first proposed by~\cite{downs1993plant}, as a large-scale chemical process for benchmarking control algorithms, as well as fault detection and diagnosis methods. We follow the description of~\cite{reinartz2021extended}. This chemical process consists of 4 chemical reactions, for the production of 2 liquid liquid components, denoted $G$ and $E$, and 4 gaseous reactants, denoted $A$, $B$, $C$ and $D$. The reactions are as follows,
\begin{equation}
    \begin{aligned}
        A(\text{g}) + C(\text{g}) + D(\text{g}) &\rightarrow G(\text{liq})&\text{Product 1,}\\
        A(\text{g}) + C(\text{g}) + E(\text{g}) &\rightarrow H(\text{liq})&\text{Product 2,}\\
        A(\text{g}) + E(\text{g}) &\rightarrow F(\text{liq})&\text{Byproduct,}\\
        3D(\text{g}) &\rightarrow 2F(\text{liq})&\text{Byproduct.}
    \end{aligned}\label{eq:tep_equations}
\end{equation}
The chemical plant is composed by 5 processing units: reactor, product condenser, vapor-liquid separator, recycle compressor and product stripper. We refer readers to~\cite{reinartz2021extended} and ~\cite{montesuma2024benchmarking} for further details on how these components work together in the system. As described by~\cite{downs1993plant}, based on equation~\ref{eq:tep_equations}, there are 6 different \emph{modes of operation}, corresponding to different product mass ratio and product rate. We detail these in Table~\ref{tab:tep_operation_modes}.

\begin{table}[ht]
    \centering
    \caption{\gls{te} process operation modes in terms of $G/H$ mass ratio and production rate.}
    \begin{tabular}{>{\centering}p{0.2\linewidth}>{\centering}p{0.2\linewidth}p{0.5\linewidth}}
        \toprule
         Mode & Mass Ratio & Production rate \\
         \midrule
         1 & 50/50 & 7038 kg h$^{-1}$ G and 7038 kg h$^{-1}$ H\\
         2 & 10/90 & 1408 kg h$^{-1}$ G and 12,669 kg h$^{-1}$ H\\
         3 & 90/10 & 10,000 kg h$^{-1}$ G and 1111 kg h$^{-1}$ H\\
         4 & 50/50 & maximum production rate\\
         5 & 10/90 & maximum production rate\\
         6 & 90/10 & maximum production rate\\
         \bottomrule
    \end{tabular}
    \label{tab:tep_operation_modes}
\end{table}

From the chemical plant that performs the equations in~\ref{eq:tep_equations}, there are 54 variables divided into measured (XME) and manipulated (XMV). We show these in Table~\ref{tab:tep_variables}. Out of the 54 variables, we do as in~\cite{reinartz2021extended}, that is, we use a subset of 34 variables, boldened in Table~\ref{tab:tep_variables}. In the simulations of~\cite{reinartz2021extended}, the \gls{te} process is simulated for 100 hours, with a sampling rate of $3$ minutes. As a result, each simulation is represented as a time series $\mathbf{x} \in \mathbb{R}^{600 \times 34}$, where $\mathbf{x}(t) \in \mathbb{R}^{34}$.

\begin{table}[ht]
    \centering
    \caption{Process variables used in the \gls{te} process, divided into measurements (XME) and manipulated (XMV). Bold rows indicate that the variable is used in our experiments.}
    \resizebox{\linewidth}{!}{
\begin{tabular}{cccccccc}
    \toprule
    Variable         & Description                                        & Variable         & Description                                           & Variable         & Description                     & Variable         & Description                     \\ 
    \midrule
    \textbf{XME(1)}  & \textbf{A Feed (kscmh)}                                     & \textbf{XME(15)} & \textbf{Stripper Level (\%)}                                   & XME(29) & Component A in Purge (mol \%)   & \textbf{XMV(2)}  & \textbf{E Feed (\%)} \\
    \textbf{XME(2)}  & \textbf{D Feed (kg/h)}                                      & \textbf{XME(16)} & \textbf{Stripper Pressure (kPa gauge)}                         & XME(30) & Component B in Purge (mol \%)   & \textbf{XMV(3)}  & \textbf{A Feed (\%)} \\
    \textbf{XME(3)}  & \textbf{E Feed (kg/h)}                                      & \textbf{XME(17)} & \textbf{Stripper Underflow (m$^{3}$/h)}                        & XME(31) & Component C in Purge (mol \%)   & \textbf{XMV(4)}  & \textbf{A \& C Feed (\%)}        \\
    \textbf{XME(4)}  & \textbf{A \& C Feed (kg/h)}                                 & \textbf{XME(18)} & \textbf{Stripper Temp (\textdegree C)}          & XME(32) & Component D in Purge (mol \%)   & \textbf{XMV(5)}  & \textbf{Compressor recycle valve (\%)}   \\
    \textbf{XME(5)}  & \textbf{Recycle Flow (kscmh)}                               & \textbf{XME(19)} & \textbf{Stripper Steam Flow (kg/h)}                            & XME(33) & Component E in Purge (mol \%)   & \textbf{XMV(6)}  & \textbf{Purge valve (\%)}                \\
    \textbf{XME(6)}  & \textbf{Reactor Feed rate (kscmh)}                          & \textbf{XME(20)} & \textbf{Compressor Work (kW)}                                  & XME(34) & Component F in Purge (mol \%)   & \textbf{XMV(7)}  & \textbf{Separator liquid flow (\%)}      \\
    \textbf{XME(7)}  & \textbf{Reactor Pressure (kscmh)}                           & \textbf{XME(21)} & \textbf{Reactor Coolant Temp (\textdegree C)}   & XME(35) & Component G in Purge (mol \%)   & \textbf{XMV(8)}  & \textbf{Stripper liquid flow (\%)}       \\
    \textbf{XME(8)}  & \textbf{Reactor Level (\%)}                                 & \textbf{XME(22)} & \textbf{Separator Coolant Temp (\textdegree C)} & XME(36) & Component H in Purge (mol \%)   & \textbf{XMV(9)}  & \textbf{Stripper steam valve (\%)}       \\
    \textbf{XME(9)}  & \textbf{Reactor Temperature (\textdegree C)} & XME(23) & Component A to Reactor (mol \%)                       & XME(37) & Component D in Product (mol \%) & \textbf{XMV(10)} & \textbf{Reactor coolant (\%)}            \\
    \textbf{XME(10)} & \textbf{Purge Rate (kscmh)}                                 & XME(24) & Component B to Reactor (mol \%)                       & XME(38) & Component E in Product (mol \%) & \textbf{XMV(11)} & \textbf{Condenser Coolant (\%)}          \\
    \textbf{XME(11)} & \textbf{Product Sep Temp (\textdegree C)}    & XME(25) & Component C to Reactor (mol \%)                       & XME(39) & Component F in Product (mol \%) & \textbf{XMV(12)} & \textbf{Agitator Speed (\%)}             \\
    \textbf{XME(12)} & \textbf{Product Sep Level (\%)}                             & XME(26)          & Component D to Reactor (mol \%)                       & XME(40) & Component G in Product (mol \%)                     &         &                                 \\
    \textbf{XME(13)} & \textbf{Product Sep Pressure (kPa gauge)}                   & XME(27) & Component E to Reactor (mol \%)                       & XME(41) & Component H in Product (mol \%)                     &                  &                                 \\
    \textbf{XME(14)} & \textbf{Product Sep Underflow (m$^{3}$/h)}                  & XME(28) & Component F to Reactor (mol \%)                       & \textbf{XMV(1)}  & \textbf{D Feed (\%)}                     &                  &                           \\
    \bottomrule     
\end{tabular}
    }
    \label{tab:tep_variables}
\end{table}

The simulations in~\cite{reinartz2021extended} are faulty or normal. For faulty simulations, a different fault type is introduced between $28$ possible categories at $t_0 = 30h$. We refer readers to the original paper for more information about the different categories. In this paper, we construct $6$ different \gls{ad} datasets, one for each mode of operation. On each mode of operation, we use $100$ normal simulations, and $1$ faulty simulation for each fault type. This process results in $128$ time series per mode of operation.

For applying \gls{ad} algorithm, we further pre-process the time series obtained from the simulation of the \gls{te} process. We decompose each time series in windows of $20$ hours, i.e., $60$ samples. Within each window, we compute the mean and standard-deviation per sensor, that is,
\begin{align*}
    \mu_{tj} = \dfrac{1}{60}\sum_{i=t}^{t+60}x_{ij}\text{, and, }\sigma_{tj} = \sqrt{\dfrac{1}{59}\sum_{i=t}^{t+60}(x_{ij}-\mu_{tj})^{2}},
\end{align*}
where $t$ is the index of the window. We use the concatenation $(\mu_{t},\sigma_{t}) \in \mathbb{R}^{68}$ as the representation for the $t-$th window. As a result of this process, we extract 30 windows for each time series. This results on $6$ datasets with $128 \times 30 = 3840$ samples each. We then proceed to apply \gls{ad} algorithms at the level of windows. Here, we note an important detail. For \gls{ot} algorithms, we compare samples based on the Euclidean distance. Computing this distance over $\mathbf{x} = (\mu,\sigma)$ and $\mathbf{y} = (\mu',\sigma')$ boils down to,
\begin{align*}
    \lVert \mathbf{x}_{i} - \mathbf{x}_{j} \rVert_{2}^{2} = \lVert \mu - \mu' \rVert_{2} + \lVert \sigma - \sigma' \rVert_{2},
\end{align*}
which is the squared $2-$Wasserstein distance $W_{2}(P, Q)^{2}$ between axis-aligned Gaussians $P = \mathcal{N}(\mu,\sigma\mathbf{I})$ and $Q = \mathcal{N}(\mu',\sigma'\mathbf{I})$.

\subsubsection{Results per anomaly percentage}

\begin{figure}[ht]
    \centering
    \begin{subfigure}{0.31\linewidth}
        \centering
        \includegraphics[width=\linewidth]{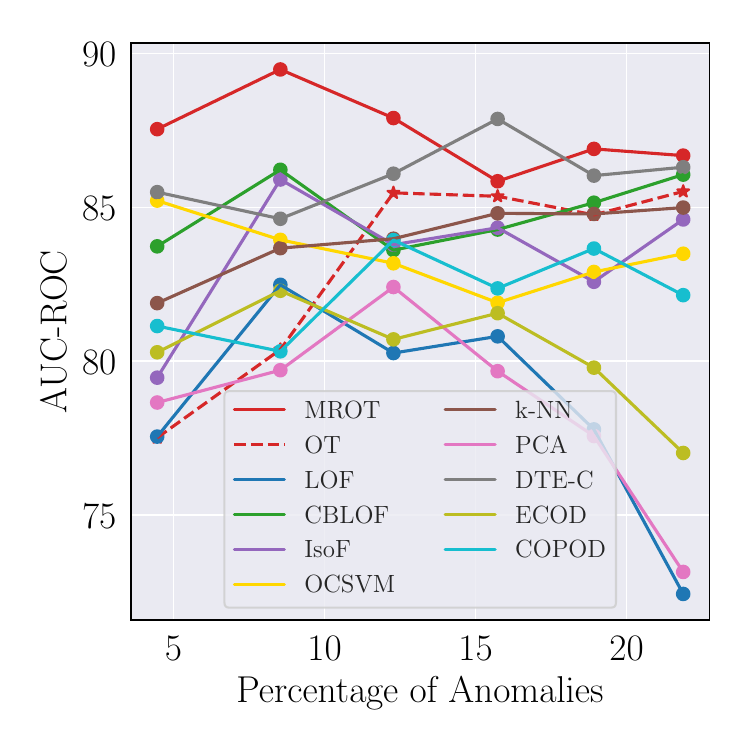}
        \caption{Mode 1.}
    \end{subfigure}\hfill
    \begin{subfigure}{0.31\linewidth}
        \centering
        \includegraphics[width=\linewidth]{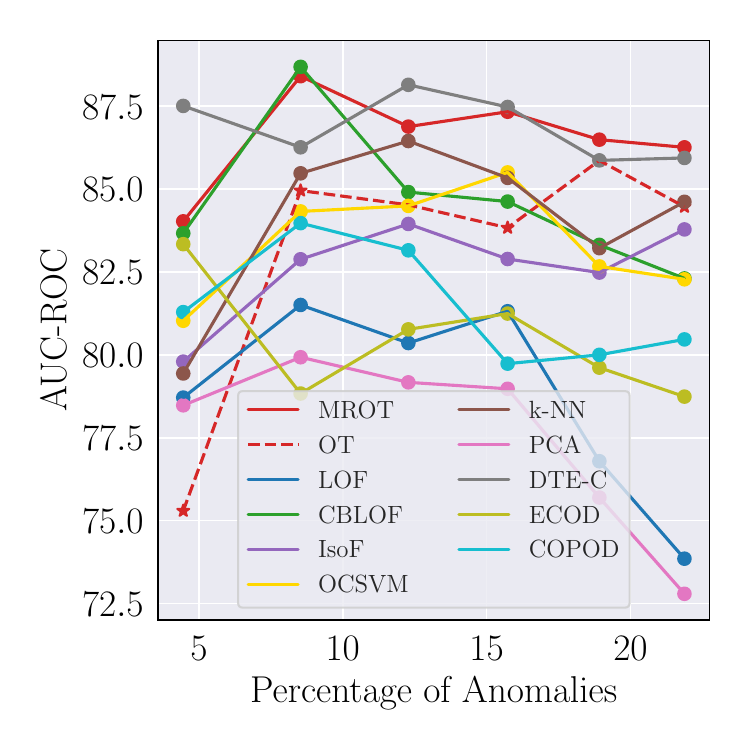}
        \caption{Mode 2.}
    \end{subfigure}\hfill
    \begin{subfigure}{0.31\linewidth}
        \centering
        \includegraphics[width=\linewidth]{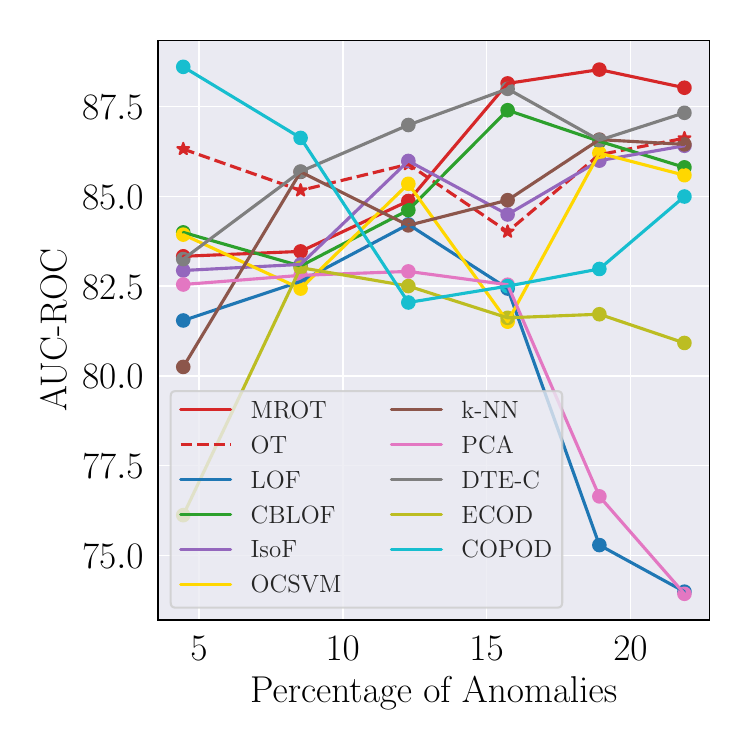}
        \caption{Mode 3.}
    \end{subfigure}\hfill
    \begin{subfigure}{0.31\linewidth}
        \centering
        \includegraphics[width=\linewidth]{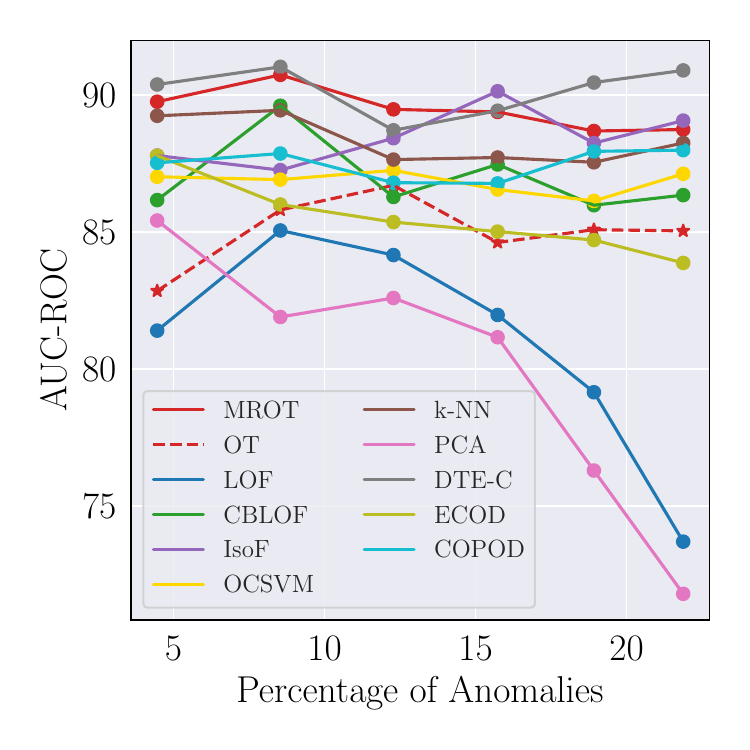}
        \caption{Mode 4.}
    \end{subfigure}\hfill
    \begin{subfigure}{0.31\linewidth}
        \centering
        \includegraphics[width=\linewidth]{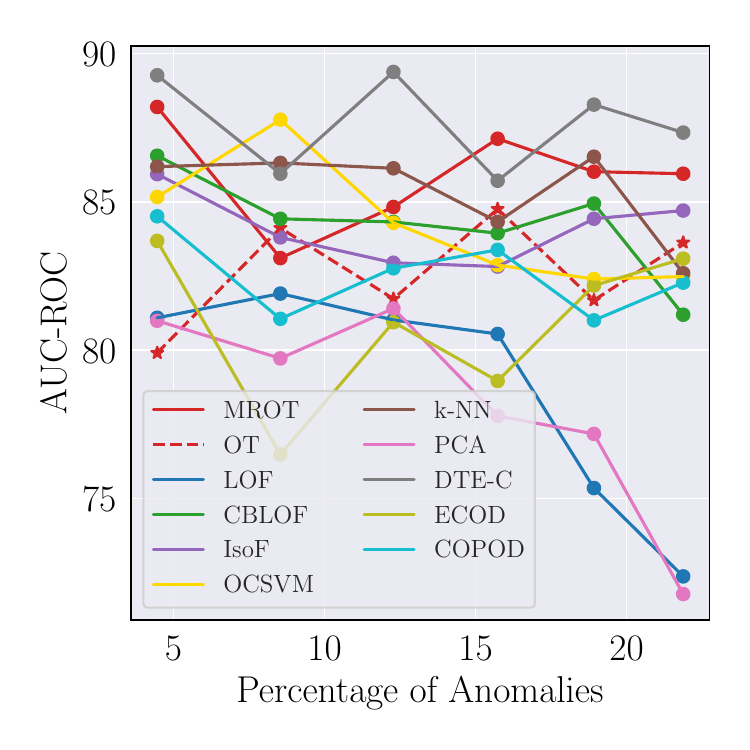}
        \caption{Mode 5.}
    \end{subfigure}\hfill
    \begin{subfigure}{0.31\linewidth}
        \centering
        \includegraphics[width=\linewidth]{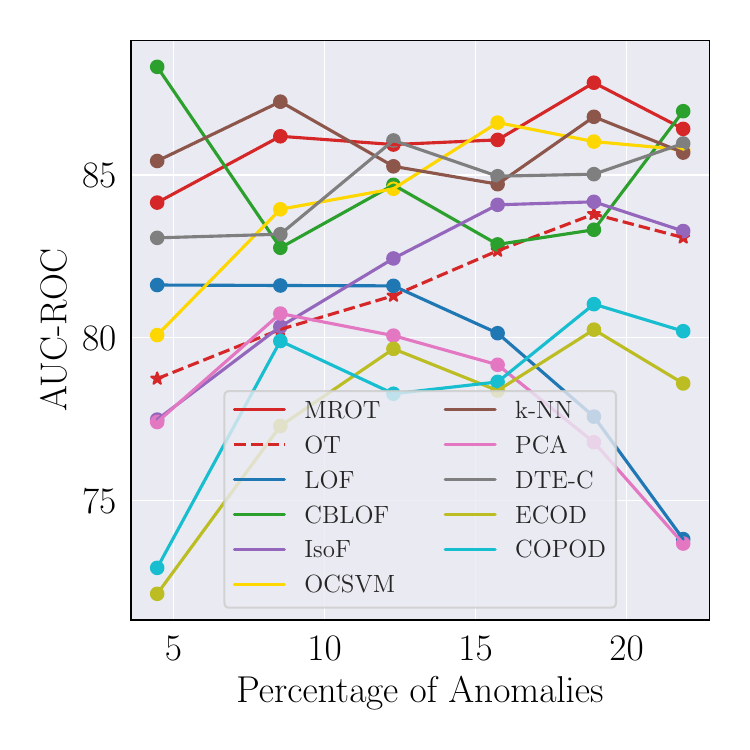}
        \caption{Mode 6.}
    \end{subfigure}\hfill
    \caption{AUC-ROC per percentage of anomalies on the \gls{te} benchmark.}
    \label{fig:tep-per-percentage-results}
\end{figure}

In Figure~\ref{fig:tep-per-percentage-results}, we show an overview of the AUC-ROC of different methods per percentage of anomalies. Overall, our \gls{munot} outperforms \gls{ot} with the regularized Coulomb cost on all modes. Furthermore, our method outperforms or stays competitive with other methods on most modes, such as 1, 3 and 6.

\end{document}